\documentclass{article}
\usepackage{ijcai24}

\usepackage{times}
\usepackage{soul}
\usepackage{url}
\usepackage[hidelinks]{hyperref}
\usepackage[utf8]{inputenc}
\usepackage[small]{caption}
\usepackage{graphicx}
\usepackage{amsmath}
\usepackage{amsthm}
\usepackage{booktabs}
\usepackage[ruled,vlined]{algorithm2e}
\usepackage[switch]{lineno}

\usepackage{subfiles}
\usepackage{subcaption}
\usepackage{multirow}

\usepackage{amsmath,amsfonts,bm}

\def\eqref#1{equation~\ref{#1}}

\def\1{\bm{1}}

\def\rvd{{\mathbf{d}}}

\def\vu{{\bm{u}}}
\def\vv{{\bm{v}}}
\def\vw{{\bm{w}}}
\def\vx{{\bm{x}}}
\def\vy{{\bm{y}}}

\DeclareMathAlphabet{\mathsfit}{\encodingdefault}{\sfdefault}{m}{sl}
\SetMathAlphabet{\mathsfit}{bold}{\encodingdefault}{\sfdefault}{bx}{n}

\def\gF{{\mathcal{F}}}

\def\gL{{\mathcal{L}}}

\def\gP{{\mathcal{P}}}

\def\gS{{\mathcal{S}}}

\def\gU{{\mathcal{U}}}

\def\gW{{\mathcal{W}}}
\def\gX{{\mathcal{X}}}

\def\sR{{\mathbb{R}}}

\newtheorem{theorem}{Theorem}
\newtheorem{definition}{Definition}
\newtheorem{proposition}{Proposition}
\newtheorem{lemma}{Lemma}
\newtheorem{remark}{Remark}

\title{Neural Sinkhorn Gradient Flow}

\author{
Huminhao Zhu$^1$
\and
Fangyikang Wang$^1$\and
Chao Zhang$^1$\and
Hanbin Zhao$^1$\And
Hui Qian$^1$\\
\affiliations
$^1$College of Computer Science and Technology, Zhejiang University\\
\emails
\{zhuhuminhao, wangfangyikang, zczju, zhaohanbin, qianhui\}@zju.edu.cn
}

\begin{document}

\maketitle

\begin{abstract}
    Wasserstein Gradient Flows (WGF) with respect to specific functionals have been widely used in the machine learning literature. 
    Recently, neural networks have been adopted to approximate certain intractable parts of the underlying Wasserstein gradient flow and result in efficient inference procedures. In this paper, we introduce the Neural Sinkhorn Gradient Flow (NSGF) model,  which parametrizes the time-varying velocity field of the Wasserstein gradient flow w.r.t. the Sinkhorn divergence to the target distribution starting a given source distribution.  
    We utilize the velocity field matching training scheme in NSGF, which only requires samples from the source and target distribution to compute an empirical velocity field approximation. 
    Our theoretical analyses show that as the sample size increases to infinity, the mean-field limit of the empirical approximation converges to the true underlying velocity field. 
    To further enhance model efficiency on high-dimensional tasks, 
    a two-phase NSGF++ model is devised, which first follows the Sinkhorn flow to approach the image manifold quickly ($\le 5$ NFEs) and then refines the samples along a simple straight flow.
    Numerical experiments with synthetic and real-world benchmark datasets support our theoretical results and demonstrate the effectiveness of the proposed methods. 
\end{abstract}

\section{Introdution}


The Wasserstein Gradient Flow (WGF) with respect to certain specific functional objective $\gF$ (denoted as $\gF$ Wasserstein gradient flow) is a powerful tool for solving optimization problems over the Wasserstein probability space.
Since the seminal work of \cite{jordan1998variational} which shows that the Fokker-Plank equation is the Wasserstein gradient flow with respect to the free energy, Wasserstein gradient flow w.r.t. different functionals have been widely used in various machine learning tasks such as Bayesian inference \cite{zhang2021dpvi}, reinforcement learning \cite{zhang2021wasserstein}, and mean-field games \cite{zhang2023mean}. 

\begin{figure*}[h!]
    \centering
    \begin{subfigure}{0.9\linewidth}
        \includegraphics[width=\linewidth]{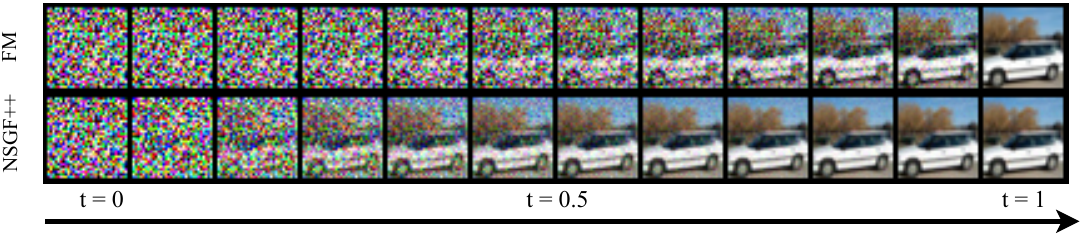}
    \end{subfigure}
    \caption{Tajectories comparison between the Flow matching and the NSGF++ model in CIFAR-10 task. we can see NSGF++ model quickly recovers the target structure and progressively optimizes the details in subsequent steps}
    \label{p1}
 \end{figure*}

\par
One recent trend in the Wasserstein gradient flow literature is to develop efficient generative modeling methods \cite{gao2019deep,gao2022deep,ansari2021refining,mokrov2021large,alvarez-melis2022optimizing,bunne2022proximal,fan2022variational}. 
In general, these methods mimic the Wasserstein gradient flow with respect to a specific distribution metric, driving a source distribution towards a target distribution. Neural networks are typically employed to approximate the computationally challenging components of the underlying Wasserstein gradient flow such as the time-dependent transport maps.
During the training process of these methods, it is common to require samples from the target distribution.  
After the training process, an inference procedure is often employed to generate new samples from the target distribution This procedure involves iteratively transporting samples from the source distribution with the assistance of the trained neural network.
Based on the chosen metric, these methods can be categorized into two main types.
\par
\textbf{Divergences Between Distributions With Exact Same Supports.} 
The first class of widely used metrics is the f-divergence, such as the Kullback-Leibler divergence and the Jensen-Shannon divergence. 
These divergences are defined based on the density ratio between two distributions and are only well-defined when dealing with distributions that have exactly the same support.
Within the scope of f-divergence Wasserstein gradient flow generative models,
neural networks are commonly utilized to formulate density-ratio estimators, as demonstrated by \cite{gao2019deep,ansari2021refining} and \cite{heng2022deep}.
However, as one can only access finite samples from target distributions in the training process, the support shift between the sample collections from the compared distributions may cause significant approximation error in the density-ratio estimators \cite{choi2022density}.
An alternative approach, proposed by \cite{fan2022variational}, circumvents these limitations by employing a dual variational formulation of the f-divergence. 
In this framework, two networks are employed to approximate the optimal variational function and the transport maps. These two components are optimized alternately. 
It's imperative to highlight that the non-convex and non-concave characteristics of their min-max objective can render the training inherently unstable \cite{hsieh2021limits}.
\par
\textbf{Divergences Between Distributions With Possible Different Supports.}
Another type of generative Wasserstein gradient flow model employs divergences that are well-defined for distributions with possible different supports. This includes free energy fuctionals \cite{mokrov2021large,bunne2022proximal}, the kernel-based metrics such as the Maximum-Mean/Sobolev Discrepancy \cite{mroueh2019sobolev,mroueh2020unbalanced} and sliced-Wasserstein distance \cite{liutkus2019sliced,du2023nonparametric}.
As these divergences can be efficiently approximated with samples, neural networks are typically used to directly model the transport maps used in the inference procedure.
In Wasserstein gradient flow methods, input convex neural networks (ICNNs, \cite{amos2017input}) are commonly used to approximate the transport map. However, recently, several works \cite{korotin2021neural} discuss the poor expressiveness of ICNNs architecture and show that it would result in poor performance in high-dimension applications.
Besides, the Maximum-Mean/Sobolev discrepancy Wasserstein gradient flow models are usually hard to train and are easy to trapped in poor local optima in practice \cite{arbel2019maximum},
since the kernel-based divergences are highly sensitive to the parameters of the kernel function \cite{li2017mmd,wang2018improving}. 
\cite{liutkus2019sliced,du2023nonparametric} consider sliced-Wasserstein WGF to build nonparametric generative Models which do not achieve high generation quality, it is an interesting work on how to combine sliced-Wasserstein WGF and neural network methods.

\par
\textbf{Contribution.}
In this paper, we investigate the Wasserstein gradient flow with respect to the Sinkhorn divergence, which is categorized under the second type of divergence and does not necessitate any kernel functions.
We introduce the \emph{Neural Sinkhorn Gradient Flow (NSGF)} model, which parametrizes the time-varying velocity field of the Sinkhorn Wasserstein gradient flow from a specified source distribution. 
The NSGF employs a velocity field matching scheme that demands only samples from the target distribution to calculate empirical velocity field approximations. 
Our theoretical analyses show that as the sample size approaches infinity, the mean-field limit of the empirical approximation converges to the true velocity field of the Sinkhorn Wasserstein gradient flow. 
Given distinct source and target data samples, our NSGF can be harnessed across a wide range of machine learning applications, including unconditional/conditional image generation, style transfer, and audio-text translation. 
To further enhance model efficiency on high-dimensional image datasets, 
a two-phase NSGF++ model is devised, which first follows the Sinkhorn flow to approach the image manifold quickly ($\le 5$ NFEs) and then refine the samples along a simple straight flow.
A novel phase-transition time predictor is proposed to transfer between the two phases.
We empirically validate NSGF on low-dimensional 2D data and NSGF++ on benchmark images (MNIST, CIFAR-10). 
Our findings indicate that our models can be trained to yield commendable results in terms of generation cost and sample quality, surpassing the performance of the neural Wasserstein gradient flow methods previously tested on CIFAR-10, to the best of our knowledge.


\section{Related Works}
\textbf{Sinkhorn Divergence in Machine Learning.}
Originally introduced in the domain of optimal transport, the Sinkhorn divergence emerged as a more computationally tractable alternative to the classical Wasserstein distance \cite{cuturi2013sinkhorn,peyre2017computational,feydy2019interpolating}.
Since its inception, Sinkhorn divergence has found applications across a range of machine learning tasks, including domain adaptation \cite{alaya2019screening,komatsu2021multi}, Sinkhorn barycenter \cite{luise2019sinkhorn,shen2020sinkhorn} and color transfer \cite{pai2021fast}.
Indeed, it has already been extended to single-step generative modeling methods, such as the Sinkhorn GAN and VAE \cite{genevay2018learning,deja2020end,patrini2020sinkhorn}. However, to the best of our knowledge, it has yet to be employed in developing efficient generative Wasserstein gradient flow models.

\par
\textbf{Neural ODE/SDE Based Diffusion Models.}
Recently, diffusion models, as a class of Neural ODE/SDE Based generative methods have achieved unprecedented success, which also transforms a simple density to the target distribution, iteratively \cite{song2019generative,ho2020denoising,song2021scorebased}.
Typically, each step of diffusion models only progresses a little by denoising a simple Gaussian noise, while each step in WGF models follows the most informative direction (in a certain sense).
Hence, diffusion models usually have a long inference trajectory.
In recent research undertakings, there has been a growing interest in exploring more informative steps within diffusion models. 
Specifically, flow matching methods \cite{lipman2023flow,liu2023flow,albergo2023building} establish correspondence between the source and target via optimal transport, subsequently crafting a probability path by directly linking data points from both ends. 
Notably, when the source and target are both Gaussians, their path is actually a Wasserstein gradient flow.
However, this property does not consistently hold for general data probabilities.
Moreover, \cite{tong2023improving,pooladian2023multisample} consider calculating the minibatch optimal transport map to guide data points connecting. 
Besides, \cite{das2023image} consider the shortest forward diffusion path for the Fisher metric and \cite{shaul2023kinetic} explore the conditional Gaussian probability path based on the principle of minimizing the Kinetic Energy.
Nonetheless, a commonality among many of these methods is their reliance on Gaussian paths for theoretical substantiation, thereby constraining the broader applicability of these techniques within real-world generative modeling.

\section{Preliminaries}
\subsection{Notations}
We denote $\vx = (x_1, \cdots , x_d) \in  \mathbb{R}^{d}$ and $\mathcal{X}\subset \mathbb{R}^{d}$ as a vector and a compact ground set in $ \mathbb{R}^{d}$, respectively. 
For a given point $\vx \in \mathcal{X}$, $\|\vx\|_p := \left(\sum_{i} x_i^p\right)^{\frac{1}{p}} $ denotes the $p$-norm on euclidean space, and $\delta_\vx$ stands for the Dirac (unit mass) distribution at point $\vx \in \mathcal{X}$.
$\mathcal{P}_2(\mathcal{X})$ denotes the set of probability measures on
$\mathcal{X}$ with finite second moment and $\mathcal{C}(\mathcal{X})$ denotes the space of continuous functions on $\mathcal{X}$. 
For a given functional 
$\mathcal{F}(\cdot):\mathcal{P}_2(\mathcal{X})\to \mathbb{R}$, $\frac{\delta\mathcal{F}(\mu_{t})}{\delta\mu}(\cdot): \mathbb{R}^{d}\to \mathbb{R}$
denotes its first variation at $\mu = \mu_{t}$. Besides, we use $\nabla$ and $\nabla\cdot()$ to denote the gradient and the divergence operator, respectively.

\subsection{Wasserstein distance and Sinkhorn divergence}
We first introduce the background of Wasserstein distance.
Given two probability measures $\mu,\nu\in \mathcal{P}_2(\mathcal{X})$, the $p$-Wasserstein distance $\mathcal{W}_p(\mu, \nu):\mathcal{P}_2(\mathcal{X}) \times \mathcal{P}_2(\mathcal{X})\to \mathbb{R}_{+}$
is defined as:
   \begin{equation}
      \label{Wasserstein}
      \mathcal{W}_p(\mu,\nu)=
          \inf_{\pi\in\Pi(\mu,\nu)}
          \left(
            \int_{\mathcal{X}\times\mathcal{X}}\left\lVert \vx-\vy\right\rVert^p \mathrm{d}\pi(\vx,\vy)
      \right) ^{\frac{1}{p}},
   \end{equation}
where $\Pi(\mu,\nu)$ denotes  the set of all probability couplings $\pi$ with marginals $\mu$ and $\nu$. 
The $\mathcal{W}_p$ distance aims to find a coupling $\pi$ so as to minimize the cost function $\left\lVert \vx-\vy\right\rVert^p$ of moving a probability mass from $\mu$ to $\nu$. It has been demonstrated that the $p$-Wasserstein distance is a valid metric on $\mathcal{P}_2(\mathcal{X})$, and $\left(\mathcal{P}_2(\mathcal{X}), \mathcal{W}_p\right)$ is referred to as the Wasserstein probability space \cite{villani2009optimal}.

Note that directly calculating $\mathcal{W}_p$ is computationally expensive, especially for high dimensional problems \cite{santambrogio2015optimal}.
Consequently, the entropy-regularized Wasserstein distance \cite{cuturi2013sinkhorn} is proposed to approximate equation
\eqref{Wasserstein} by regularizing the original problem with an entropy term:
\begin{definition}
    The entropy-regularized Wasserstein distance is formally defined as:
    \begin{equation}\label{entropy-regularized OT}
        \small
        \begin{aligned}
        &\mathcal{W}_{p,\varepsilon}(\mu,\nu)= \\
        &\inf_{\pi\in\Pi(\mu,\nu)}
        \left[
        \left(
            \int_{\mathcal{X}\times\mathcal{X}}\left\lVert \vx-\vy\right\rVert^p \mathrm{d}\pi(\vx,\vy)
        \right)^{\frac{1}{p}} 
         + \varepsilon KL(\pi|\mu \otimes \nu)  
        \right], 
        \end{aligned}
    \end{equation}
    where $\varepsilon > 0$ is a regularization coefficient, $\mu \otimes \nu$ denotes the product measure, i.e., $\mu \otimes \nu(\vx,\vy) =\mu(\vx)\nu(\vy)$, and $KL(\pi|\mu \otimes \nu)$ denotes the KL-divergence between $\pi$ and $\mu \otimes \nu$. 
\end{definition}
Generally, the computational cost of $\mathcal{W}_{p,\varepsilon}$ is much lower than $\mathcal{W}_{p}$, and can be efficiently calculated with Sinkhorn algorithms \cite{cuturi2013sinkhorn}.
Without loss of generality, we fix $p=2$ and abbreviate $\mathcal{W}_{2,\varepsilon} := \mathcal{W}_{\varepsilon}$ for ease of notion in the whole paper. 
According to Fenchel-Rockafellar theorem, the entropy-regularized Wasserstein problem $\mathcal{W}_{\varepsilon}$ \eqref{entropy-regularized OT} has an equivalent dual formulation, which is given as follows \cite{peyre2017computational}:
\begin{equation}
	\begin{aligned}\label{dual form}
		\mathcal{W}_{\varepsilon}(\mu, \nu) &=  \max _{f, g \in \mathcal{C}(\mathcal{X})}\langle\mu, f\rangle+\langle\nu, g\rangle  \\
		& -\varepsilon\left\langle\mu \otimes \nu, \exp \left(\frac{1}{\varepsilon}(f \oplus g-\mathrm{C})\right)-1\right\rangle,
	\end{aligned}
\end{equation}
where $C$ is the cost function in \eqref{entropy-regularized OT} and $f \oplus g$ is the tensor sum: $(x, y)\in \gX^{2}\mapsto f(x)+g(y) $. The maximizers $f_{\mu, \nu}$ and $g_{\mu, \nu}$ of \eqref{dual form} are called the $\gW_{\varepsilon}$-potentials of $\mathcal{W}_{\varepsilon}{(\mu, \nu)}$.
The following lemma states the optimality condition for the $\gW_{\varepsilon}$-potentials: 
\begin{lemma}\label{W-potentials}
	(Optimality \cite{cuturi2013sinkhorn})
	The $\gW_{\varepsilon}$-potentials $(f_{\mu, \nu}, g_{\mu, \nu})$ exist and are unique $(\mu, \nu)-a.e.$ up to an additive constant (i.e. $\forall K \in \sR, (f_{\mu, \nu} + K, g_{\mu, \nu} - K)$ is optimal). Moreover,
	\begin{equation}
		\mathcal{W}_{\varepsilon}(\mu,\nu) = \langle \mu, f_{\mu, \nu} \rangle + \langle \nu, g_{\mu, \nu} \rangle. 
	\end{equation}
\end{lemma}
We describe such the method in Appendix \ref{appA} for completeness. 
Note that, although computationally more efficient than the $\mathcal{W}_{p}$ distance, the $\mathcal{W}_{\varepsilon}$ distance is not a true metric, as there exists $\mu\in\mathcal{P}_2(\mathcal{X})$ such that $\mathcal{W}_{\varepsilon}(\mu,\mu) \neq 0$ when $\varepsilon \neq 0$, which restricts the applicability of $\mathcal{W}_{\varepsilon}$. 
As a result, the following Sinkhorn divergence $\mathcal{S}_{\varepsilon}(\mu, \nu):\mathcal{P}_2(\mathcal{X}) \times \mathcal{P}_2(\mathcal{X})\to \mathbb{R}$ is proposed \cite{peyre2017computational}:
\begin{definition}
   Sinkhorn divergence:
   \begin{equation}\label{Sinkhorn divergence}
    \mathcal{S}_{\varepsilon}(\mu, \nu)=\mathcal{W}_{\varepsilon}(\mu, \nu)-
    \frac{1}{2}\left(\mathcal{W}_{\varepsilon}(\mu, \mu)+\mathcal{W}_{\varepsilon}(\nu, \nu)\right).
   \end{equation}
\end{definition}
$\mathcal{S}_{\varepsilon}(\mu, \nu)$  is nonnegative, bi-convex thus a valid metric on $\mathcal{P}_2(\mathcal{X}) $ and metricize the convergence in law. 
Actually $\mathcal{S}_{\varepsilon}(\mu, \nu)$ interpolates the Wasserstein distance ($\epsilon \to 0$) and the Maximum Mean Discrepancy ($\epsilon \to \infty$) \cite{feydy2019interpolating}. 

\subsection{Gradient flows}
Consider an optimization problem over $\mathcal{P}_2(\mathcal{X})$: 
\begin{equation}
    \min_{\mu\in \mathcal{P}_2(\mathcal{X})}\mathcal{F}(\mu):=\mathcal{D}(\mu|\mu^{*}).
\end{equation}
where $\mu^{*}$ is the target distribution, $\mathcal{D}$ is the divergence we choose. We consider now the problem of transporting mass from an initial distribution $\mu_{0}$ to a target distribution $\mu^{*}$, by finding a continuous probability path $\mu_{t}$ starting from $\mu_{0} = \mu$ that converges to $\mu^{*}$ while decreasing $\gF(\mu_{t})$. 
To solve this optimization problem, one can consider a descent flow of $\mathcal{F}(\mu)$ in the Wasserstein space,
which transports any initial distribution $\mu_0$ towards the target distribution $\mu^{*}$. Specifically, the descent flow of $\gF(\mu)$ is described by the following continuity equation \cite{ambrosio2005gradient,villani2009optimal,santambrogio2017euclidean}:
\begin{equation} \label{continuity equation}
   \frac{\partial \mu_t(x)}{\partial t} = -\nabla\cdot(\mu_t(x)\vv_{t}(x)).
\end{equation}  
where $\vv_{\mu_t} : \gX \to \gX$ is a velocity field that defines the direction of position transportation.
To ensure a descent of $\mathcal{F}(\mu_t)$ over time $t$, the velocity field $\vv_{\mu_t}$ should satisfy the following inequality (\cite{ambrosio2005gradient}):
\begin{equation}
    \frac{\mathrm{d}\mathcal{F}(\mu_t)}{\mathrm{d}t} = \int \langle\nabla \frac{\delta \mathcal{F}(\mu_t)}{\delta \mu},\vv_{t} \rangle \mathrm{d} \mu_t \leq 0 . 
\end{equation} 
A straightforward choice of $\vv_{t}$ is $\vv_{t} = -\nabla\frac{\delta \mathcal{F}(\mu_t)}{\delta \mu}$, 
which is actually the steepest descent direction of $\mathcal{F}(\mu_t)$. When we select this $\vv_t$, we refer to the aforementioned continuous equation as the \emph{Wasserstein gradient flow} of $\mathcal{F}$. We give the definition of the first variation in the appendix for the sake of completeness of the article.

\section{Methodology}
In this section, we first introduce the  Sinkhorn Wasserstein gradient flow and investigate its convergence properties. Then, we develop our Neural Sinkhorn Gradient Flow model, which consists of a velocity field matching training procedure and a velocity field guided inference procedure.
Moreover, we theoretically show that the mean-field limit of the empirical approximation used in the training procedure converges to the true velocity field of the Sinkhorn Wasserstein gradient flow. 
%
\subsection{Sinkhorn Wasserstein gradient flow}
Based on the definition of the Sinkhorn divergence, we construct our Sinkhorn objective $\gF_\varepsilon(\cdot)=\gS_{\varepsilon}(\cdot, \mu^*)$, 
where $\mu^*$ denotes the target distribution.
The following theorem gives the first variation of the Sinkhorn objective.

\begin{theorem}\label{first variation of Sinkhorn divergence}
(First variation of the Sinkhorn objective \cite{luise2019sinkhorn})
Let $\varepsilon > 0$. Let $(f_{\mu, \mu^*}, g_{\mu, \mu^*})$ be the $\gW_{\varepsilon}$-potentials of $\gW_{\varepsilon}{(\mu, \mu^*)}$ and $(f_{\mu, \mu}, g_{\mu, \mu})$ be the $\gW_{\varepsilon}$-potentials of $\gW_{\varepsilon}{(\mu, \mu)}$. The first variation of the Sinkhorn objective $\gF_\varepsilon$ is
   \begin{equation}\label{first variation}
      \frac{\delta \gF_\varepsilon}{\delta \mu} =  f_{\mu, \mu^\ast} - f_{\mu, \mu}.
   \end{equation}
\end{theorem}

According to Theorem \ref{first variation of Sinkhorn divergence}, we can construct the Sinkhorn Wasserstein gradient flow by setting the velocity field $\vv_{t}$ in the continuity equation \eqref{continuity equation} as $\vv^{\gF_{\varepsilon}}_{\mu_t} = -\nabla\frac{\delta \gF_\varepsilon(\mu_t)}{\delta \mu_t} = \nabla f_{\mu_t, \mu_t} - \nabla f_{\mu_t, \mu^\ast}$.
\begin{proposition}
   Consider the Sinkhorn Wasserstein gradient flow described by the following continuity equation:
   \begin{equation}\label{PDE flow}
      \frac{\partial \mu_t(\vx)}{\partial t} = -\nabla\cdot(\mu_t(\vx)(\nabla f_{\mu_t, \mu_t}(\vx) - \nabla f_{\mu_t, \mu^\ast}(\vx))).
   \end{equation}
   The following local descending property of $\mathcal{F}_\varepsilon$ holds:
   \begin{equation}\label{ODE of Sinkhorn}
      \frac{\mathrm{d} \mathcal{F}_\varepsilon\left(\mu_t\right)}{\mathrm{d} t}=-\int\left\|\nabla f_{\mu_t, \mu_t}(\vx) - \nabla f_{\mu_t, \mu^\ast}(\vx)\right\|^2 \mathrm{~d} \mu_t,
   \end{equation}
	where the r.h.s.  equals 0 if and only if $\mu_t = \mu^*$. 
\end{proposition}
\subsection{Velocity-fields Matching}
We now present our NSGF method, the core of which lies in training a neural network to approximate the time-varying velocity field $\vv_{\mu_t}^{\gF_{\varepsilon}}$ induced by Sinkhorn Wasserstein gradient flow.  
Given a target probability density path $\mu_t(x)$ and it's corresponding velocity field $\vv_{\mu_t}^{\gS_{\varepsilon}}$, which generates $\mu_t(x)$, we define the velocity field matching objective as follows:
\begin{equation} \label{goal}
   \min _\theta \mathbf{E}_{t\sim [0, T], x \sim \mu_t}\left[\left\|\vv^\theta(x, t)-\vv_{\mu_t}^{\gS_{\varepsilon}}\left(x \right)\right\|^2\right]. 
\end{equation}
\par 
To construct our algorithm, we utilize independently and identically distributed ($\mathrm{i.i.d}$) samples denoted as $\{Y_i \}_{i=1}^{n}\in \mathbb{R}^{d}$ , which are drawn from an unknown target distribution $\mu^{*}$ a common practice in the field of generative modeling. Given the current set of samples $\{\tilde{X}_i^t \}_{i=1}^{n}\sim \mu_t $, our method calculates the velocity field using the $\gW_{\varepsilon}$-potentials (Lemma \ref{W-potentials}) $f_{\tilde{\mu}_t, \tilde{\mu}^{*}} $ and $f_{\tilde{\mu}_t, \tilde{\mu}_t}$ based on samples. Here, $\tilde{\mu}_t$ and $\tilde{\mu}^{*}$ represent discrete Dirac distributions. 
\begin{remark}
   In the discrete case, $\gW_{\varepsilon}$-potentials \eqref{W-potentials} can be computed by a standard method in \cite{genevay2016stochastic}. In practice, we use the efficient implementation of the Sinkhorn algorithm with GPU acceleration from the GeomLoss package \cite{feydy2019interpolating}. 
\end{remark}
The corresponding finite sample velocity field approximation can be computed as follows:
\begin{equation} 
   \hat{\vv}_{\tilde{\mu}_t}^{\gF_{\varepsilon}}(\tilde{X}_i^t)=\nabla_{\tilde{X}_i^t} f_{\tilde{\mu}_t, \tilde{\mu}_t}(\tilde{X}_i^t)-\nabla_{\tilde{X}_i^t} f_{\tilde{\mu}_t, \tilde{\mu}^{*}}(\tilde{X}_i^t).
\end{equation}
Subsequently, we derive the particle formulation corresponding to the flow formulation \eqref{PDE flow}.
\begin{equation}\label{ODE}
   \mathrm{d}\tilde{X}^t_i= \hat{\vv}_{\tilde{\mu}_t}^{\gF_\epsilon}\left(\tilde{X}_i^t\right)\mathrm{d}t, i=1,2, \cdots n.
\end{equation}
In the following proposition, we investigate the mean-field limit of the particle set $\{\tilde{X}_i^t\}_{i = 1, \cdots, M}$.
\begin{theorem}(Mean-field limits.)\label{theorem2-1}
   Suppose the empirical distribution $\tilde{\mu}_0$ of $M$ particles weakly converges to a distribution $\mu_0$ when $M\to \infty$. Then, the path of equation \eqref{ODE} starting from $\tilde{\mu}_0$ weakly converges to a solution of the following partial differential equation starting from $\mu_0$ when $M\to \infty$:
   \begin{equation}\label{PDE15}
      \frac{\partial \mu_t(x)}{\partial t} = -\nabla\cdot(\mu_t(x)\nabla\frac{\delta \gF_\varepsilon(\mu_t)}{\delta \mu_t}).
   \end{equation}
   which is actually the gradient flow of Sinkhorn divergence $\gF_{\varepsilon}$ in the Wasserstein space.
\end{theorem}
The following proposition shows that the goal of the velocity field matching objective \eqref{goal} can be regarded as approximating the steepest local descent direction with neural networks.
\begin{proposition}(Steepest local descent direction.)\label{Frechet derivative of Sinkhorn divergence}
   Consider the infinitesimal transport $T(x) = x + \lambda \phi$. The Fr{\'e}chet derivative under this particular perturbation,
   \begin{equation}
      \begin{aligned}
         &\frac{\rvd }{\rvd \lambda} \gF_\varepsilon(T_{\#}\mu)|_{\lambda = 0}  = 
         \lim _{\lambda \rightarrow 0} \frac{\gF_\varepsilon\left(T_{\#}\mu\right)-\gF_\varepsilon\left(\mu\right)}{\lambda} \\
         =& \int_{\gX} \nabla f_{\mu, \mu^\ast}(\vx)\phi(\vx)d\mu  - \int_{\gX} \nabla f_{\mu, \mu}(\vx)\phi(\vx)\rvd\mu
         ,
      \end{aligned}
   \end{equation}
   and the steepest local descent direction is $\phi = \frac{\nabla f_{\mu, \mu^\ast}(\vx)-f_{\mu, \mu}(\vx)}{\| \nabla f_{\mu, \mu^\ast}(\vx)-f_{\mu, \mu}(\vx) \|}$.
\end{proposition}

\subsection{Minibatch Sinkhorn Gradient Flow and Experience Replay}
\begin{algorithm}[t]
   \resizebox{0.42\textwidth}{!}{
   \begin{minipage}{\linewidth}
   \SetKwInOut{KwIn}{Input}
   \SetKwInOut{KwOut}{Output}

   \KwIn{number of time steps $T$, batch size $n$, gradient flow step size $\eta>0$, empirical or samplable distribution $\mu_0$ and $\mu^*$, neural network parameters $\theta$, optimizer step size $\gamma > 0$}
   \tcc{Build trajectory pool}
   \While{$\text{Building}$}{
      \tcc{Sample batches of size $n$ $i.i.d.$ from the datasets}
      $ \tilde{X}_i^0 \sim \mu_0,\quad \tilde{Y_i} \sim \mu^*, i=1,2, \cdots n $.\\
      \For{$t = 0, 1, \cdots T$}{
         $\text{calculate} f_{\tilde{\mu}_t, \tilde{\mu}_t}\left(\tilde{X}_i^t\right),  f_{\tilde{\mu}_t,\tilde{\mu}^*}\left(\tilde{X}_i^t\right)$ . \\
         $\hat{\vv}_{\mu_t}^{\gF_\epsilon}\left(\tilde{X}_i^t\right)=\nabla f_{\tilde{\mu}_t, \tilde{\mu}_t}\left(\tilde{X}_i^t\right) - \nabla f_{\tilde{\mu}_t,\tilde{\mu}^*}\left(\tilde{X}_i^t\right)$ .\\
         $\tilde{X}_i^{t+1}=\tilde{X}_i^t+\eta \hat{\vv}_t^{\gF_\epsilon}\left(\tilde{X}_i^t\right)$. \\
         $\text { store all} \left(\tilde{X}_i^t, \hat{\vv}_t^{\gF_\epsilon} \left(\tilde{X}_i^{t}\right)\right) \text{pair into the pool},\quad i=1,2, \cdots n .$
      }
   }
   \tcc{velocity field matching}
   \While{$\text{Not convergence}$}{
      $\text{from trajectory pool sample pair} \left(\tilde{X}_i^t, \hat{\vv}_t^{\gF_\epsilon} \left(\tilde{X}_i^{t}\right)\right) $.\\
      $\mathcal{L}(\theta)= \left\|\vv^\theta(\tilde{X}_i^t, t)-\hat{\vv}_{\mu_t}^{\gF_{\varepsilon}}\left(\tilde{X}_i^t \right)\right\|^2$, \\
      $\theta \leftarrow \theta-\gamma \nabla_{\theta} \mathcal{L}\left(\theta\right)$ .
   }

   \KwOut{$\theta$ parameterize the time-varying velocity field}

\caption{\textbf{Velocity field matching training} }
\label{velocity field}

\end{minipage}
}
\end{algorithm}

\begin{algorithm}[t]
   \resizebox{0.47\textwidth}{!}{\begin{minipage}{\linewidth}
   \SetKwInOut{KwIn}{Input}
   \SetKwInOut{KwOut}{Output}

   \KwIn{number of time steps $T$, inference step size $\eta$, time-varying velocity field $\vv^\theta$, prior samples $ \tilde{X}_i^0 \sim \tilde{\mu}_0$}
   
   \For{$t = 0, 1, \cdots T$}{
      $\tilde{X}_i^{t+1}=\tilde{X}_i^t+\eta \vv^\theta\left(\tilde{X}_i^t, t\right), i=1,2, \cdots n . $ 
   }
   
   \KwOut{$\tilde{X}_i^{T}$ as the results.}

\caption{\textbf{Inference via velocity field} }
\label{velocity field inference}
\end{minipage}
}
\end{algorithm} 

According to Theorem \ref{theorem2-1}, we construct our NSGF method based on minibatches.
We utilize a set of discrete targets, denoted as $\{Y_i \}_{i=1}^{n}$, where $n$ represents the batch size, to construct the Sinkhorn Gradient Flow starting from random Gaussian noise or other initial distributions.  
As indicated by Theorem \ref{theorem2-1}, the mean-field limit converges to the true Sinkhorn Gradient flow when the batch size approaches $\infty$.
Note that in practice, we only use a moderate batch size for computation efficiency, and the experimental results demonstrate that this works well for practical generative tasks. 

Considering the balance between expensive training costs and training quality, we opted to first build a trajectory pool of Sinkhorn gradient flow and then sample from it to construct the velocity field matching algorithm. Our method draws inspiration from experience replay, a common technique in reinforcement learning, adapting it to enhance our model's effectiveness \cite{mnih2013playing,silver2016mastering}.
Once we calculate the time-varying velocity field $\hat{\vv}_{\mu_t}^s (\tilde{X}_i^{t})$, we can parameterize the velocity field using a straightforward regression method.
The velocity field matching training procedure is outlined in Algorithm \ref{velocity field}.




Once obtained a feasible velocity field approximation $\vv^\theta$, one can generate new samples by iteratively employing the explicit Euler discretization of the Equation \eqref{ODE} to drive the samples to the target. 
Note that various other numerical schemes, such as the implicit Euler method \cite{platen2010numerical} and Runge-Kutta methods \cite{butcher1964implicit}, can be employed. In this study, we opt for the first-order explicit Euler discretization method \cite{suli2003introduction} due to its simplicity and ease of implementation. We leave the exploration of higher-order algorithms for future research.

\subsection{NSGF++}

\begin{figure}[t]
   \centering
   \begin{subfigure}{0.95\linewidth}
       \includegraphics[width=\linewidth]{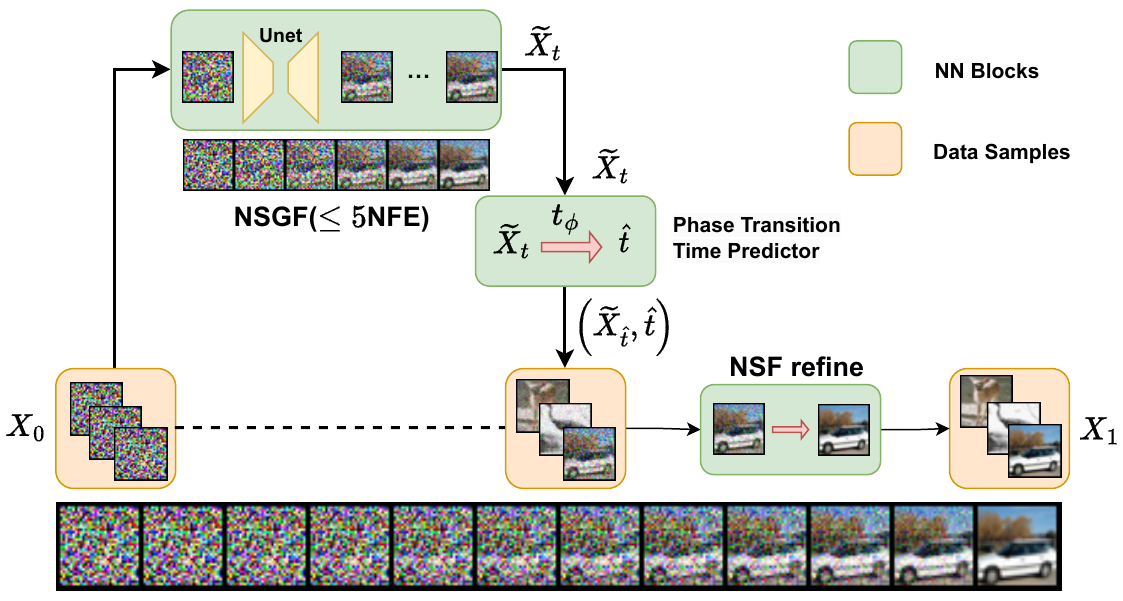}
   \end{subfigure}
   \caption{NSGF++ framework}
   \label{p2}
   \vspace{-3mm}
\end{figure}

In this subsection, we propose an approach to enhance the performance of NSGF on high-dimensional datasets.
As a trajectory pool should be first constructed in the velocity field matching training procedure of NSGF,
the storage and computation costs would greatly hinter the usage of NSGF in large-scale tasks.
To tackle this problem, we propose a two-phase NSGF++ algorithm, which first follows the Sinkhorn gradient flow to approach the image manifold quickly and then refine the samples along a simple straight flow.
Specifically, our NSGF++ model consists of three components, (1) a NSGF model trained on $T \le 5$ time steps, (2) a Neural Straight Flow (NSF) model trained via velocity field matching on a straight flow $X_t  \sim (1-t) P_0 + t P_1, t\in [0,1]$, which has also been used in existing FM models,  
(3) a phase-transition 
time predictor to transfer from NSGF to the NSF.
Here, we train the time predictor $t_\phi: \mathcal{X} \to [0,1]$ with the following regression objective:
$$
L(\phi) = E_{t\in \gU(0, 1), X_{t} \sim P_{t}} ||t - t_{\phi}(X_t)||^2.
$$
Note that the training of the straight NSF model and the time predictor is simulated free and need no extra storage.
As a result, the training cost of NSFG++ is similar to existing FM models, since the computation cost of NSF is nearly the same as FM, and the 5-step NSGF and the time predictor is easy to train.

In the inference of NSGF++, we first follow the NSGF with less than 5 NFEs form $X_0 \sim P_0$ to obtain $\tilde X_t$, then transfer it with the time predictor $t_\phi$, and obtain our final output by refining the transferred sample with the NSF model from $t_\phi(\tilde X_t)$, as shown in Figure \ref{p2}.

\begin{table*}[t]
	\centering
    \resizebox{1.0\textwidth}{!}{
	\begin{tabular}{c|ccccc|ccccc}
		\toprule
		\multirow{2}*{Algorithm} & \multicolumn{5}{c}{2-Wasserstein distance (10 steps)} & \multicolumn{5}{c }{2-Wasserstein distance (100 steps)} \\
		~    & 8gaussians & 8gaussians-moons & moons & scurve & checkerboard & 8gaussians & 8gaussians-moons & moons & scurve & checkerboard \\
		\hline
        NSGF (ours) & \textbf{0.285} & \textbf{0.144} & \textbf{0.077} & \textbf{0.117} & \textbf{0.252} & 0.278 & \textbf{0.144} & \textbf{0.067} & \textbf{0.110} & \textbf{0.147}\\
        JKO-Flow   & 0.290 & 0.177  & 0.085 & 0.135 & 0.269 & 0.274 & 0.167 & 0.085 & 0.123 & 0.160\\
        EPT   & 0.295 & 0.180  & 0.082 & 0.138 & 0.277 & 0.289 & 0.176 & 0.080 & 0.118 & 0.163 \\
        \hline
        OT-CFM     & 0.289 & 0.173  & 0.088 & 0.149 & 0.253 & \textbf{0.269} & 0.165 & 0.078 & 0.127 & 0.159\\
        \hline
	     1-RF       & 0.427 & 0.294  & 0.107 & 0.169 & 0.396 & 0.415 & 0.293 & 0.099 & 0.136 & 0.166\\
	     2-RF       & 0.428 & 0.311  & 0.125 & 0.171 & 0.421 & 0.430 & 0.311 & 0.121 & 0.136 & 0.170\\
        3-RF       & 0.421 & 0.298  & 0.110 & 0.170 & 0.413 & 0.414 & 0.297 & 0.103 & 0.140 & 0.170\\
        SI         & 0.435 & 0.324  & 0.134 & 0.187 & 0.427 & 0.411 & 0.294 & 0.096 & 0.139 & 0.166\\
        FM         & 0.423 & 0.292  & 0.111 & 0.171 & 0.417 & 0.415 & 0.290 & 0.097 & 0.135 & 0.165\\
		\bottomrule
	\end{tabular}}
	\caption{Comparison of neural gradient-flow-based methods and neural ODE-based diffusion models over five data sets with 10/100 Euler steps. The principle of steps in JKO-flow means backward Eulerian method steps (JKO steps).}
	\label{2D}
	 \vspace{-5mm}
\end{table*}

\begin{figure*}[t]
   \centering
   \begin{subfigure}{0.24\linewidth}
       \includegraphics[width=\linewidth]{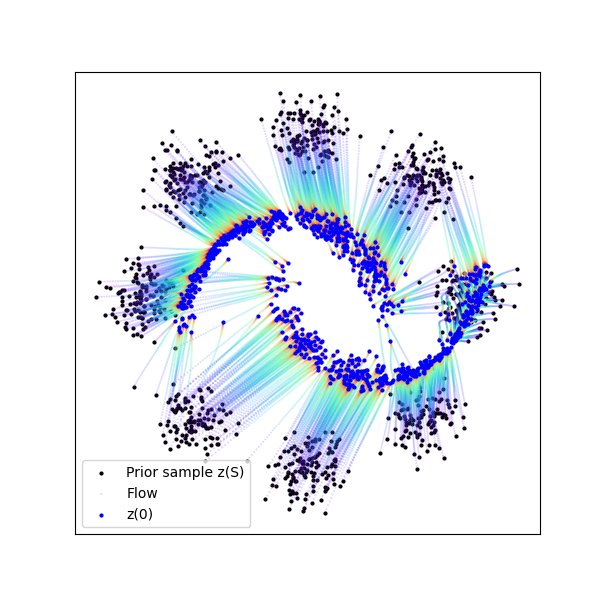}
       \caption{NSGF}
   \end{subfigure}
   \begin{subfigure}{0.24\linewidth}
      \includegraphics[width=\linewidth]{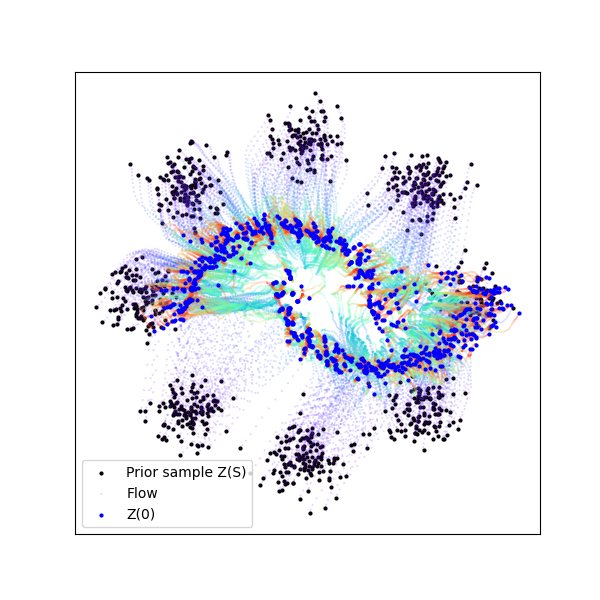}
      \caption{EPT}
  \end{subfigure}
   \begin{subfigure}{0.24\linewidth}
       \includegraphics[width=\linewidth]{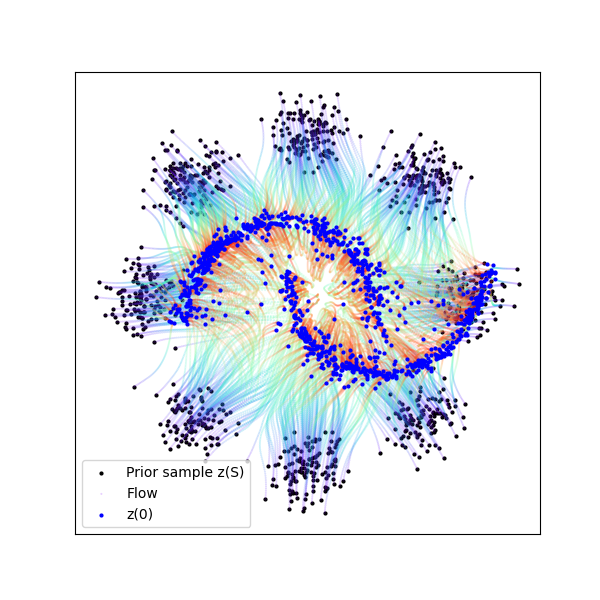}
       \caption{FM}
   \end{subfigure}
   \begin{subfigure}{0.24\linewidth}
       \includegraphics[width=\linewidth]{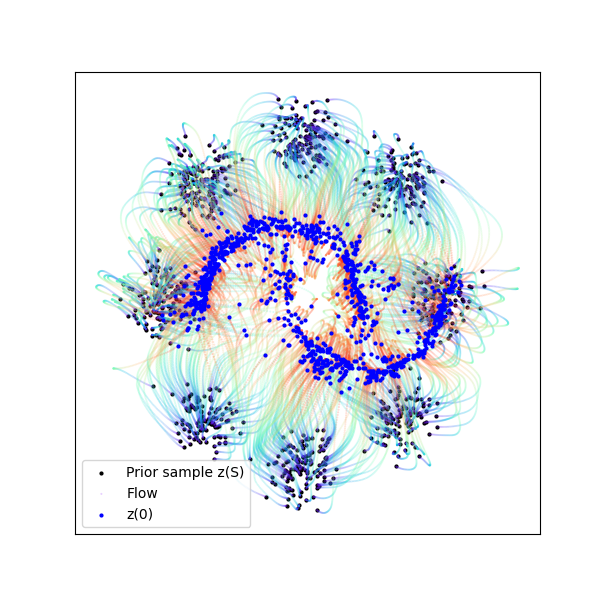}
       \caption{SI}
   \end{subfigure}
   \caption{Visualization results for 2D generated paths. We show different methods that drive the particle from the prior distribution (black) to the target distribution (blue). The color change of the flow shows the different number of steps (from blue to red means from $0$ to $T$).}
    \vspace{-5mm}
   \label{8gaussians-moons flow}
\end{figure*}

\begin{figure}[t]
   \captionsetup[subfigure]{labelformat=empty}
   \centering
   \tiny      
   \makebox[0pt][r]{\makebox[15pt]{\raisebox{15pt}{\rotatebox[origin=c]{90}{NSGF}}}}
   \begin{subfigure}{.19\linewidth}
   \includegraphics[width=\linewidth,height= .8\linewidth]{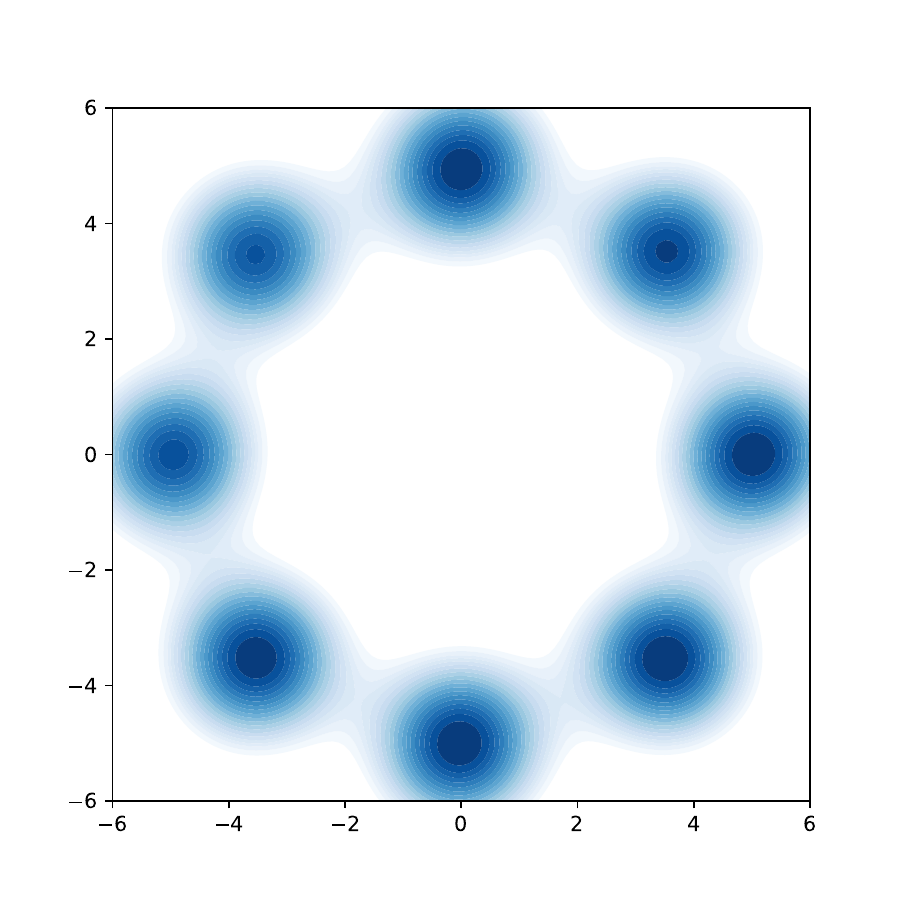}
   \end{subfigure}
   \begin{subfigure}{.19\linewidth}
         \includegraphics[width=\linewidth,height= .8\linewidth]{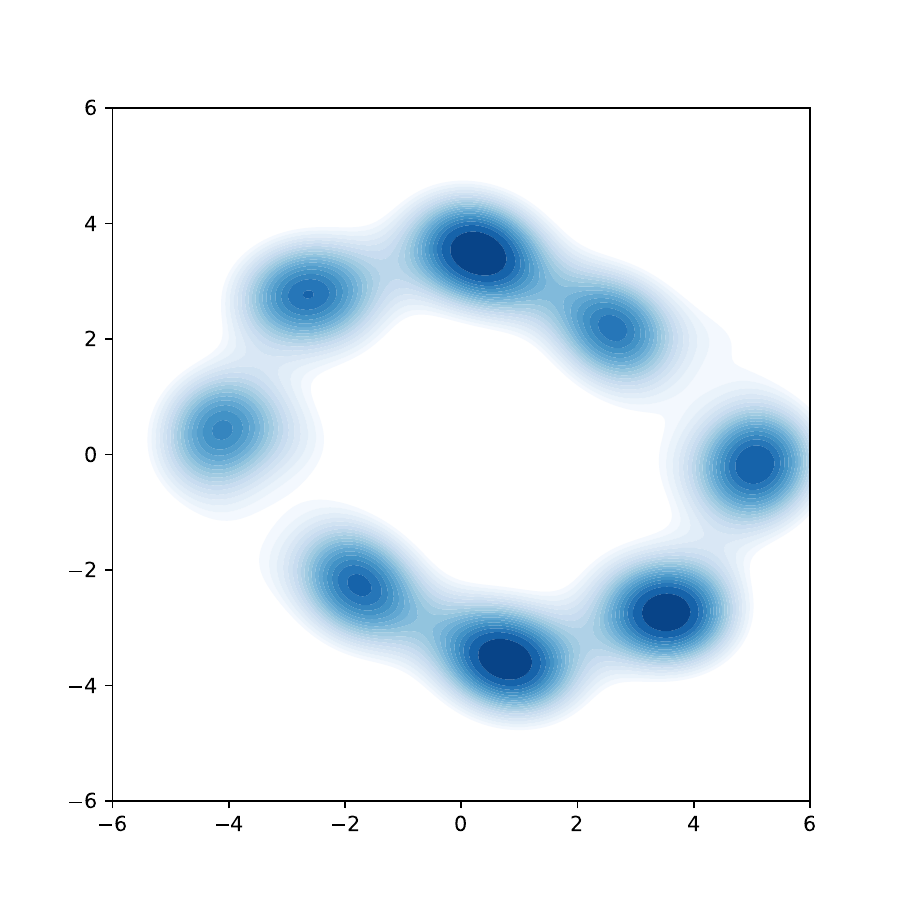}
   \end{subfigure}
   \begin{subfigure}{.19\linewidth}
         \includegraphics[width=\linewidth,height= .8\linewidth]{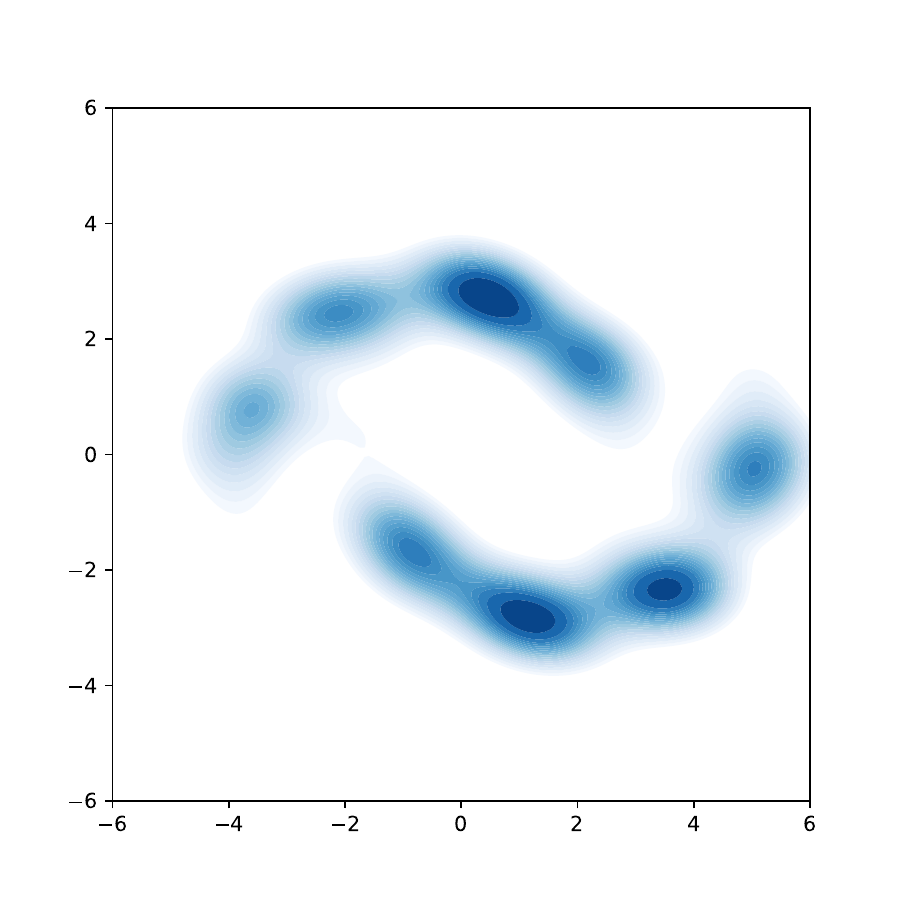}
   \end{subfigure}
   \begin{subfigure}{.19\linewidth}
         \includegraphics[width=\linewidth,height= .8\linewidth]{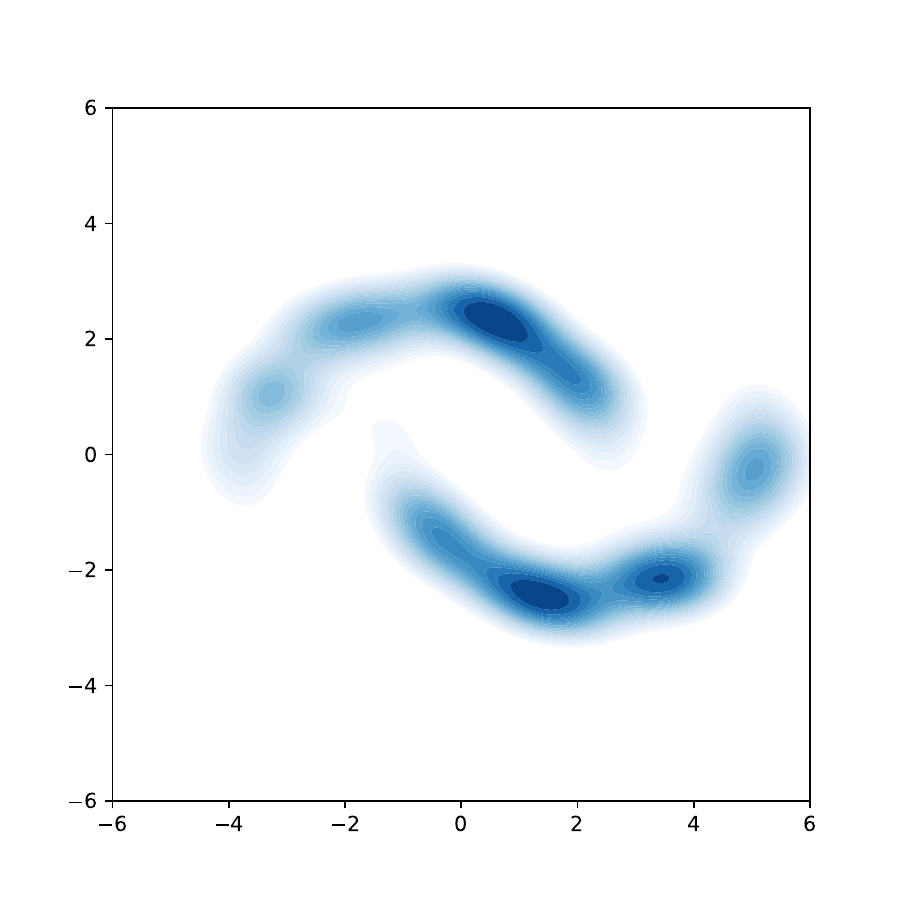}
   \end{subfigure}
   \begin{subfigure}{.19\linewidth}
         \includegraphics[width=\linewidth,height= .8\linewidth]{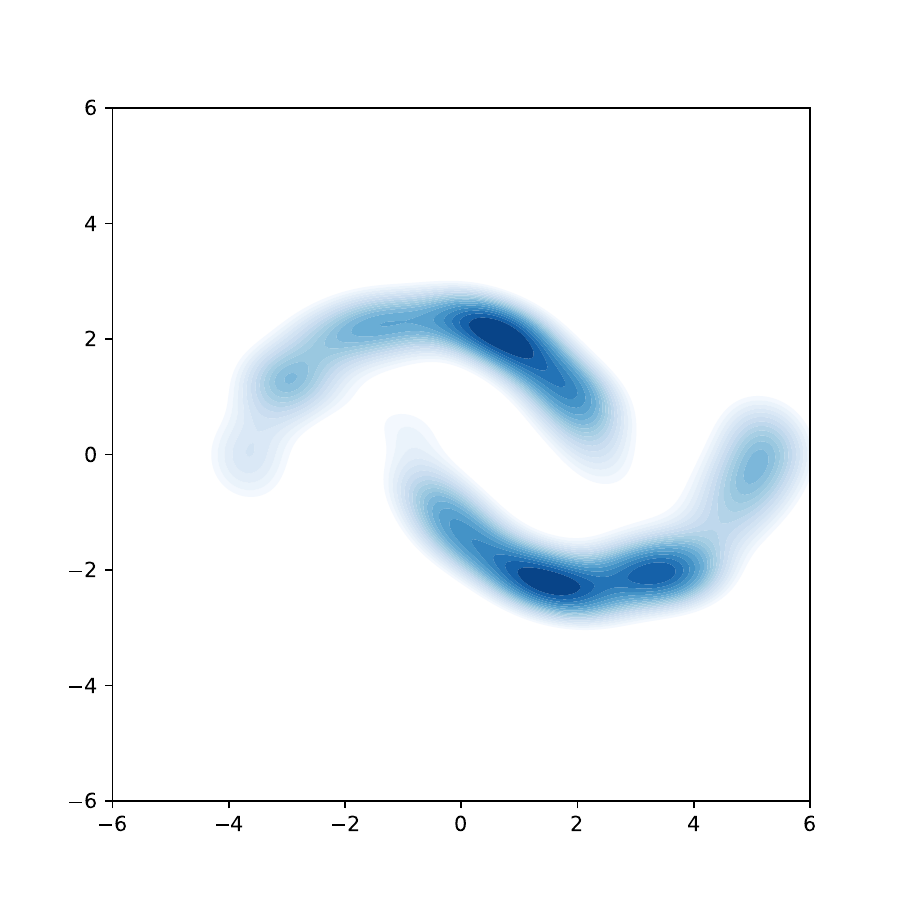}
   \end{subfigure}

   \makebox[0pt][r]{\makebox[15pt]{\raisebox{15pt}{\rotatebox[origin=c]{90}{EPT}}}}
   \begin{subfigure}{.19\linewidth}
      \includegraphics[width=\linewidth,height= .8\linewidth]{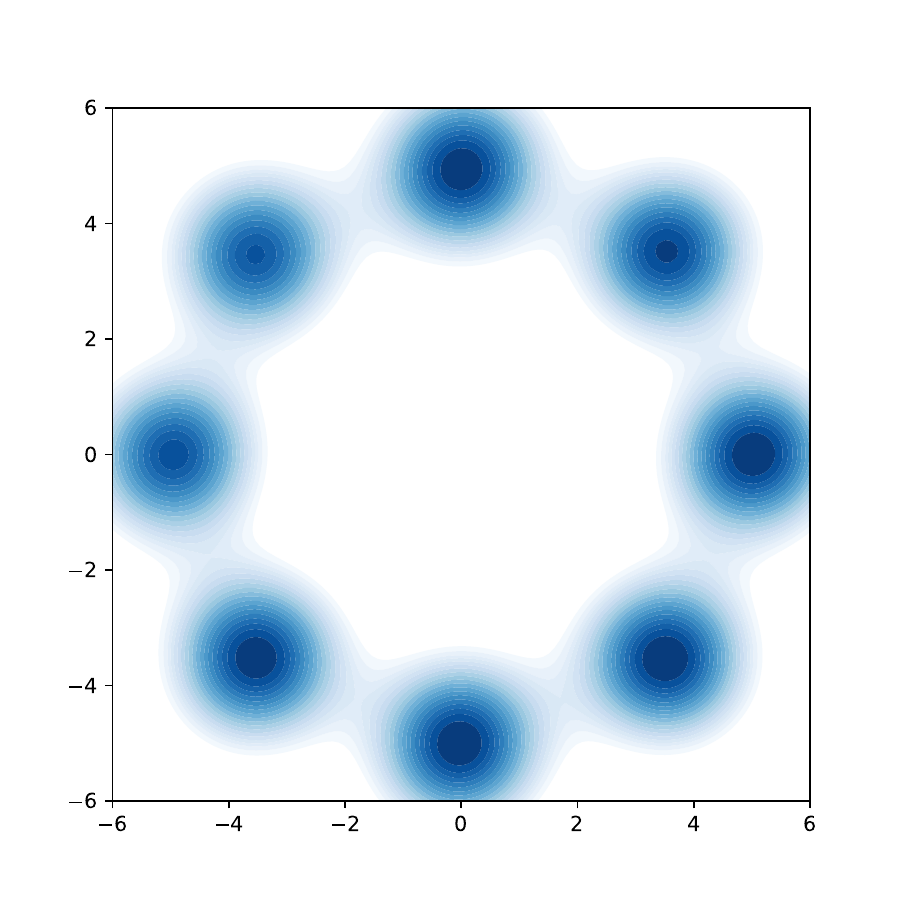}
   \end{subfigure}
   \begin{subfigure}{.19\linewidth}
      \includegraphics[width=\linewidth,height= .8\linewidth]{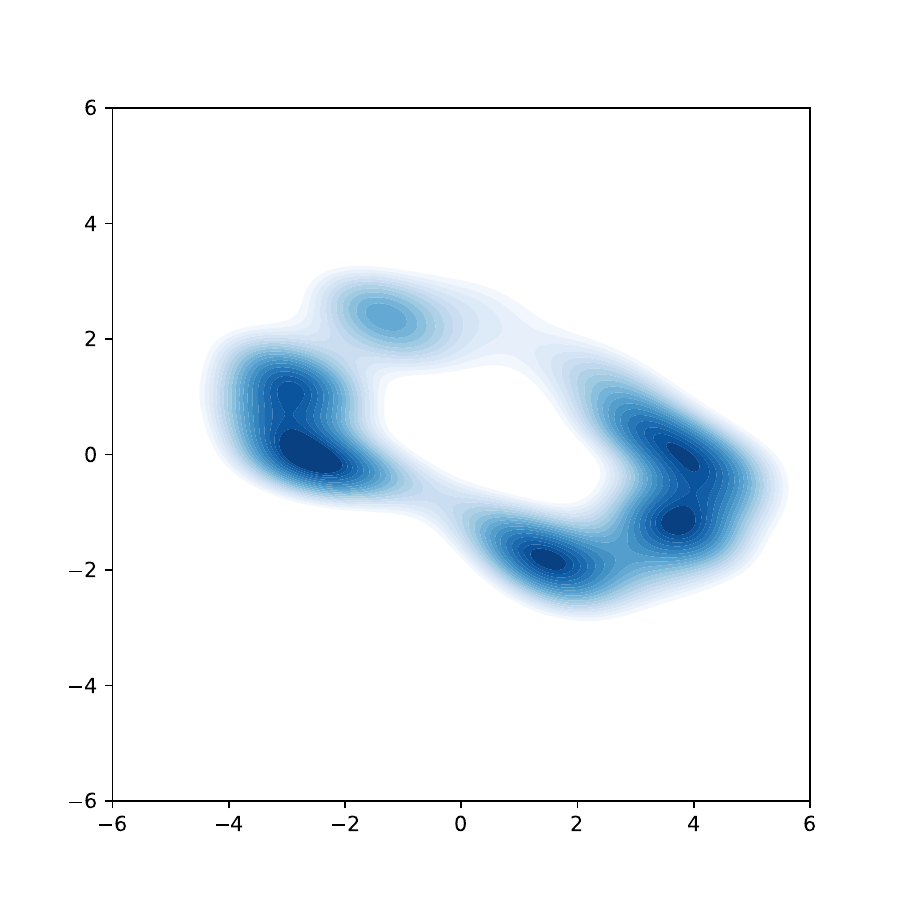}
   \end{subfigure}
   \begin{subfigure}{.19\linewidth}
      \includegraphics[width=\linewidth,height= .8\linewidth]{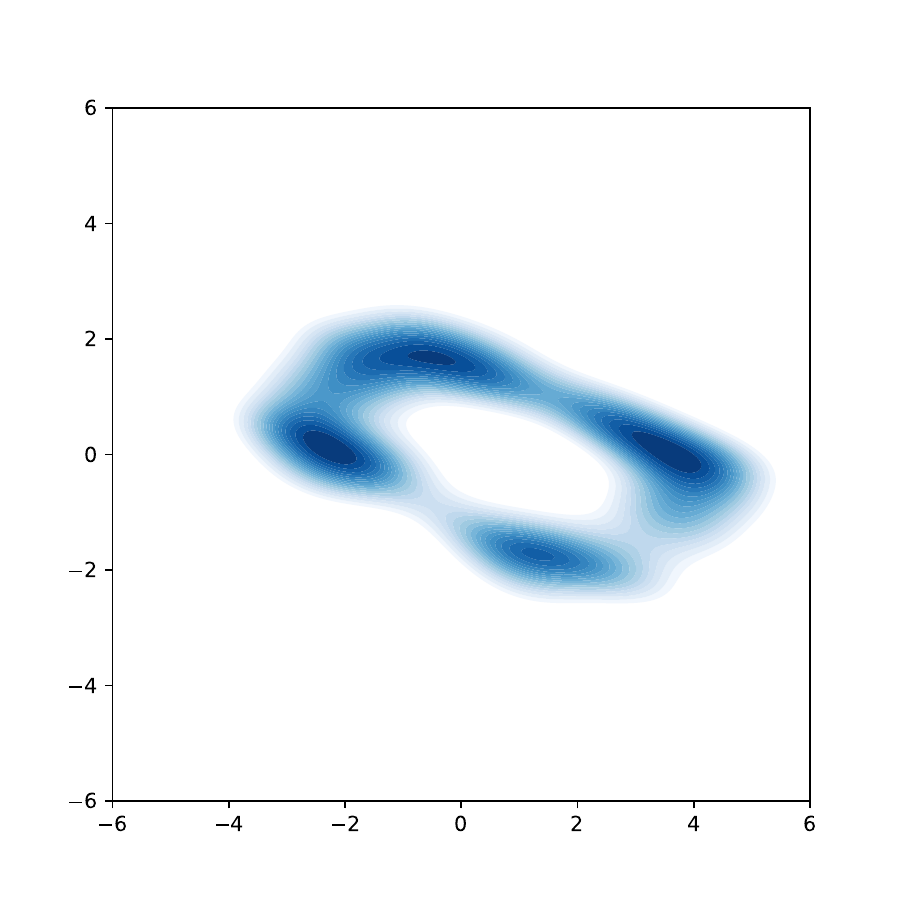}
   \end{subfigure}
   \begin{subfigure}{.19\linewidth}
      \includegraphics[width=\linewidth,height= .8\linewidth]{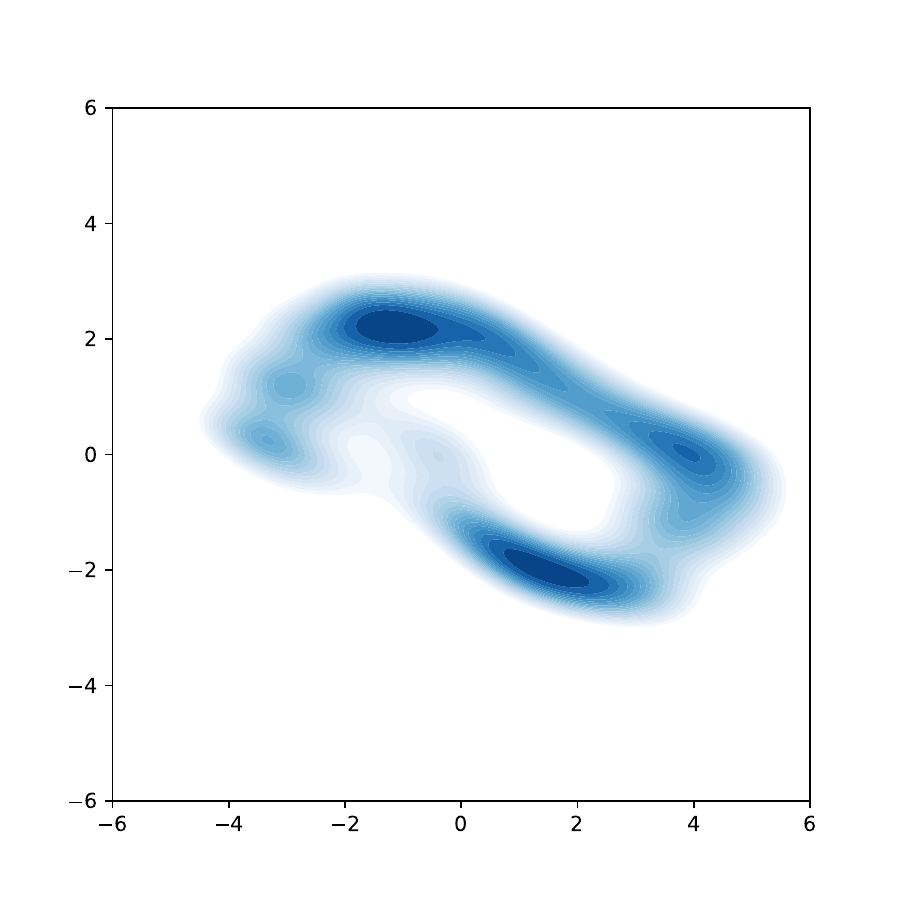}
   \end{subfigure}
   \begin{subfigure}{.19\linewidth}
      \includegraphics[width=\linewidth,height= .8\linewidth]{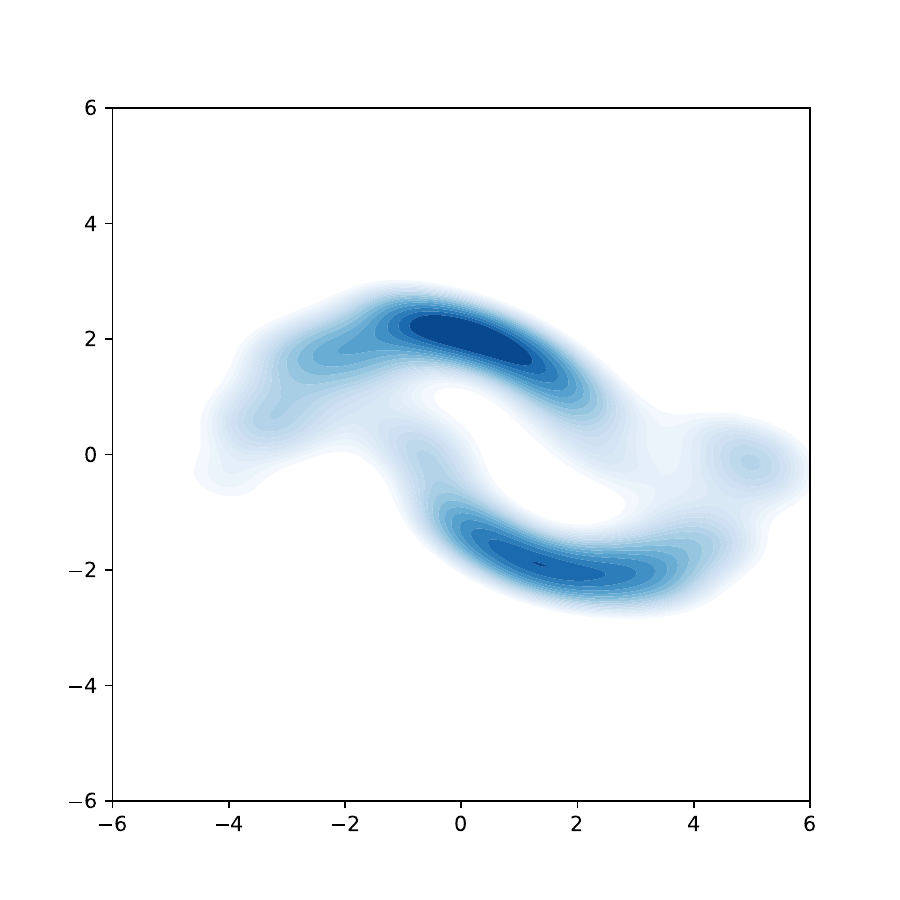}
   \end{subfigure}

   \makebox[0pt][r]{\makebox[15pt]{\raisebox{15pt}{\rotatebox[origin=c]{90}{FM}}}}%
   \begin{subfigure}{.19\linewidth}
      \includegraphics[width=\linewidth,height= .8\linewidth]{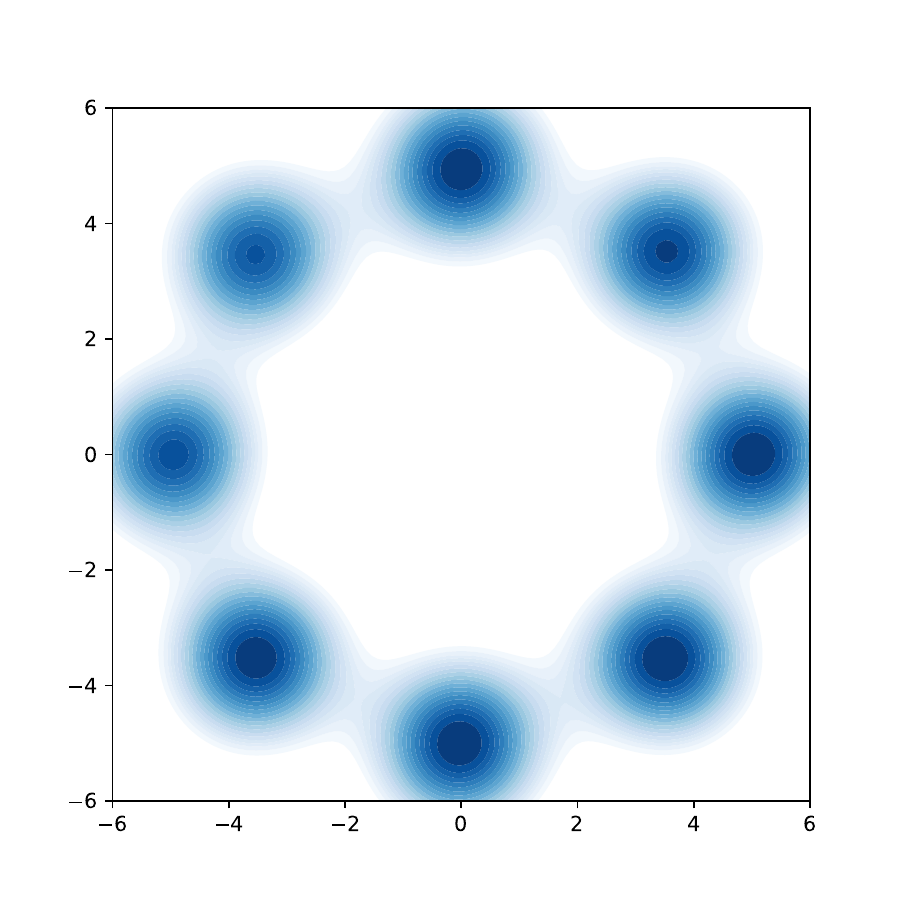}
   \end{subfigure}
   \begin{subfigure}{.19\linewidth}
         \includegraphics[width=\linewidth,height= .8\linewidth]{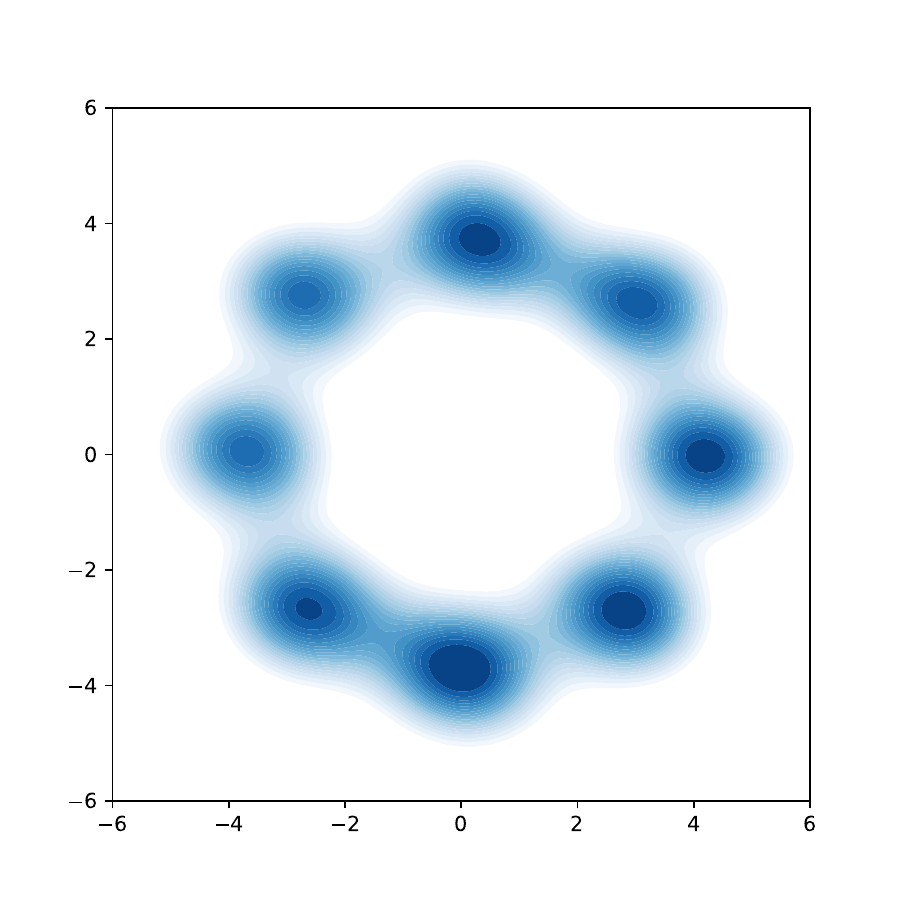}
   \end{subfigure}
   \begin{subfigure}{.19\linewidth}
         \includegraphics[width=\linewidth,height= .8\linewidth]{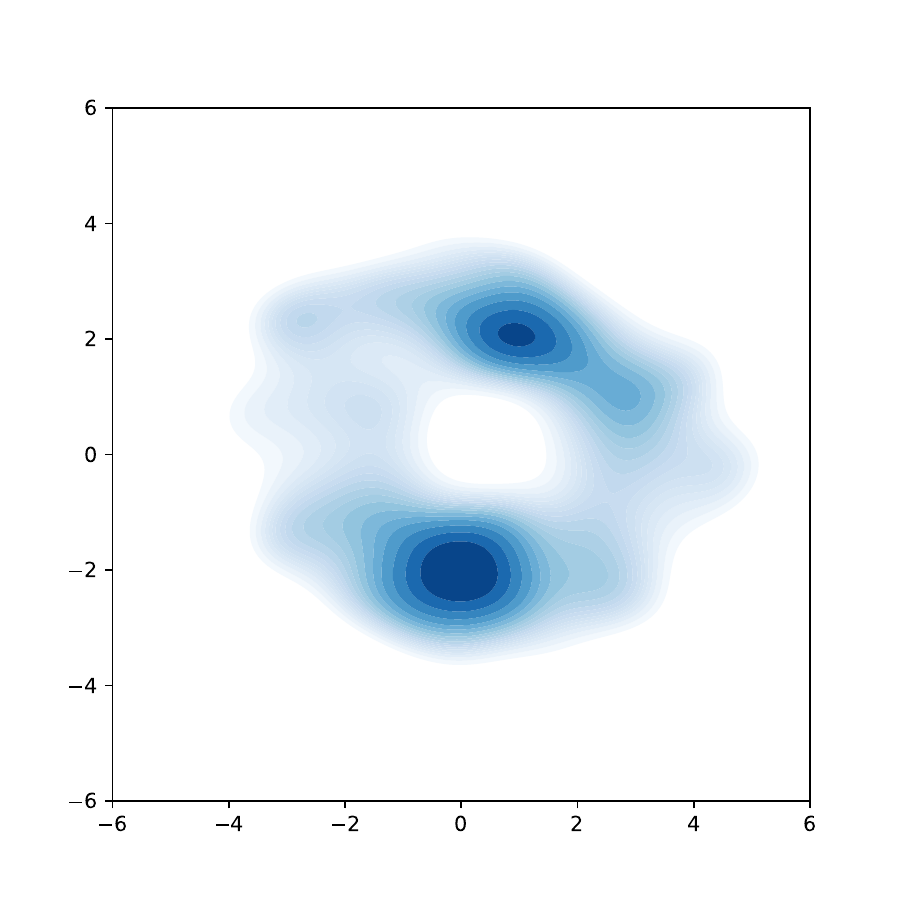}
   \end{subfigure}
   \begin{subfigure}{.19\linewidth}
         \includegraphics[width=\linewidth,height= .8\linewidth]{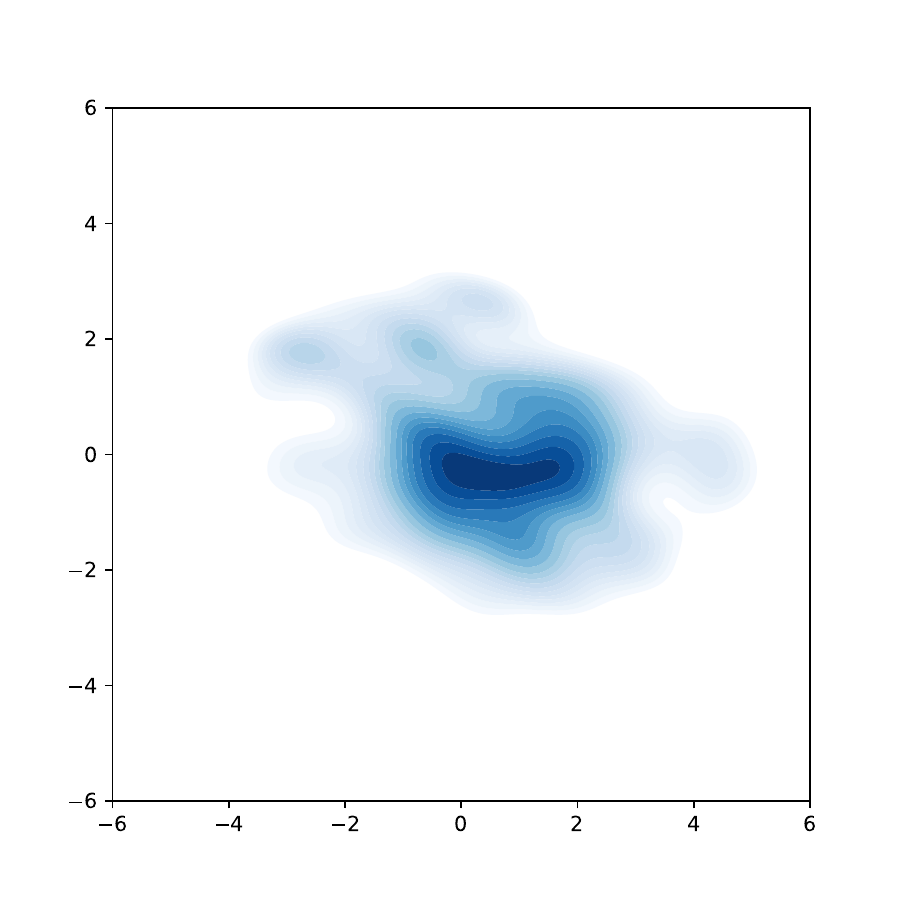}
   \end{subfigure}
   \begin{subfigure}{.19\linewidth}
         \includegraphics[width=\linewidth,height= .8\linewidth]{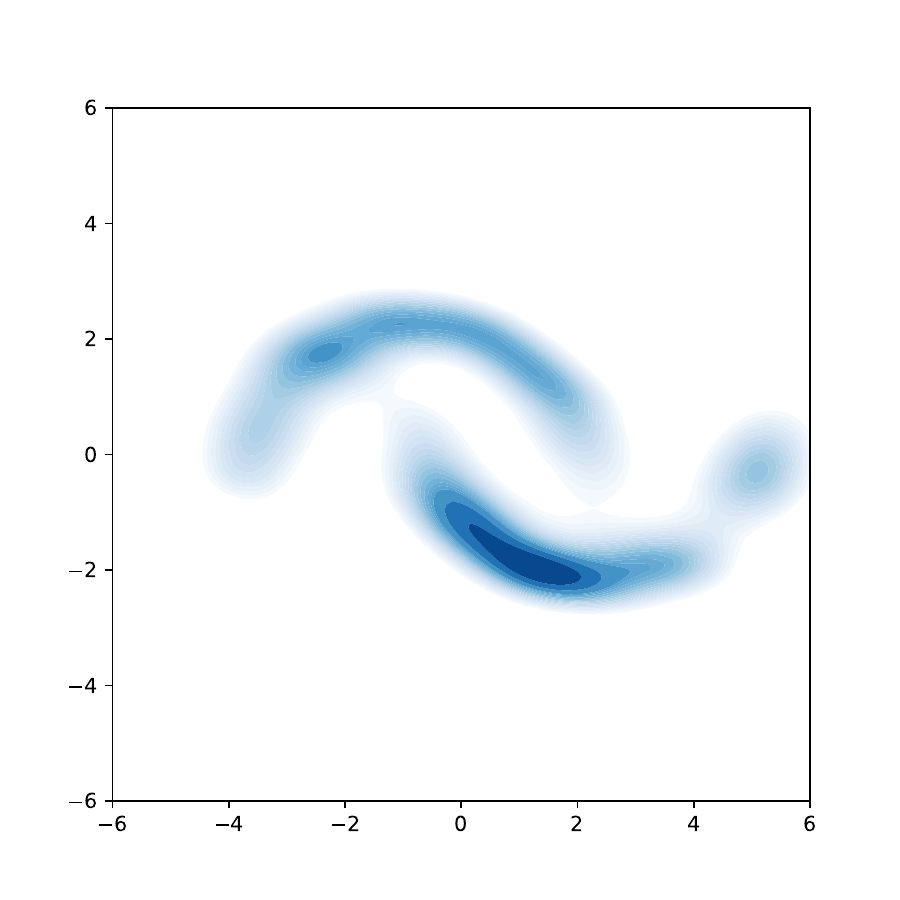}
   \end{subfigure}

   \makebox[0pt][r]{\makebox[15pt]{\raisebox{15pt}{\rotatebox[origin=c]{90}{SI}}}}
   \begin{subfigure}{.19\linewidth}
      \includegraphics[width=\linewidth,height= .9\linewidth]{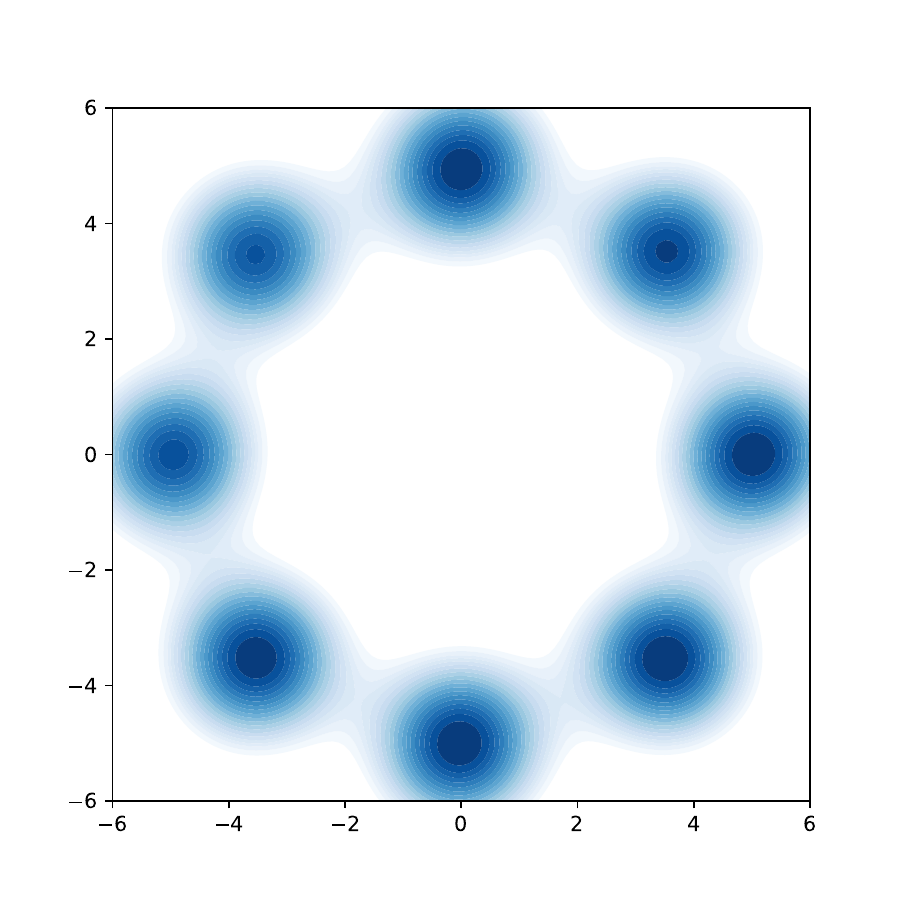}
   \end{subfigure}
   \begin{subfigure}{.19\linewidth}
      \includegraphics[width=\linewidth,height= .9\linewidth]{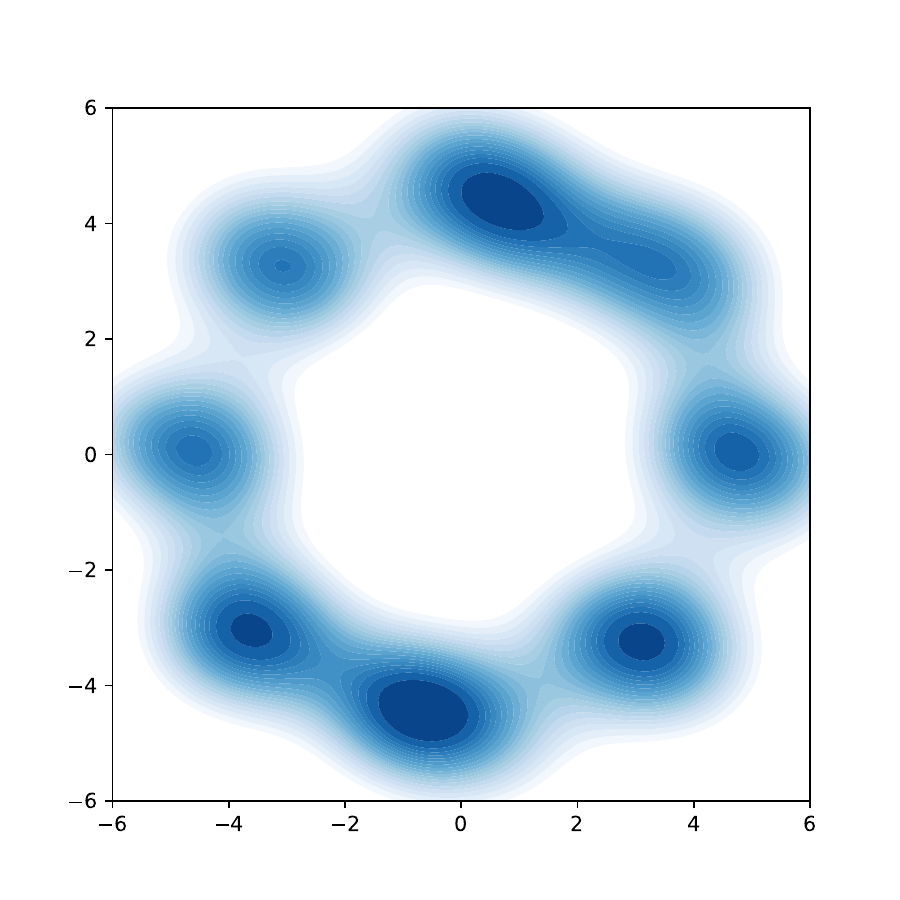}
   \end{subfigure}
   \begin{subfigure}{.19\linewidth}
      \includegraphics[width=\linewidth,height= .9\linewidth]{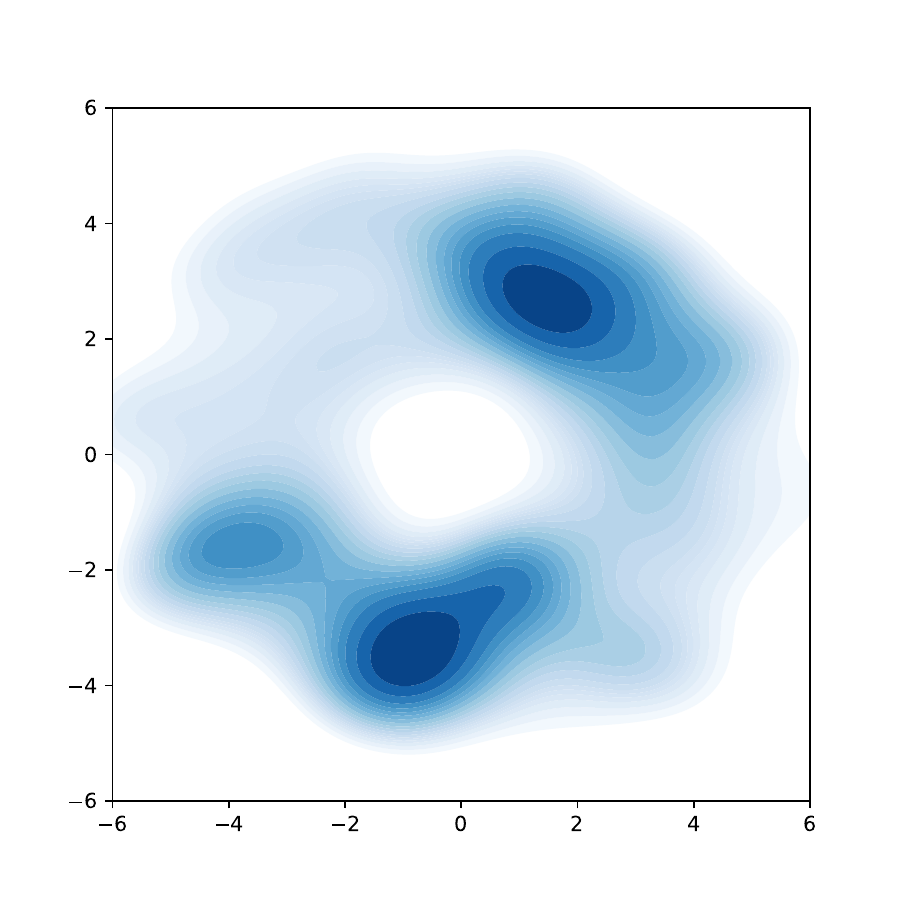}
   \end{subfigure}
   \begin{subfigure}{.19\linewidth}
      \includegraphics[width=\linewidth,height= .9\linewidth]{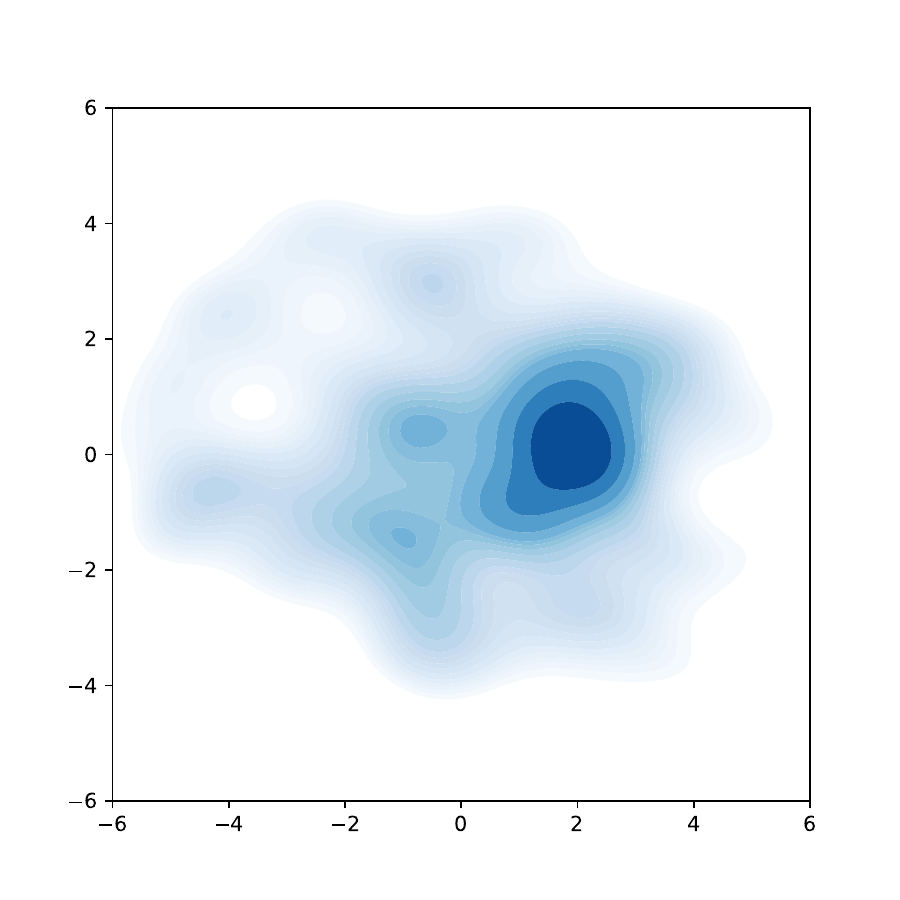}
   \end{subfigure}
   \begin{subfigure}{.19\linewidth}
      \includegraphics[width=\linewidth,height= .9\linewidth]{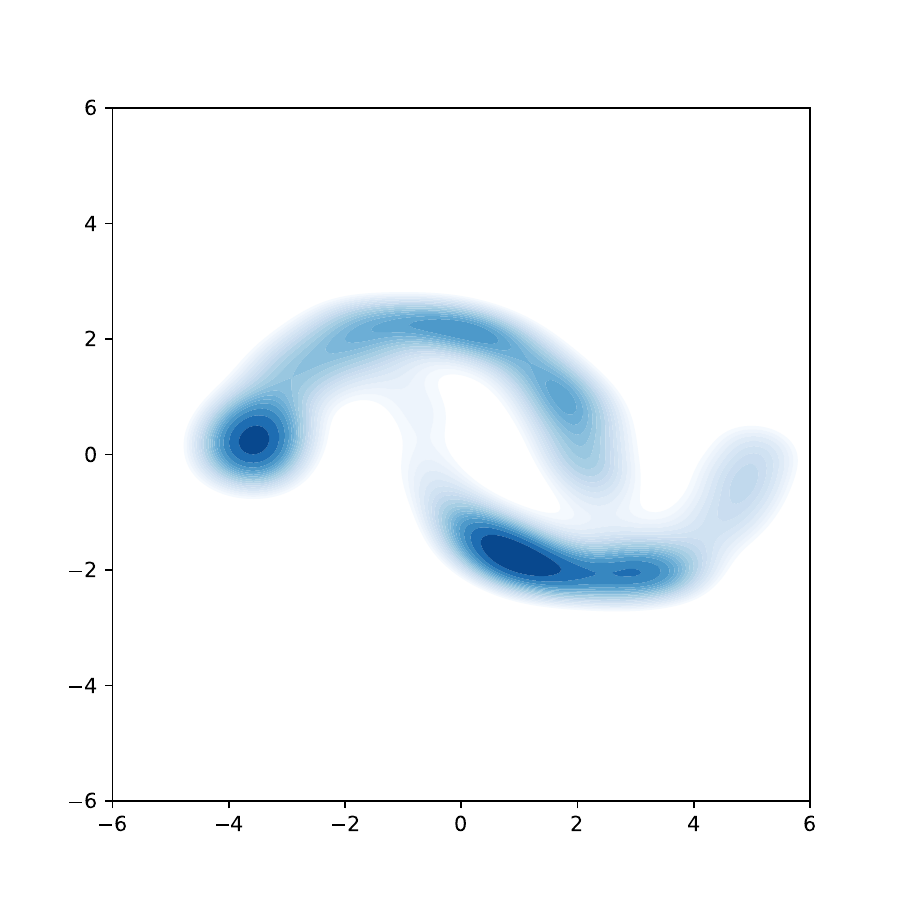}
   \end{subfigure}

   \caption{2-Wasserstein Distance of the generated process utilizing neural ODE-based diffusion models and NSGF. The FM/SI methods reduce noise roughly linearly, while NSGF quickly recovers the target structure and progressively optimizes the details in subsequent steps.}
   \label{KEDplot of 8gaussian to moons}
   \vspace{-5mm}
   \end{figure}

\section{Experiments}
We conduct an empirical investigation of the NSGF-based generative models (the standard NSGF and its two-phase variant NSGF++) across a range of experiments. Initially, we demonstrate how NSGF guides the evolution and convergence of particles from the initial distribution toward the target distribution in 2D simulation experiments. 
Subsequently, our attention turns to real-world image benchmarks, such as MNIST and CIFAR-10. To improve the efficiency in those high-dimensional tasks, we adopt the two-phase variant NSGF++ instead of the standard NSGF. Our method's adaptability to high-dimensional spaces is exemplified through experiments conducted on these datasets.
\subsection{2D simulation data}
We assess the performance of various generative modeling models in low dimensions. Specifically, we conduct a comparative analysis between our method, NSGF, and several neural ODE-based diffusion models, including Flow Matching (FM; \cite{lipman2023flow}), Rectified Flow (1,2,3-RF; \cite{liu2023flow}), Optimal Transport Condition Flow Matching (OT-CFM; \cite{tong2023improving,pooladian2023multisample}), Stochastic Interpolant (SI; \cite{albergo2023building}), and neural gradient-flow-based models such as JKO-Flow \cite{fan2022variational} and EPT \cite{gao2022deep}. Our evaluation involves learning 2D distributions adapted from \cite{grathwohl2018ffjord}, which include multiple modes. 
\par

Table \ref{2D} provides a comprehensive overview of our 2D experimental results, clearly illustrating the generalization capabilities of NSGF. Even when employing fewer steps. It is evident that neural gradient-flow-based models consistently outperform neural ODE-based diffusion models, particularly in low-step settings. This observation suggests that neural gradient-flow-based models generate more informative paths, enabling effective generation with a reduced number of steps. Furthermore, our results showcase the best performances among neural gradient-flow-based models, indicating that we have successfully introduced a lower error in approximating Wasserstein gradient flows. More complete details of the experiment can be found in the appendix \ref{appC}.
In the absence of specific additional assertions, we adopted Euler steps as the inference steps.

We present additional comparisons between neural ODE-based diffusion models and neural gradient-flow-based models, represented by NSGF and EPT, in Figure \ref{8gaussians-moons flow}, \ref{KEDplot of 8gaussian to moons}, which illustrates the flow at different steps from $0$ to $T$. Our observations reveal that the velocity field induced by NSGF exhibits notably high-speed values right from the outset. This is attributed to the fact that NSGF follows the steepest descent direction within the probability space. In contrast, neural ODE-based diffusion models, particularly those based on stochastic interpolation, do not follow the steepest descent path in 2D experiments. Even with the proposed rectified flow method by \cite{liu2023flow} to straighten the path, these methods still necessitate more steps to reach the desired outcome.



\subsection{Image benchmark data}
In this section, we illustrate the scalability of our algorithm to the high-dimensional setting by applying our methods to real image datasets. Notably, we leverage the two-phase variant (NSGF++) instead of the standard NSGF to enhance efficiency in high-dimensional spaces. It is worth mentioning that we achieve this improvement by constructing a significantly smaller training pool compared with the standard NSGF pool (10\% of the usual size), thus requiring only 5\% of the typical training duration.
We evaluate NSGF++ on MNIST and CIFAR10 to show our generating ability. Due to the limit of the space, we defer the generative images and comparison results of MNIST in appendix \ref{MNISTFID}.




\begin{table}[tb]
   \small
	\centering
   \resizebox{0.40\textwidth}{!}{
	\begin{tabular}{c|ccc}
		\toprule
		\multirow{2}*{Algorithm} & \multicolumn{3}{c}{CIFAR 10}  \\
		~  & IS($\uparrow$)  & FID($\downarrow$) & NFE($\downarrow$) \\
		\hline
      \textbf{NSGF++ (ours)} & \textbf{8.86}  & \textbf{5.55}  & \textbf{59}  \\
        EPT\shortcite{gao2022deep}      & /  & 46.63  &  10k  \\
        JKO-Flow\shortcite{fan2022variational} & 7.48 & 23.7 & $>$150  \\
        DGGF\shortcite{heng2022deep}    & /  & 28.12  &  110 \\
        \hline
        OT-CFM\shortcite{tong2023improving}  & /  & 11.14  & 100  \\        
        \hline
        FM\shortcite{lipman2023flow}    & /  & 6.35  & 142  \\
        RF\shortcite{liu2023flow} & \textbf{9.20}  & \textbf{4.88} & 100 \\
        SI\shortcite{albergo2023building}& /    & 10.27 & / \\
		\bottomrule
	\end{tabular}}
	\caption{Comparison of Neural Wasserstein gradient flow methods and Neural ODE-based diffusion models over CIFAR-10}
	\label{CIFAR10_FID}

\end{table}

We report sample quality using the standard Fr{\'e}chet Inception Distance (FID) \cite{heusel2017gans}, Inception Score (IS) \cite{salimans2016improved} and compute cost using the number of function evaluations (NFE). These are all standard metrics throughout the literature. 
\par

Table \ref{CIFAR10_FID} presents the results, including the Fréchet Inception Distance (FID), Inception Score (IS), and the number of function evaluations (NFE), comparing the empirical distribution generated by each algorithm with the target distribution. While our current implementation may not yet rival state-of-the-art methods, it demonstrates promising outcomes, particularly in terms of generating quality (FID), outperforming neural gradient-flow-based models (EPT, \cite{gao2022deep}; JKO-Flow, \cite{fan2022variational}; DGGF,(LSIF-$\mathcal{X}^2$) \cite{heng2022deep}) with fewer steps.
It's essential to emphasize that this work represents an initial exploration of this particular model category and has not undergone optimization using common training techniques found in recent diffusion-based approaches. Such techniques include the use of exponential moving averages, truncations, learning rate warm-ups, and similar strategies. Furthermore, it's worth noting that training neural gradient-flow-based models like NSGF in high-dimensional spaces can be challenging. Balancing the optimization of per-step information with the limitations of the neural network's expressive power presents an intriguing research avenue that warrants further investigation.

\begin{figure}[t]
   \centering
   \begin{subfigure}{0.85\linewidth}
       \includegraphics[width=\linewidth]{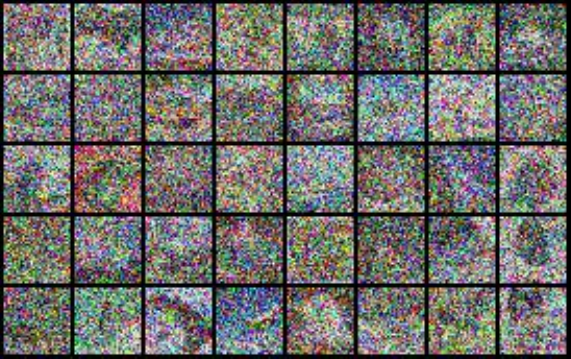}
   \end{subfigure}
   \begin{subfigure}{0.85\linewidth}
      \includegraphics[width=\linewidth]{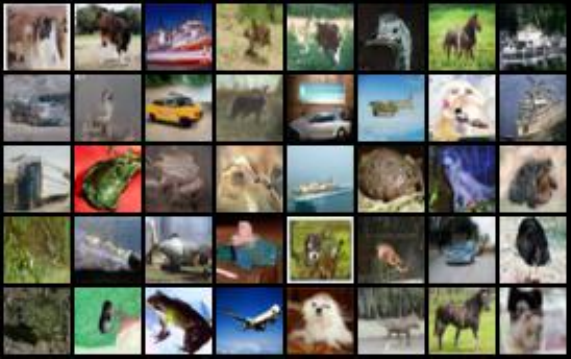}
  \end{subfigure}
   \caption{The inference result of our NSGF++ model. The first row shows the result after 5 NSGF steps and the second row shows the final results.}
   \label{p2}
\end{figure}
\section{Conclusion}
This paper delves into the realm of Wasserstein gradient flow w.r.t. the Sinkhorn divergence as an alternative to kernel methods. Our main investigation revolves around the Neural Sinkhorn Gradient Flow (NSGF) model, which introduces a parameterized velocity field that evolves over time in the Sinkhorn gradient flow. One noteworthy aspect of the NSGF is its efficient velocity field matching, which relies solely on samples from the target distribution for empirical approximations. The combination of rigorous theoretical foundations and empirical observations demonstrates that our approximations of the velocity field converge toward their true counterparts as the sample sizes grow.
To further enhance model efficiency on high-dimensional tasks, 
    a two-phase NSGF++ model is devised, which first follows the Sinkhorn flow to approach the image manifold quickly and then refine the samples along a simple straight flow.
Through extensive empirical experiments on well-known datasets like MNIST and CIFAR-10, we validate the effectiveness of the proposed methods.

\newpage
\begin{small}
    \bibliographystyle{named}
    \bibliography{ijcai24}

\begin{thebibliography}{}

\bibitem[\protect\citeauthoryear{Alaya \bgroup \em et al.\egroup
  }{2019}]{alaya2019screening}
Mokhtar~Z Alaya, Maxime Berar, Gilles Gasso, and Alain Rakotomamonjy.
\newblock Screening sinkhorn algorithm for regularized optimal transport.
\newblock {\em NeurIPS}, 32, 2019.

\bibitem[\protect\citeauthoryear{Albergo and
  Vanden-Eijnden}{2023}]{albergo2023building}
Michael~Samuel Albergo and Eric Vanden-Eijnden.
\newblock Building normalizing flows with stochastic interpolants.
\newblock In {\em The Eleventh ICLR}, 2023.

\bibitem[\protect\citeauthoryear{Alvarez-Melis \bgroup \em et al.\egroup
  }{2022}]{alvarez-melis2022optimizing}
David Alvarez-Melis, Yair Schiff, and Youssef Mroueh.
\newblock Optimizing functionals on the space of probabilities with input
  convex neural networks.
\newblock {\em Transactions on Machine Learning Research}, 2022.

\bibitem[\protect\citeauthoryear{Ambrosio \bgroup \em et al.\egroup
  }{2005}]{ambrosio2005gradient}
Luigi Ambrosio, Nicola Gigli, and Giuseppe Savar{\'e}.
\newblock {\em Gradient flows: in metric spaces and in the space of probability
  measures}.
\newblock Springer Science \& Business Media, 2005.

\bibitem[\protect\citeauthoryear{Amos \bgroup \em et al.\egroup
  }{2017}]{amos2017input}
Brandon Amos, Lei Xu, and J~Zico Kolter.
\newblock Input convex neural networks.
\newblock In {\em ICML}, pages 146--155. PMLR, 2017.

\bibitem[\protect\citeauthoryear{Ansari \bgroup \em et al.\egroup
  }{2021}]{ansari2021refining}
Abdul~Fatir Ansari, Ming~Liang Ang, and Harold Soh.
\newblock Refining deep generative models via discriminator gradient flow.
\newblock In {\em ICLR}, 2021.

\bibitem[\protect\citeauthoryear{Arbel \bgroup \em et al.\egroup
  }{2019}]{arbel2019maximum}
Michael Arbel, Anna Korba, Adil Salim, and Arthur Gretton.
\newblock Maximum mean discrepancy gradient flow.
\newblock {\em NeurIPS}, 32, 2019.

\bibitem[\protect\citeauthoryear{Bunne \bgroup \em et al.\egroup
  }{2022}]{bunne2022proximal}
Charlotte Bunne, Laetitia Papaxanthos, Andreas Krause, and Marco Cuturi.
\newblock Proximal optimal transport modeling of population dynamics.
\newblock In {\em International Conference on Artificial Intelligence and
  Statistics}, pages 6511--6528. PMLR, 2022.

\bibitem[\protect\citeauthoryear{Butcher}{1964}]{butcher1964implicit}
John~C Butcher.
\newblock Implicit runge-kutta processes.
\newblock {\em Mathematics of computation}, 18(85):50--64, 1964.

\bibitem[\protect\citeauthoryear{Choi \bgroup \em et al.\egroup
  }{2022}]{choi2022density}
Kristy Choi, Chenlin Meng, Yang Song, and Stefano Ermon.
\newblock Density ratio estimation via infinitesimal classification.
\newblock In {\em International Conference on Artificial Intelligence and
  Statistics}, pages 2552--2573. PMLR, 2022.

\bibitem[\protect\citeauthoryear{Cuturi}{2013}]{cuturi2013sinkhorn}
Marco Cuturi.
\newblock Sinkhorn distances: Lightspeed computation of optimal transport.
\newblock {\em NeurIPS}, 26, 2013.

\bibitem[\protect\citeauthoryear{Dai and Seljak}{2020}]{dai2020sliced}
Biwei Dai and Uros Seljak.
\newblock Sliced iterative normalizing flows.
\newblock {\em arXiv preprint arXiv:2007.00674}, 2020.

\bibitem[\protect\citeauthoryear{Damodaran \bgroup \em et al.\egroup
  }{2018}]{damodaran2018deepjdot}
Bharath~Bhushan Damodaran, Benjamin Kellenberger, R{\'e}mi Flamary, Devis Tuia,
  and Nicolas Courty.
\newblock Deepjdot: Deep joint distribution optimal transport for unsupervised
  domain adaptation.
\newblock In {\em Proceedings of the European conference on computer vision
  (ECCV)}, pages 447--463, 2018.

\bibitem[\protect\citeauthoryear{Das \bgroup \em et al.\egroup
  }{2023}]{das2023image}
Ayan Das, Stathi Fotiadis, Anil Batra, Farhang Nabiei, FengTing Liao, Sattar
  Vakili, Da-shan Shiu, and Alberto Bernacchia.
\newblock Image generation with shortest path diffusion.
\newblock {\em arXiv preprint arXiv:2306.00501}, 2023.

\bibitem[\protect\citeauthoryear{Deja \bgroup \em et al.\egroup
  }{2020}]{deja2020end}
Kamil Deja, Jan Dubi{\'n}ski, Piotr Nowak, Sandro Wenzel, Przemys{\l}aw Spurek,
  and Tomasz Trzcinski.
\newblock End-to-end sinkhorn autoencoder with noise generator.
\newblock {\em IEEE Access}, 9:7211--7219, 2020.

\bibitem[\protect\citeauthoryear{Dhariwal and
  Nichol}{2021}]{dhariwal2021diffusion}
Prafulla Dhariwal and Alexander Nichol.
\newblock Diffusion models beat gans on image synthesis.
\newblock {\em Advances in neural information processing systems},
  34:8780--8794, 2021.

\bibitem[\protect\citeauthoryear{Du \bgroup \em et al.\egroup
  }{2023}]{du2023nonparametric}
Chao Du, Tianbo Li, Tianyu Pang, YAN Shuicheng, and Min Lin.
\newblock Nonparametric generative modeling with conditional sliced-wasserstein
  flows.
\newblock 2023.

\bibitem[\protect\citeauthoryear{Fan \bgroup \em et al.\egroup
  }{2022}]{fan2022variational}
Jiaojiao Fan, Qinsheng Zhang, Amirhossein Taghvaei, and Yongxin Chen.
\newblock Variational wasserstein gradient flow.
\newblock In {\em ICML}, pages 6185--6215. PMLR, 2022.

\bibitem[\protect\citeauthoryear{Fatras \bgroup \em et al.\egroup
  }{2019}]{fatras2019learning}
Kilian Fatras, Younes Zine, R{\'e}mi Flamary, R{\'e}mi Gribonval, and Nicolas
  Courty.
\newblock Learning with minibatch wasserstein: asymptotic and gradient
  properties.
\newblock {\em arXiv preprint arXiv:1910.04091}, 2019.

\bibitem[\protect\citeauthoryear{Fatras \bgroup \em et al.\egroup
  }{2021a}]{fatras2021unbalanced}
Kilian Fatras, Thibault S{\'e}journ{\'e}, R{\'e}mi Flamary, and Nicolas Courty.
\newblock Unbalanced minibatch optimal transport; applications to domain
  adaptation.
\newblock In {\em ICML}, pages 3186--3197. PMLR, 2021.

\bibitem[\protect\citeauthoryear{Fatras \bgroup \em et al.\egroup
  }{2021b}]{fatras2021minibatch}
Kilian Fatras, Younes Zine, Szymon Majewski, R{\'e}mi Flamary, R{\'e}mi
  Gribonval, and Nicolas Courty.
\newblock Minibatch optimal transport distances; analysis and applications.
\newblock {\em arXiv preprint arXiv:2101.01792}, 2021.

\bibitem[\protect\citeauthoryear{Feydy \bgroup \em et al.\egroup
  }{2019}]{feydy2019interpolating}
Jean Feydy, Thibault S{\'e}journ{\'e}, Fran{\c{c}}ois-Xavier Vialard, Shun-ichi
  Amari, Alain Trouv{\'e}, and Gabriel Peyr{\'e}.
\newblock Interpolating between optimal transport and mmd using sinkhorn
  divergences.
\newblock In {\em The 22nd International Conference on Artificial Intelligence
  and Statistics}, pages 2681--2690. PMLR, 2019.

\bibitem[\protect\citeauthoryear{Folland}{1999}]{folland1999real}
Gerald~B Folland.
\newblock {\em Real analysis: modern techniques and their applications},
  volume~40.
\newblock John Wiley \& Sons, 1999.

\bibitem[\protect\citeauthoryear{Gao \bgroup \em et al.\egroup
  }{2019}]{gao2019deep}
Yuan Gao, Yuling Jiao, Yang Wang, Yao Wang, Can Yang, and Shunkang Zhang.
\newblock Deep generative learning via variational gradient flow.
\newblock In {\em ICML}, pages 2093--2101. PMLR, 2019.

\bibitem[\protect\citeauthoryear{Gao \bgroup \em et al.\egroup
  }{2022}]{gao2022deep}
Yuan Gao, Jian Huang, Yuling Jiao, Jin Liu, Xiliang Lu, and Zhijian Yang.
\newblock Deep generative learning via euler particle transport.
\newblock In {\em Mathematical and Scientific Machine Learning}, pages
  336--368. PMLR, 2022.

\bibitem[\protect\citeauthoryear{Genevay \bgroup \em et al.\egroup
  }{2016}]{genevay2016stochastic}
Aude Genevay, Marco Cuturi, Gabriel Peyr{\'e}, and Francis Bach.
\newblock Stochastic optimization for large-scale optimal transport.
\newblock {\em NeurIPS}, 29, 2016.

\bibitem[\protect\citeauthoryear{Genevay \bgroup \em et al.\egroup
  }{2018}]{genevay2018learning}
Aude Genevay, Gabriel Peyr{\'e}, and Marco Cuturi.
\newblock Learning generative models with sinkhorn divergences.
\newblock In {\em International Conference on Artificial Intelligence and
  Statistics}, pages 1608--1617. PMLR, 2018.

\bibitem[\protect\citeauthoryear{Goodfellow \bgroup \em et al.\egroup
  }{2014}]{goodfellow2014generative}
Ian Goodfellow, Jean Pouget-Abadie, Mehdi Mirza, Bing Xu, David Warde-Farley,
  Sherjil Ozair, Aaron Courville, and Yoshua Bengio.
\newblock Generative adversarial nets.
\newblock {\em NeurIPS}, 27, 2014.

\bibitem[\protect\citeauthoryear{Grathwohl \bgroup \em et al.\egroup
  }{2018}]{grathwohl2018ffjord}
Will Grathwohl, Ricky~TQ Chen, Jesse Bettencourt, Ilya Sutskever, and David
  Duvenaud.
\newblock Ffjord: Free-form continuous dynamics for scalable reversible
  generative models.
\newblock {\em arXiv preprint arXiv:1810.01367}, 2018.

\bibitem[\protect\citeauthoryear{Heng \bgroup \em et al.\egroup
  }{2022}]{heng2022deep}
Alvin Heng, Abdul~Fatir Ansari, and Harold Soh.
\newblock Deep generative wasserstein gradient flows.
\newblock 2022.

\bibitem[\protect\citeauthoryear{Heusel \bgroup \em et al.\egroup
  }{2017}]{heusel2017gans}
Martin Heusel, Hubert Ramsauer, Thomas Unterthiner, Bernhard Nessler, and Sepp
  Hochreiter.
\newblock Gans trained by a two time-scale update rule converge to a local nash
  equilibrium.
\newblock {\em NeurIPS}, 30, 2017.

\bibitem[\protect\citeauthoryear{Ho \bgroup \em et al.\egroup
  }{2020}]{ho2020denoising}
Jonathan Ho, Ajay Jain, and Pieter Abbeel.
\newblock Denoising diffusion probabilistic models.
\newblock {\em NeurIPS}, 33:6840--6851, 2020.

\bibitem[\protect\citeauthoryear{Hsieh \bgroup \em et al.\egroup
  }{2021}]{hsieh2021limits}
Ya-Ping Hsieh, Panayotis Mertikopoulos, and Volkan Cevher.
\newblock The limits of min-max optimization algorithms: Convergence to
  spurious non-critical sets.
\newblock In {\em ICML}, pages 4337--4348. PMLR, 2021.

\bibitem[\protect\citeauthoryear{Jordan \bgroup \em et al.\egroup
  }{1998}]{jordan1998variational}
Richard Jordan, David Kinderlehrer, and Felix Otto.
\newblock The variational formulation of the fokker--planck equation.
\newblock {\em SIAM journal on mathematical analysis}, 29(1):1--17, 1998.

\bibitem[\protect\citeauthoryear{Kalantari}{2007}]{Kalantari2007}
Iraj Kalantari.
\newblock {\em Induction over the Continuum}, pages 145--154.
\newblock Springer Netherlands, Dordrecht, 2007.

\bibitem[\protect\citeauthoryear{Komatsu \bgroup \em et al.\egroup
  }{2021}]{komatsu2021multi}
Tatsuya Komatsu, Tomoko Matsui, and Junbin Gao.
\newblock Multi-source domain adaptation with sinkhorn barycenter.
\newblock In {\em 2021 29th European Signal Processing Conference (EUSIPCO)},
  pages 1371--1375. IEEE, 2021.

\bibitem[\protect\citeauthoryear{Korotin \bgroup \em et al.\egroup
  }{2021}]{korotin2021neural}
Alexander Korotin, Lingxiao Li, Aude Genevay, Justin~M Solomon, Alexander
  Filippov, and Evgeny Burnaev.
\newblock Do neural optimal transport solvers work? a continuous wasserstein-2
  benchmark.
\newblock {\em NeurIPS}, 34:14593--14605, 2021.

\bibitem[\protect\citeauthoryear{Li \bgroup \em et al.\egroup
  }{2017}]{li2017mmd}
Chun-Liang Li, Wei-Cheng Chang, Yu~Cheng, Yiming Yang, and Barnab{\'a}s
  P{\'o}czos.
\newblock Mmd gan: Towards deeper understanding of moment matching network.
\newblock {\em NeurIPS}, 30, 2017.

\bibitem[\protect\citeauthoryear{Lipman \bgroup \em et al.\egroup
  }{2023}]{lipman2023flow}
Yaron Lipman, Ricky T.~Q. Chen, Heli Ben-Hamu, Maximilian Nickel, and Matthew
  Le.
\newblock Flow matching for generative modeling.
\newblock In {\em The Eleventh ICLR}, 2023.

\bibitem[\protect\citeauthoryear{Liu \bgroup \em et al.\egroup
  }{2023}]{liu2023flow}
Xingchao Liu, Chengyue Gong, and qiang liu.
\newblock Flow straight and fast: Learning to generate and transfer data with
  rectified flow.
\newblock In {\em The Eleventh ICLR}, 2023.

\bibitem[\protect\citeauthoryear{Liutkus \bgroup \em et al.\egroup
  }{2019}]{liutkus2019sliced}
Antoine Liutkus, Umut Simsekli, Szymon Majewski, Alain Durmus, and
  Fabian-Robert St{\"o}ter.
\newblock Sliced-wasserstein flows: Nonparametric generative modeling via
  optimal transport and diffusions.
\newblock In {\em ICML}, pages 4104--4113. PMLR, 2019.

\bibitem[\protect\citeauthoryear{Luise \bgroup \em et al.\egroup
  }{2019}]{luise2019sinkhorn}
Giulia Luise, Saverio Salzo, Massimiliano Pontil, and Carlo Ciliberto.
\newblock Sinkhorn barycenters with free support via frank-wolfe algorithm.
\newblock {\em NeurIPS}, 32, 2019.

\bibitem[\protect\citeauthoryear{Mnih \bgroup \em et al.\egroup
  }{2013}]{mnih2013playing}
Volodymyr Mnih, Koray Kavukcuoglu, David Silver, Alex Graves, Ioannis
  Antonoglou, Daan Wierstra, and Martin Riedmiller.
\newblock Playing atari with deep reinforcement learning.
\newblock {\em arXiv preprint arXiv:1312.5602}, 2013.

\bibitem[\protect\citeauthoryear{Mokrov \bgroup \em et al.\egroup
  }{2021}]{mokrov2021large}
Petr Mokrov, Alexander Korotin, Lingxiao Li, Aude Genevay, Justin~M Solomon,
  and Evgeny Burnaev.
\newblock Large-scale wasserstein gradient flows.
\newblock {\em NeurIPS}, 34:15243--15256, 2021.

\bibitem[\protect\citeauthoryear{Mroueh and
  Rigotti}{2020}]{mroueh2020unbalanced}
Youssef Mroueh and Mattia Rigotti.
\newblock Unbalanced sobolev descent.
\newblock {\em NeurIPS}, 33:17034--17043, 2020.

\bibitem[\protect\citeauthoryear{Mroueh \bgroup \em et al.\egroup
  }{2019}]{mroueh2019sobolev}
Youssef Mroueh, Tom Sercu, and Anant Raj.
\newblock Sobolev descent.
\newblock In {\em The 22nd International Conference on Artificial Intelligence
  and Statistics}, pages 2976--2985. PMLR, 2019.

\bibitem[\protect\citeauthoryear{Pai \bgroup \em et al.\egroup
  }{2021}]{pai2021fast}
Gautam Pai, Jing Ren, Simone Melzi, Peter Wonka, and Maks Ovsjanikov.
\newblock Fast sinkhorn filters: Using matrix scaling for non-rigid shape
  correspondence with functional maps.
\newblock In {\em Proceedings of the IEEE/CVF Conference on Computer Vision and
  Pattern Recognition}, pages 384--393, 2021.

\bibitem[\protect\citeauthoryear{Patrini \bgroup \em et al.\egroup
  }{2020}]{patrini2020sinkhorn}
Giorgio Patrini, Rianne Van~den Berg, Patrick Forre, Marcello Carioni, Samarth
  Bhargav, Max Welling, Tim Genewein, and Frank Nielsen.
\newblock Sinkhorn autoencoders.
\newblock In {\em Uncertainty in Artificial Intelligence}, pages 733--743.
  PMLR, 2020.

\bibitem[\protect\citeauthoryear{Peyr{\'e} \bgroup \em et al.\egroup
  }{2017}]{peyre2017computational}
Gabriel Peyr{\'e}, Marco Cuturi, et~al.
\newblock Computational optimal transport.
\newblock {\em Center for Research in Economics and Statistics Working Papers},
  (2017-86), 2017.

\bibitem[\protect\citeauthoryear{Platen and
  Bruti-Liberati}{2010}]{platen2010numerical}
Eckhard Platen and Nicola Bruti-Liberati.
\newblock {\em Numerical solution of stochastic differential equations with
  jumps in finance}, volume~64.
\newblock Springer Science \& Business Media, 2010.

\bibitem[\protect\citeauthoryear{Pooladian \bgroup \em et al.\egroup
  }{2023}]{pooladian2023multisample}
Aram-Alexandre Pooladian, Heli Ben-Hamu, Carles Domingo-Enrich, Brandon Amos,
  Yaron Lipman, and Ricky Chen.
\newblock Multisample flow matching: Straightening flows with minibatch
  couplings.
\newblock {\em arXiv preprint arXiv:2304.14772}, 2023.

\bibitem[\protect\citeauthoryear{Salimans \bgroup \em et al.\egroup
  }{2016}]{salimans2016improved}
Tim Salimans, Ian Goodfellow, Wojciech Zaremba, Vicki Cheung, Alec Radford, and
  Xi~Chen.
\newblock Improved techniques for training gans.
\newblock {\em NeurIPS}, 29, 2016.

\bibitem[\protect\citeauthoryear{Santambrogio}{2015}]{santambrogio2015optimal}
Filippo Santambrogio.
\newblock Optimal transport for applied mathematicians.
\newblock {\em Birk{\"a}user, NY}, 55(58-63):94, 2015.

\bibitem[\protect\citeauthoryear{Santambrogio}{2017}]{santambrogio2017euclidean}
Filippo Santambrogio.
\newblock $\{$Euclidean, metric, and Wasserstein$\}$ gradient flows: an
  overview.
\newblock {\em Bulletin of Mathematical Sciences}, 7:87--154, 2017.

\bibitem[\protect\citeauthoryear{Shaul \bgroup \em et al.\egroup
  }{2023}]{shaul2023kinetic}
Neta Shaul, Ricky~TQ Chen, Maximilian Nickel, Matthew Le, and Yaron Lipman.
\newblock On kinetic optimal probability paths for generative models.
\newblock In {\em ICML}, pages 30883--30907. PMLR, 2023.

\bibitem[\protect\citeauthoryear{Shen \bgroup \em et al.\egroup
  }{2020}]{shen2020sinkhorn}
Zebang Shen, Zhenfu Wang, Alejandro Ribeiro, and Hamed Hassani.
\newblock Sinkhorn barycenter via functional gradient descent.
\newblock {\em NeurIPS}, 33:986--996, 2020.

\bibitem[\protect\citeauthoryear{Silver \bgroup \em et al.\egroup
  }{2016}]{silver2016mastering}
David Silver, Aja Huang, Chris~J Maddison, Arthur Guez, Laurent Sifre, George
  Van Den~Driessche, Julian Schrittwieser, Ioannis Antonoglou, Veda
  Panneershelvam, Marc Lanctot, et~al.
\newblock Mastering the game of go with deep neural networks and tree search.
\newblock {\em nature}, 529(7587):484--489, 2016.

\bibitem[\protect\citeauthoryear{Song and Ermon}{2019}]{song2019generative}
Yang Song and Stefano Ermon.
\newblock Generative modeling by estimating gradients of the data distribution.
\newblock {\em NeurIPS}, 32, 2019.

\bibitem[\protect\citeauthoryear{Song \bgroup \em et al.\egroup
  }{2021}]{song2021scorebased}
Yang Song, Jascha Sohl-Dickstein, Diederik~P Kingma, Abhishek Kumar, Stefano
  Ermon, and Ben Poole.
\newblock Score-based generative modeling through stochastic differential
  equations.
\newblock In {\em ICLR}, 2021.

\bibitem[\protect\citeauthoryear{S{\"u}li and
  Mayers}{2003}]{suli2003introduction}
Endre S{\"u}li and David~F Mayers.
\newblock {\em An introduction to numerical analysis}.
\newblock Cambridge university press, 2003.

\bibitem[\protect\citeauthoryear{Tong \bgroup \em et al.\egroup
  }{2023}]{tong2023improving}
Alexander Tong, Nikolay Malkin, Guillaume Huguet, Yanlei Zhang, Jarrid
  Rector-Brooks, Kilian FATRAS, Guy Wolf, and Yoshua Bengio.
\newblock Improving and generalizing flow-based generative models with
  minibatch optimal transport.
\newblock In {\em ICML Workshop on New Frontiers in Learning, Control, and
  Dynamical Systems}, 2023.

\bibitem[\protect\citeauthoryear{Villani and others}{2009}]{villani2009optimal}
C{\'e}dric Villani et~al.
\newblock {\em Optimal transport: old and new}, volume 338.
\newblock Springer, 2009.

\bibitem[\protect\citeauthoryear{Wang \bgroup \em et al.\egroup
  }{2018}]{wang2018improving}
Wei Wang, Yuan Sun, and Saman Halgamuge.
\newblock Improving mmd-gan training with repulsive loss function.
\newblock {\em arXiv preprint arXiv:1812.09916}, 2018.

\bibitem[\protect\citeauthoryear{Zhang and Katsoulakis}{2023}]{zhang2023mean}
Benjamin~J Zhang and Markos~A Katsoulakis.
\newblock A mean-field games laboratory for generative modeling.
\newblock {\em arXiv preprint arXiv:2304.13534}, 2023.

\bibitem[\protect\citeauthoryear{Zhang \bgroup \em et al.\egroup
  }{2021a}]{zhang2021dpvi}
Chao Zhang, Zhijian Li, Hui Qian, and Xin Du.
\newblock Dpvi: A dynamic-weight particle-based variational inference
  framework.
\newblock {\em arXiv preprint arXiv:2112.00945}, 2021.

\bibitem[\protect\citeauthoryear{Zhang \bgroup \em et al.\egroup
  }{2021b}]{zhang2021wasserstein}
Yufeng Zhang, Siyu Chen, Zhuoran Yang, Michael Jordan, and Zhaoran Wang.
\newblock Wasserstein flow meets replicator dynamics: A mean-field analysis of
  representation learning in actor-critic.
\newblock {\em NeurIPS}, 34:15993--16006, 2021.

\end{thebibliography}
\end{small}

\appendix
\section{Computation of $\gW_{\varepsilon}$-potentials}\label{appA}
The $\gW_{\varepsilon}$-potentials is the cornerstone to conduct NSGF. Hence, a key component of our method is to efficiently compute this quantity. \cite{genevay2016stochastic} provided an  efficient method when both $\mu$ and $\nu$ are discrete measures so that we can calculate $\gW_{\varepsilon}$-potential in terms of samples. In particular, when $\mu$ is discrete, f can be simply represented by a finite-dimensional vector since only its values on $supp(\mu)$ matter. 

To more clearly explain the relationship between the calculation of $\gW_{\varepsilon}$-protentials and the composition of our algorithm, we provide the following explanation: In practice, we actually calculate the $\gW_{\varepsilon}$-potentials for the empirical distribution of discrete minibatches and construct Sinkhorn WGF based on this. Therefore, in fact, the $\mu$ and $\nu$ in the subsequent text refer to $(\tilde{X}_i^t)_{i=1}^{n}$ and $(\tilde{Y}_i^t)_{i=1}^{n}$ in the Algorithm \ref{velocity field}.
We first introduce another property of the entropy-regularized optimal transport problem.
\begin{lemma}\label{Optimality of sinkhorn potentials}
   Define the Sinkhorn mapping: $\mathcal{A}: \mathcal{C}(\mathcal{X}) \times \mathcal{M}_1^{+}(\mathcal{X}) \rightarrow \mathcal{C}(\mathcal{X})$ 
   \begin{equation}
      \mathcal{A}(f, \mu)(\vy)=-\varepsilon \log \int_{\mathcal{X}} \exp ((f(\vx)-c(\vx, \vy)) / \varepsilon) \mathbf{d} \mu(\vx).
   \end{equation}
   The pair $(f_{\mu, \nu}, g_{\mu, \nu})$ are the $\gW_{\varepsilon}$-potentials of the entropy-regularized optimal transport problem \ref{entropy-regularized OT} if they satisfy:
   \begin{equation}
      \begin{aligned}
         f_{\mu, \nu} &=\mathcal{A}(g_{\mu, \nu}, \nu), \mu-\text { a.e. } \quad \text { and } \\
         \quad g_{\mu, \nu} & =\mathcal{A}(f_{\mu, \nu}, \mu), \nu-\text { a.e.},
      \end{aligned}
   \end{equation}
   or equivalently
   \begin{equation}
      \begin{aligned}
         \int_{\gX} h(\vx, \vy) \rvd \nu (\vy) &= 1 , \mu-\text { a.e. }, \\
         \int_{\gX} h(\vx, \vy) \rvd \mu (\vx) &= 1 , \nu-\text { a.e. },
      \end{aligned}
   \end{equation}
   where $h(\vx, \vy):= \exp{\frac{1}{\gamma}(f(\vx)+g(\vx)-c(\vx, \vy))}$.
\end{lemma}
To be more precise, by plugging in the optimality condition on $g_{\mu, \nu}$ in \ref{W-potentials}, the dual problem \ref{entropy-regularized OT} becomes:
\begin{equation}
   \mathrm{OT}_\varepsilon(\mu, \nu)=\max _{f \in \mathcal{C}}\langle f, \mu\rangle+\langle\mathcal{A}(f, \mu), \nu\rangle
\end{equation} 
Viewing the discrete measure $\mu$ as a weight vector $\vw_\mu$ on $supp(\mu)$, we have:
\begin{equation}
   \mathrm{OT}_\varepsilon(\mu, \nu)=\max _{\mathbf{f} \in \mathbb{R}^d}\left\{F(\mathbf{f}):=\mathbf{f}^{\top} \vw_\mu+\mathbb{E}_{y \sim \nu}[\mathcal{A}(\mathbf{f}, \mu)(y)]\right\},
\end{equation}
that is, we get a standard concave stochastic optimization problem, where the randomness of the problem comes from $\nu$ \cite{genevay2016stochastic}. Hence, the problem can be solved using stochastic gradient descent (SGD). 
In our methods, we can treat the computation of $\gW_{\varepsilon}$-potentials as a Blackbox.  
In practice, we use the efficient implementation of the Sinkhorn algorithm with GPU acceleration from the GeomLoss package \cite{feydy2019interpolating}.


\section{Theory of Sinkhorn Wasserstein gradient flow}
\begin{definition}(First variation of Functionals over Probability). \label{First variation of Functionals over Probability}
   Given a functional $\gF: \gP(\gX) \to \sR^+$, we shell perturb measure $\mu$ with a perturbation $\chi$ so that $\mu + t \chi$ belongs to $\gP(\gX)$ for small $t$ ($\int \mathrm{d}\chi = 0$). We treat $\gF(\mu)$, as a functional over probability in its second argument and compute its first variation as follows:
   \begin{equation}
	\resizebox{.98\linewidth}{!}{$
	\displaystyle
		\left.\frac{d}{d t} \gF\left(\mu+t \chi\right)\right|_{t=0}=\lim _{t \rightarrow 0} \frac{\gF\left(\mu+t \chi\right)-\gF\left(\mu\right)}{t}\\
		:=\int \frac{\delta \gF}{\delta \mu}\left(\mu\right) d \chi .$}
   \end{equation}

\end{definition}
\subsection{Proof of Theorem \ref{first variation of Sinkhorn divergence}}\label{Proof of Theorem 1}
\begin{proof}
According to definition \ref{First variation of Functionals over Probability}, given $\gF_{\varepsilon}(\cdot) = \gS_{\varepsilon}(\cdot, \mu^*)$ and $t$ in a neighborhood of $0$, we define $\mu_{t} = \mu + t\delta \mu $ 
\begin{equation*}
	\begin{aligned}
	\lim_{t \to 0} \frac{1}{t} \left( \gF_{\varepsilon}(\mu_t) - \gF_{\varepsilon}(\mu) \right)
	& =\underbrace{\lim_{t \to 0}\frac{1}{t} \left( \gW_{\varepsilon}(\mu_t, \mu^*) - \gW_{\varepsilon}(\mu, \mu^*) \right) }_{\Delta_t^{ \text{first part}}}\\
	& -\underbrace{\lim_{t \to 0}\frac{1}{2t}  \left( \gW_{\varepsilon}(\mu_t, \mu_t) - \gW_{\varepsilon}(\mu, \mu)\right) }_{\Delta_t^{ \text{second part}}}\\
	\end{aligned}	
\end{equation*}
We first analysis $\Delta_t^{ \text{first part}} :=  \lim_{t \to 0}\frac{1}{t} \left( \gW_{\varepsilon}(\mu_t, \mu^*) - \gW_{\varepsilon}(\mu, \mu^*) \right)$.
First, let us remark that $(f, g)$ is the a suboptimal pair of dual potentials $\gW_{\varepsilon, \mu^*}{(\mu)}$ for short. Recall \ref{dual form},
\begin{equation*}
	\resizebox{.98\linewidth}{!}{$
	\displaystyle
	\gW_{\varepsilon} \geq \langle \mu_t, f \rangle + \langle \mu^*, g \rangle - \varepsilon\left\langle\mu_t \otimes \mu^*, \exp \left(\frac{1}{\varepsilon}(f \oplus g-\mathrm{C})\right)-1\right\rangle,$}
\end{equation*}
and thus, since
\begin{equation*}
	\resizebox{.98\linewidth}{!}{$
	\displaystyle
	\gW_{\varepsilon} \geq \langle \mu, f \rangle + \langle \mu^*, g \rangle - \varepsilon\left\langle\mu \otimes \mu^*, \exp \left(\frac{1}{\varepsilon}(f \oplus g-\mathrm{C})\right)-1\right\rangle,$}
\end{equation*}
one has
\begin{equation*}
	\small
	\begin{aligned}
	\Delta_t^{ \text{first part}} & \geq \langle \delta \mu, f \rangle - \varepsilon\left\langle \delta\mu \otimes \mu^*, \exp \left(\frac{1}{\varepsilon}(f \oplus g-\mathrm{C})\right)\right\rangle + o(1) \\
	& \geq 	\langle \delta \mu, f-\varepsilon \rangle + o(1)
	\end{aligned}
\end{equation*}

Conversely, let us denote by $(f_t, g_t)$ the optimal pair of potentials for $\gW_{\varepsilon}(\mu_t, \mu^*)$ satisfying $g_t(x_o) = 0$ for some arbitrary anchor point $x_o \in \gX$. As $(f_t, g_t)$ are suboptimal potentials for $\gW_{\varepsilon}(\mu, \mu^*)$ we get that 
\begin{equation*}
	\resizebox{.98\linewidth}{!}{$
	\displaystyle
	\gW_{\varepsilon} \geq \langle \mu, f_t \rangle + \langle \mu^*, g_t \rangle - \varepsilon\left\langle\mu_t \otimes \mu^*, \exp \left(\frac{1}{\varepsilon}(f \oplus g-\mathrm{C})\right)-1\right\rangle,$}
\end{equation*}
and thus, since
\begin{equation*}
	\resizebox{.98\linewidth}{!}{$
	\displaystyle
	\gW_{\varepsilon} \geq \langle \mu_t, f_t \rangle + \langle \mu^*, g_t \rangle - \varepsilon\left\langle\mu_t \otimes \mu^*, \exp \left(\frac{1}{\varepsilon}(f_t \oplus g-\mathrm{C})\right)-1\right\rangle,$}
\end{equation*}
one has
\begin{equation*}
	\small
	\begin{aligned}
		\Delta_t^{ \text{first part}} & \geq \langle \delta \mu, f_t \rangle - \varepsilon\left\langle \delta\mu \otimes \mu^*_t, \exp \left(\frac{1}{\varepsilon}(f_t \oplus g-\mathrm{C})\right)\right\rangle + o(1) \\
		& \geq 	\langle \delta \mu, f_t-\varepsilon \rangle + o(1)
	\end{aligned}
\end{equation*}
Now, let us remark that as $t$ goes to $0$, $\mu + t \delta \mu \rightharpoonup \mu$. $f_t$ and $g_t$ converge uniformly towards $f$ and $g$ according to Proposition 13 \cite{feydy2019interpolating}. we get
\begin{equation*}
	\Delta_t^{ \text{first part}} = \langle \delta \mu, f \rangle 
\end{equation*}
Simular to analysis 
$$\Delta_t^{ \text{first part}} :=  lim_{t \to 0}\frac{1}{t} \left( \gW_{\varepsilon}(\mu_t, \mu^*) - \gW_{\varepsilon}(\mu, \mu^*) \right)$$ we define $$\Delta_t^{ \text{second part}} := lim_{t \to 0}\frac{1}{2t} \left( \gW_{\varepsilon}(\mu_t, \mu_t) - \gW_{\varepsilon}(\mu, \mu)\right)$$, we have:
\begin{equation*}
	\Delta_t^{ \text{second part}} = \langle \delta \mu, f^{\prime} \rangle
\end{equation*}
to be more clearly, we denote $f = f_{\mu, \mu^*}$ and $f^{\prime} = f_{\mu, \mu}$
thus,
\begin{equation*}
	\lim_{t \to 0} \frac{1}{t} \left( \gF_{\varepsilon}(\mu_t) - \gF_{\varepsilon}(\mu) \right) = \langle \delta \mu, f_{\mu, \mu^*} - f_{\mu, \mu} \rangle.
\end{equation*}
So the first variation of $\gF_{\varepsilon}$ is:
\begin{equation*}
	\frac{\delta \gF_{\varepsilon}}{\delta \mu} =  f_{\mu, \mu^*} - f_{\mu, \mu}.
\end{equation*}
\end{proof}

\subsection{Proof of Proposition \ref{Frechet derivative of Sinkhorn divergence}} \label{proof of proposition 2}
Following the lines of our proof in Theorem \ref{first variation of Sinkhorn divergence}, we give the following proof.
\begin{lemma}(Fr{\'e}chet derivative of entropy-regularized Wasserstein distance)\label{Frechet derivative of entropy-regularized Wasserstein distance}
   Let $\varepsilon > 0$. We shall fix in the following a measure $\mu^*$ and let $(f_{\mu, \mu^*}, g_{\mu, \mu^*})$ be the $\gW_{\varepsilon}$ potentials of $\gW_{\varepsilon}{(\mu, \mu^*)}$ according to lemma \ref{W-potentials}. Consider the infinitesimal transport $T(x) = x + \lambda \phi$. We have the Fr{\'e}chet derivative under this particular perturbation:
   \begin{equation}
	\begin{aligned}
		\frac{d }{d \lambda} \gW_\varepsilon(T_{\#}\mu, \mu^*)|_{\lambda = 0} 
		& = \lim _{\lambda \rightarrow 0} \frac{\gW_\varepsilon\left(T_{\#}\mu, \mu^*\right)-\gW_\varepsilon\left(\mu, \mu^*\right)}{\lambda} \\
		& = \int_{\gX} \nabla f_{\mu, \mu^*}(\vx) \phi(\vx) \rvd \mu(\vx) 
        .
	\end{aligned}  
   \end{equation}
\end{lemma}

\begin{proof}
 let $f = f_{\mu, \mu^*}$ and $g = g_{\mu, \mu^*}$ be the $\gW_{\varepsilon}$-potentials \ref{W-potentials} to $\gW_{\varepsilon}(\mu, \mu^*)$ for short. By \ref{dual form} and the optimality of $(f, g)$, we have follows:
 \begin{equation*}
	\gW_{\varepsilon}(\mu, \mu^*)= \langle f, \mu \rangle + \langle g, \mu^* \rangle.
 \end{equation*}
 However, $(f , g)$ are not necessarily the optimal dual variables for $\gW_{\varepsilon}(T_{\#}\mu, \mu^*)$, recall the lemma \ref{Optimality of sinkhorn potentials}:
 \begin{equation*}
	\gW_{\varepsilon}(T_{\#}\mu, \mu^*) \geq \langle f, T_{\#}\mu \rangle + \langle g, \mu^* \rangle - \varepsilon \langle h-1, T_{\#}\mu \otimes \mu^* \rangle,
 \end{equation*}
 where $\int_{\gX} h(\vx, \vy) \rvd \mu^* (\vy) = 1 $ and hence $\langle h-1, T_{\#}\mu \otimes \mu^* \rangle = 0$.
 Thus:
 \begin{equation*}
	\gW_{\varepsilon}(T_{\#}\mu, \mu^*) - \gW_{\varepsilon}(\mu, \mu^*) \geq \langle f, T_{\#}\mu - \mu \rangle.
 \end{equation*}
 Use the change-of-variables formula of the push-forward measure to obtain:
 \begin{equation*}
	\begin{aligned}
		\frac{1}{\lambda} \langle f, T_{\#}\mu-\mu \rangle &= \frac{1}{\lambda} \int_{\gX} ((f\circ T)(\vx) - f(\vx)) \rvd \mu(\vx) \\ 
		&= \int_{\gX} \nabla f(x+\lambda^{\prime}\phi(x))\phi(x) \rvd \mu(\vx),
	\end{aligned}
 \end{equation*}
 where $\lambda^{\prime}\in [0, \lambda]$ is from the mean value theorem.
 Here we assume $\nabla f$ is Lipschitz continuous follow Proposition 12 in \cite{feydy2019interpolating} and Lemma A.4 form \cite{shen2020sinkhorn}. We have:
 \begin{equation*}
	\lim_{\lambda \to 0} \frac{1}{\lambda} \langle f, T_{\#}\mu - \mu \rangle =  \int_{\gX} \nabla f(\vx) \phi(\vx) \rvd \mu(\vx).
 \end{equation*}
Hence:
\begin{equation*}
	\resizebox{.98\linewidth}{!}{$
	\displaystyle
	\lim_{\lambda \to 0} \frac{1}{\lambda} \left( \gW_{\varepsilon}(T_{\#}\mu, \mu^*) - \gW_{\varepsilon}(\mu, \mu^*) \right) \geq \int_{\gX} \nabla f(\vx) \phi(\vx) \rvd \mu(\vx).$}
\end{equation*}
Similarly, let $f^{\prime}$ and $g^{\prime}$ be the $\gW_{\varepsilon}$ potentials to $\gW_{\varepsilon}(T_{\#}\mu, \mu^*)$, we have:
\begin{equation*}
	\gW_{\varepsilon}(\mu, \mu^*) \geq \langle f^{\prime}, \mu \rangle + \langle g, \mu^* \rangle - \varepsilon \langle h-1, \mu \otimes \mu^* \rangle,
\end{equation*}
where $\int_{\gX} h(\vx, \vy) \rvd \mu^* (\vy) = 1 $ and hence $\langle h-1, \mu \otimes \mu^* \rangle = 0$. Thus:
\begin{equation*}
	\gW_{\varepsilon}(T_{\#}\mu, \mu^*) - \gW_{\varepsilon}(\mu, \mu^*) \leq \langle f^{\prime}, T_{\#}\mu - \mu \rangle.
\end{equation*}
Same as above, use the change-of-variables formula and the mean value theorem:
\begin{equation*}
	\frac{1}{\lambda} \langle f^{\prime}, T_{\#}\mu - \mu^* \rangle = \int_{\gX} \nabla f^{\prime}(x+\lambda^{\prime}\phi(x))\phi(x) \rvd \mu(\vx),
\end{equation*}
Thus:
\begin{equation*}
	\begin{aligned}
		\lim_{\lambda \to 0}\frac{1}{\lambda} (\gW_{\varepsilon}(T_{\#}\mu, \mu^*) - \gW_{\varepsilon}(\mu, \mu^*))  \leq \\ \int_{\gX}\lim_{\lambda \to 0} \nabla f^{\prime}(\vx+\lambda^{\prime}\phi(\vx))\phi(x) \rvd \mu(\vx) .
	\end{aligned}
\end{equation*}
Assume that $\nabla f^{\prime}$ is Lipschitz continuous and $f^{\prime}\to f$ as $\lambda \to 0$. Consequently we have $\lim_{\lambda \to 0}\nabla f^{\prime}(x+\lambda^{\prime}\phi(\vx))$ and hence:
\begin{equation*}
	\lim_{\lambda \to 0} \frac{1}{\lambda} \left( \gW_{\varepsilon}(T_{\#}\mu, \mu^*) - \gW_{\varepsilon}(\mu, \mu^*) \right) = \int_{\gX} \nabla f(\vx) \phi(\vx) \rvd \mu(\vx).
\end{equation*}

According to lemma \ref{Frechet derivative of entropy-regularized Wasserstein distance}, we have:
\begin{equation*}
	\small
	\begin{aligned}
		\left.\frac{d }{d \lambda} \gF_\varepsilon(T_{\#}\mu)\right|_{\lambda = 0} 
		&= \int_{\gX} \nabla f_{\mu, \mu^\ast}(\vx) \phi(\vx) \rvd \mu(\vx) \\ &-  \frac{1}{2} \cdot \int_{\gX} 2\nabla f_{\mu, \mu}(\vx) \phi(\vx) \rvd \mu(\vx) \\
		&=\int_{\gX} \nabla f_{\mu, \mu^\ast}(\vx)\phi(\vx)d\mu - \int_{\gX} \nabla f_{\mu, \mu}(\vx)\phi(\vx)\rvd\mu
		.
	\end{aligned}
\end{equation*}
\end{proof}
\subsection{Proof of theorem \ref{theorem2-1}}
\label{Proof of theorem 2}
\begin{proof}
First, we define $\Psi(\mu) = \int h \mathrm{d}\mu$ where $h:\mathbb{R}^d\to \mathbb{R}$ is an arbitrary bounded and continuous function and $\frac{\delta \Psi(\mu)}{\delta \mu }(\vx)$ denotes the first variation of functional $\Psi$ at $\mu$ satisfying: 
	\begin{align*}
		\int \frac{\delta \Psi(\mu)}{\delta \mu }(\vx)  \xi(\vx) d \vx = \lim_{\epsilon \to 0} \frac{\Psi(\mu + \epsilon \xi) - \Psi(\mu)}{\epsilon}
	\end{align*}
	for all signed measure $\int \xi(\vx) d \vx =0$. We also have the following:
	\begin{align*}
		\frac{\delta \Psi(\mu)}{\delta \mu }(\cdot) = \frac{\delta \int h \mathrm{d}\mu}{\delta \mu }(\cdot) = h(\cdot)
	\end{align*}

	Assume $\mu_t$ is a flow satisfies the following:
	\begin{align*}
		\partial_t \Psi[\mu_t] = (\mathcal{L} \Psi)[\mu_t],
	\end{align*}
	where, 
	\begin{align}\label{lpsi} 
		\mathcal{L} \Psi[\mu_t] = - \int \langle \nabla\frac{\delta\gF_{\varepsilon}(\mu_t)}{\delta\mu} (\vx), \nabla_\vx \frac{\delta \Psi(\mu_t)}{\delta \mu }(\vx)  \rangle \mu_t(\vx)\mathrm{d} \vx
	\end{align}
	 Notably, $\mu_t$ is a solution of equation \ref{theorem2-1}.
	
	Next, let $\tilde{\mu}_t^M$ be the distribution produced by the equation \ref{ODE} at time $t$. Under mild assumption of $\tilde{\mu}_0^M \rightharpoonup \mu_0$, we want to show that the mean-field limit of $\tilde{\mu}_t^M$ as $M \to \infty$ is $\mu_t$ by showing that $\lim_{M\to \infty} \Psi(\mu_t^M) = \Psi (\mu_t)$ \cite{folland1999real}.

	For the measure valued flow $\tilde{\mu}_t^M$ \eqref{ODE}, the infinitesimal  generator of $\Psi$ w.r.t. $ \tilde \mu_t^M$ is defined as follows: 
	\begin{align*}
		(\gL \Psi)[\tilde{\mu}_t^M] := \lim_{\epsilon\to 0^+} \frac{\Psi(\tilde{\mu}_{t+\epsilon}^M) - \Psi(\tilde{\mu}_t^M)}{\epsilon},
	\end{align*}
	According to the definition of first variation, it can be calculated that 
	\begin{equation*}
		\small
		\begin{aligned}
		(\gL \Psi)[\tilde{\mu}_t^M]  
		=&\lim_{\epsilon\to 0^+} \frac{\Psi[\sum_{i=1}^M \frac{1}{M} \delta_{\vx^i_{t+\epsilon}}] - \Psi(\sum_{i=1}^M \frac{1}{M} \delta_{\vx^i_t})}{\epsilon}
		\\
		=&\int \frac{\delta \Psi(\tilde{\mu}_t^M)}{\delta \mu }(\vx) \sum_{i=1}^M \frac{1}{M} \partial_t \rho(\vx^i_t)  d \vx
	\end{aligned}
	\end{equation*}
		 
	Then we adopt the \emph{Induction over the Continuum} to prove $\lim_{n\to \infty} \Psi(\tilde{\mu}_t^M) = \Psi (\mu_t)$ for all $t>0$. Here $t\in \mathbb{R}^+$ satisfy the requirement of well ordering and the existence of a greatest lower bound for non-empty subsets, so \emph{Induction over the Continuum} is reasonable \cite{Kalantari2007}.

	1. As for $t=0$, our assumption of $\tilde{\mu}_0^M \rightharpoonup \mu_0$ suffice.
	
	2. For the case of $t=t^*$, we first hypothesis that for $t<t^*$, $\tilde{\mu}_t^M \rightharpoonup \mu_t$ as $M \to \infty$. Then for $t<t^*$ we have:
	\begin{equation*}
		\small
	\begin{aligned}
		\lim_{M\to \infty}&(\gL \Psi)[\tilde{\mu}_t^M] \\
		=&
		\lim_{M\to \infty}\int \frac{\delta \Psi(\tilde{\mu}_t^M)}{\delta \mu }(\vx) \sum_{i=1}^M \frac{1}{M} \partial_t \rho(\vx^i_t)  d \vx
		\\
		=&
		-\lim_{M\to \infty}\int \langle \nabla \frac{\delta\gF_{\varepsilon}(\mu_t)}{\delta\mu}(\vx), \nabla_\vx \frac{\delta \Psi(\tilde{\mu}_t^M)}{\delta \mu }(\vx)  \rangle \tilde{\mu}_t^M(\vx)\mathrm{d} \vx
		\\
		=&
		-\int \langle \nabla \frac{\delta\gF_{\varepsilon}(\mu_t)}{\delta\mu}(\vx), \nabla_\vx \frac{\delta \Psi(\mu_t)}{\delta \mu }(\vx)  \rangle \mu_t(\vx)\mathrm{d} \vx.
	\end{aligned}
	\end{equation*}	
	Because $\lim_{M\to \infty} \Psi(\tilde{\mu}_0^M) = \Psi (\mu_0)$ at $t = 0$ and $\lim_{M\to \infty}(\partial_t \Psi)[\tilde{\mu}_t^M] = (\partial_t \Psi)[\mu_t]$ for all $t < t^*$, we have $\lim_{M\to \infty} \Psi(\tilde{\mu}_{t^*}^M) = \Psi (\mu_{t^*})$.

	Combining (1) and (2), we can reach to the conclusion that $\lim_{M\to \infty} \Psi(\mu_t^M) = \Psi (\mu_t)$ for all $t$.
	which indicates that $\tilde{\mu}_t^M \rightharpoonup \mu_t$ if $\tilde{\mu}_0^M \rightharpoonup \mu_0$.
	Since $\mu_t$ solves the partial differential equation \ref{PDE flow}, we conclude that the path of \eqref{ODE} starting from $\tilde{\mu}^M_0$ 
	weakly converges to a solution of the partial differential equation \eqref{PDE flow} starting from $\mu_0$ as $M\to\infty$.
	
\end{proof}

\subsection{Descending property}
\begin{proposition}\label{proof of Descending property}
	Consider the Sinkhorn gradient flow \ref{PDE flow}, the differentiation of $\gF_{\varepsilon}(\mu_t)$ with respect to the time $t$ 
		satisfies: 
		\begin{align}\label{wfr-descent}
			\frac{\mathrm{d}\gF_{\varepsilon}(\mu_t)}{\mathrm{d}t} = &-\int{\left\|\nabla\left(\frac{\delta\gF_{\varepsilon}(\mu_t)}{\delta\mu_t}\right) \right\|^2\mathrm{d}\mu_t}\leq 0
		\end{align}
\end{proposition}

\begin{proof}
		By substituting $\Psi(\cdot) = \gF_{\varepsilon}(\cdot)$ in \eqref{lpsi}, we directly reach to the above equality.
		
\end{proof}

\section{Minibatch Optimal Transport}
For large datasets, the computation and storage of the optimal transport plan can be challenging due to OT's cubic time and quadratic memory complexities relative to the number of samples \cite{cuturi2013sinkhorn,genevay2016stochastic,peyre2017computational}. The minibatch approximation offers a viable solution for enhancing calculation efficiency. Theoretical analysis of using the minibatch approximation for transportation plans is provided by \cite{fatras2019learning,fatras2021minibatch}. Although minibatch OT introduces some errors compared to the exact OT solution, its efficiency in computing approximate OT is clear, and it has seen successful applications in domains like domain adaptation \cite{damodaran2018deepjdot,fatras2021unbalanced} and generative modeling \cite{genevay2018learning}.

More recently, \cite{pooladian2023multisample,tong2023improving} introduced OT-CFM and empirically demonstrated that using minibatch approximation of optimal transport in flow matching methods \cite{liu2023flow,lipman2023flow} can straighten the flow's trajectory and yield more consistent samples. OT-CFM specifically focuses on minibatch initial and target samples, continuing to use random linear interpolation paths. In contrast, NSGF leverages minibatch $\mathcal{W}_{\varepsilon}$-potentials to construct Sinkhorn gradient flows in minibatches. Our method also involves performing velocity field matching on the flow's discretized form, marking a separate and innovative direction in the field.
\section{NSGF++}
We introduce a two-phase NSGF++ algorithm that initially employs the Sinkhorn gradient flow for rapid approximation to the image manifold, followed by sample refinement using a straightforward straight flow. The NSGF++ model comprises three key components:
\begin{itemize}
   \item An NSGF model trained for $T \le 5$ time steps.
   \item A phase-transition time predictor, denoted as $t_\phi: \mathcal{X} \to [0,1]$, which facilitates the transition from NSGF to NSF.
   \item A Neural Straight Flow (NSF) model, trained through velocity field matching on a linear interpolation straight flow $X_t \sim (1-t) P_0 + t P_1, t\in [0,1]$.
\end{itemize}
the detailed algorithm is outlined in \ref{NSGF++ training}.

\begin{algorithm*}[t]
   \resizebox{0.95\textwidth}{!}{\begin{minipage}{\linewidth}
   \SetKwInOut{KwIn}{Input}
   \SetKwInOut{KwOut}{Output}

   \KwIn{number of time steps $T$, batch size $n$, gradient flow step size $\eta>0$, empirical or samplable distribution $\mu_0$ and $\mu^*$, neural network parameters $\theta$, optimizer step size $\gamma > 0$}
   \tcc{NSGF model}
   \tcc{Build trajectory pool}
   \While{$\text{Building}$}{
      \tcc{Sample batches of size $n$ $i.i.d.$ from the datasets}
      $ \tilde{X}_i^0 \sim \mu_0,\quad \tilde{Y_i} \sim \mu^*, i=1,2, \cdots n $.\\
      \For{$t = 0, 1, \cdots T$}{
         $\text{calculate} f_{\tilde{\mu}_t, \tilde{\mu}_t}\left(\tilde{X}_i^t\right),  f_{\tilde{\mu}_t,\tilde{\mu}^*}\left(\tilde{X}_i^t\right)$ . \\
         $\hat{\vv}_{\mu_t}^{\gF_\epsilon}\left(\tilde{X}_i^t\right)=\nabla f_{\tilde{\mu}_t, \tilde{\mu}_t}\left(\tilde{X}_i^t\right) - \nabla f_{\tilde{\mu}_t,\tilde{\mu}^*}\left(\tilde{X}_i^t\right)$ .\\
         $\tilde{X}_i^{t+1}=\tilde{X}_i^t+\eta \hat{\vv}_t^{\gF_\epsilon}\left(\tilde{X}_i^t\right)$. \\
         $\text { store all} \left(\tilde{X}_i^t, \hat{\vv}_t^{\gF_\epsilon} \left(\tilde{X}_i^{t}\right)\right) \text{pair into the pool},\quad i=1,2, \cdots n .$
      }
   }
   \tcc{velocity field matching}
   \While{$\text{Not convergence}$}{
      $\text{from trajectory pool sample pair} \left(\tilde{X}_i^t, \hat{\vv}_t^{\gF_\epsilon} \left(\tilde{X}_i^{t}\right)\right) $.\\
      $\mathcal{L}(\theta)= \left\|\vv^\theta(\tilde{X}_i^t, t)-\hat{\vv}_{\mu_t}^{\gF_{\varepsilon}}\left(\tilde{X}_i^t \right)\right\|^2$, \\
      $\theta \leftarrow \theta-\gamma \nabla_{\theta} \mathcal{L}\left(\theta\right)$ .
   }

   \tcc{phase trainsition time predictor}
   \While{Training}{
      $ \tilde{X}_i^0 \sim \mu_0,\quad \tilde{Y_i} \sim \mu^*, i=1,2, \cdots n $.\\
      $t \sim \mathcal{U}(0,1).$ \\
      $X_t = \operatorname{t} \tilde{Y_i}+(1-t) \tilde{X}_i^0$ \\
      $\mathcal{L}(\phi) = E_{t\in \gU(0, 1), X_{t} \sim P_{t}} ||t - t_{\phi}(X_t)||^2.$ \\
      $\phi \leftarrow \phi-\gamma^{\prime} \nabla_{\phi} \mathcal{L}\left(\phi\right).$
   }
   \tcc{NSF model}
   \While{Training}{
      $ \tilde{X}_i^0 \sim \mu_0,\quad \tilde{Y_i} \sim \mu^*, i=1,2, \cdots n $.\\
      $t \sim \mathcal{U}(0,1).$ \\ 
      $X_t = \operatorname{t} \tilde{Y_i}+(1-t) \tilde{X}_i^0.$ \\ 
      $\mathcal{L}_{\mathrm{NSF}}(\delta) \leftarrow\left\|u_\delta(t, X_t)-\left(\tilde{Y_i}-\tilde{X}_i^0\right)\right\|^2.$ \\ 
      $\delta \leftarrow \delta - \gamma^{\prime\prime}\nabla_\delta \mathcal{L}_{\mathrm{NSF}}(\delta).$
   }
     
   \KwOut{$\vv_\theta$ parameterize the time-varying velocity field of NSGF, $t_\phi$ parameterize the phase trainsition time predictor, $\vu_\delta$ parameterize the NSF model}

\caption{\textbf{NSGF++ Training} }
\label{NSGF++ training}
\end{minipage}
}
\end{algorithm*} 

In the inference process of NSGF++, we initially apply the NSGF with fewer than 5 NFE, starting from $X_0 \sim P_0$, to obtain an intermediate output $\tilde X_T$. This output is then processed using the time predictor $t_\phi$. The final output is achieved by refining this intermediate result with the NSF model, starting from the state $t_\phi(\tilde X_t)$. The detailed algorithm is outlined in \ref{NSGF++ inference}.
\begin{algorithm}[t]
    \resizebox{0.47\textwidth}{!}{\begin{minipage}{\linewidth}
    \SetKwInOut{KwIn}{Input}
    \SetKwInOut{KwOut}{Output}
 
    \KwIn{number of NSGF time steps $T \leq 5$, NSGF++ inference step size $\eta$, NSGF velocity field $\vv^\theta$, phase trainsition time predictor $t_{\phi}$, NSF inference step size $\omega$, NSF model $\vu^\delta$, prior samples $ \tilde{X}_i^0 \sim \tilde{\mu}_0$, $ODEsolver(X, model, starttime, endtime, steps)$}
    
    \tcc{NSGF phase}
    \For{$t = 0, 1, \cdots T$}{
       $\tilde{X}_i^{t+1}=\tilde{X}_i^t+\eta \vv^\theta\left(\tilde{X}_i^t, t\right), i=1,2, \cdots n . $ 
    }
    
    \tcc{phase trainsition time predict}
    $\hat{t} = t_{\phi}(\tilde{X}_i^{T}).$\\

    \tcc{NSF refine phase}
    $\hat{T} = (1 - \hat{t}) / \omega.$ \\
    $\tilde{X} = ODEsolver(\tilde{X}_i^{T}, \vu_\delta,\hat{t}, 1, \hat{T}).$
      
    \KwOut{$\tilde{X}$ as the results.}
 
 \caption{\textbf{NSGF++ Inference} }
 \label{NSGF++ inference}
 \end{minipage}
 }
 \end{algorithm} 
\section{Experimrnts}\label{appC}
\subsection{2D simulated data}
For the 2D experiments, we closely follow \cite{tong2023improving} and the released code at https: //github.com/atong01/conditional-flow-matching (code released under MIT license), and use the same synthetic datasets and the 2-Wasserstein distance between the test set and samples simulated using NSGF as the evaluation metric. We use 1024 samples in the test set since we find the  We use a simple MLP with 3 hidden layers and 256 hidden units to parameterize the velocity matching networks. We use batch size 256 and 10/100 steps with a uniform schedule at sampling time. For both Nerual gradient-flow-based models and Nerual ODE-based Models, we train for 20000 steps in total. Note that FM cannot be used for the 8gaussians-moons task since it requires a Gaussian source, but we still conducted experiments with the algorithm and found competitive experimental results. We believe that this is because FM is essentially very close to 1-RF in its algorithmic design, and that the Gaussian source condition can be meaningfully relaxed in practice, as confirmed in \cite{tong2023improving}. The experiments are run using one 3090 GPU and take approximately less than 60 minutes (for both training and testing).

For the neural gradient-flow-based models, we solely implemented the EPT without the outer loop, as the outer loop can be likened to a GAN-like distillation approach \cite{goodfellow2014generative}. Notably, the original EPT \cite{gao2022deep} recommends iterating for $20,000$ rounds with an exceedingly small step size; however, to ensure a fair comparison, we employed the same number of steps as the other methods while adapting the step size accordingly. It's worth mentioning that for the JKO-Flow, we used the recommended parameter setting of $10$ steps, as suggested in \cite{fan2022variational}, but we also provide results for $100$ steps for comparative purposes. All the results for Neural Gradient flow-based models were trained and sampled following the standard procedures outlined in their respective papers. 

\subsection{Image benchmark data}
For the MNIST/CIFAR-10 experiments, we summarize the setup of our NSGF++ model here, where the exact parameter choices can be seen in the source code.
For the calculation of $\gW_{\varepsilon}$-potentials, we use the GeomLoss package \cite{feydy2019interpolating} with blur = $0.5$, $1.0$ or $2.0$, scaling = $0.80$, $0.85$ or $0.95$ depends on learning rate of Sinkhorn gradient flow. 
We find using an incremental lr scheme will improve training performance. More detailed experiments we will leave for future work.
We used the Adam optimizer with $\beta_1 = 0.9, \beta_2 = 0.999$ and no weight decay. 
Here we list different part of our NSGF++ model separately.
First, we use the UNet architecture from \cite{dhariwal2021diffusion}. For MNIST, we use channels = 32, depth = 1, channels multiple = [1, 2, 2], heads = 1, attention resolution = 16, dropout = 0.0.
For CIFAR-10, we use channels = 128, depth = 2, channels multiple = [1, 2, 2, 2], heads = 4, heads channels = 64, attention resolution = 16, dropout = 0.0. 
We use the same UNet architecture in our neural straight flow model.



For the phase transition time predictor, we employed an efficiently designed convolutional neural network (CNN) capable of achieving satisfactory results while optimizing training time. The CNN used in our study consists of a structured architecture featuring four convolutional layers with filter depths of 32, 64, 128, and 256. Each layer uses a 3x3 kernel, a stride of 1, and padding of 1, coupled with ReLU activation and 2x2 average pooling for effective feature downsampling. The network culminates in a fully connected layer that transforms the flattened features into a single value, further refined through a sigmoid activation function for regression tasks targeting time value outputs. This architecture is tailored for processing image inputs to predict continuous time values within a specific range.
The training parameters are as follows:
Batch size: $128$
Learning rate: $10^{-4}$.
For sampling, a 5-step Euler integration is applied in the NSGF phase on MNIST and CIFAR-10 datasets.
Training the phase transition time predictor is efficient and methodically streamlined. Utilizing a well-structured CNN as its backbone, the model reaches peak performance in merely 20 minutes, covering 40,000 iterations. This training efficiency is a significant advantage, especially for applications that demand rapid model adaptation.

For the MNIST/CIFAR-10 experiments, a considerable amount of storage space is required when establishing the trajectory pool during the first phase of the algorithm. For MNIST dataset, setting the batch size to 256 and saving 1500 batches 5-step minibatch Sinkhorn gradient flow trajectories requires less than 20GB of storage space and with CIFAR-10, setting the batch size to 128 and saving 2500 batches 5-step minibatch Sinkhorn gradient flow trajectories requires about 45GB of storage space. In situations where storage space is limited, we suggest dynamically adding and removing trajectories in the trajectory pool to meet the training requirements. Identifying a more effective trade-off between training time and storage space utilization is a direction for future improvement.

\begin{figure}[t]
   \centering
   \begin{subfigure}{0.3\linewidth}
       \includegraphics[width=\linewidth]{figures/8gaussians-moons/flows/NSGF.png}
       \caption{NSGF}
    \end{subfigure}
    \begin{subfigure}{.3\linewidth}
       \includegraphics[width=\linewidth]{figures/8gaussians-moons/flows/CFM.png}
       \caption{CFM}
    \end{subfigure}
    \begin{subfigure}{.3\linewidth}
       \includegraphics[width=\linewidth]{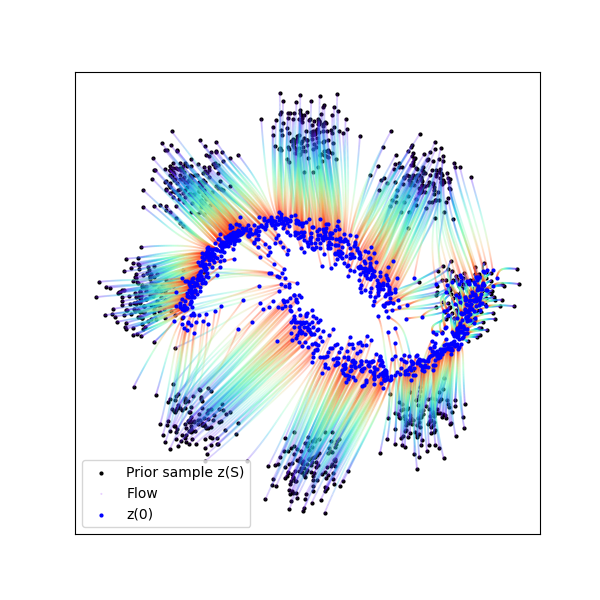}
       \caption{OT-CFM}
    \end{subfigure} 
    \begin{subfigure}{.3\linewidth}
       \includegraphics[width=\linewidth]{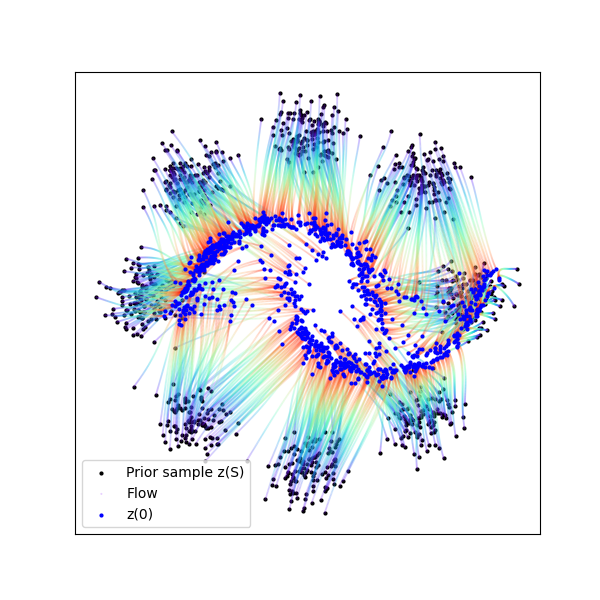}
       \caption{1-RF}
    \end{subfigure}
    \begin{subfigure}{.3\linewidth}
       \includegraphics[width=\linewidth]{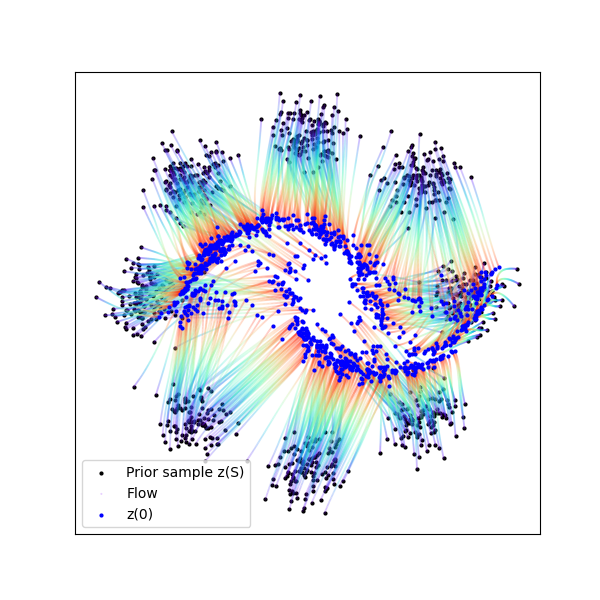}
       \caption{2-RF}
    \end{subfigure}
    \begin{subfigure}{.3\linewidth}
       \includegraphics[width=\linewidth]{figures/8gaussians-moons/flows/SI.png}
       \caption{SI}
    \end{subfigure}\qquad
   \vspace{-5mm}
   \caption{Visualization results for 2D generated paths. We show different methods that drive the particle from the prior distribution (black) to the target distribution (blue). The color change of the flow shows the different number of steps (from blue to red means from $0$ to $T$). We can see NSGF using fewer steps than OT-CFM}
   \label{2D stimulate data more results}
   \vspace{-5mm}
   \end{figure}

\subsection{Supplementary experimental results}
\subsubsection*{2D simulated data}
\begin{figure}[t]
   \captionsetup[subfigure]{labelformat=empty}
   \centering
   \tiny      
   \makebox[0pt][r]{\makebox[15pt]{\raisebox{15pt}{\rotatebox[origin=c]{90}{NSGF}}}}
   \begin{subfigure}{.19\linewidth}
   \includegraphics[width=\linewidth,height= .8\linewidth]{figures/8gaussians-moons/KDEplot/NSGF/1.pdf}
   \end{subfigure}
   \begin{subfigure}{.19\linewidth}
         \includegraphics[width=\linewidth,height= .8\linewidth]{figures/8gaussians-moons/KDEplot/NSGF/2.pdf}
   \end{subfigure}
   \begin{subfigure}{.19\linewidth}
         \includegraphics[width=\linewidth,height= .8\linewidth]{figures/8gaussians-moons/KDEplot/NSGF/3.pdf}
   \end{subfigure}
   \begin{subfigure}{.19\linewidth}
         \includegraphics[width=\linewidth,height= .8\linewidth]{figures/8gaussians-moons/KDEplot/NSGF/4.pdf}
   \end{subfigure}
   \begin{subfigure}{.19\linewidth}
         \includegraphics[width=\linewidth,height= .8\linewidth]{figures/8gaussians-moons/KDEplot/NSGF/5.pdf}
   \end{subfigure}

   \makebox[0pt][r]{\makebox[15pt]{\raisebox{15pt}{\rotatebox[origin=c]{90}{EPT}}}}
   \begin{subfigure}{.19\linewidth}
      \includegraphics[width=\linewidth,height= .8\linewidth]{figures/8gaussians-moons/KDEplot/EPT/1.pdf}
   \end{subfigure}
   \begin{subfigure}{.19\linewidth}
      \includegraphics[width=\linewidth,height= .8\linewidth]{figures/8gaussians-moons/KDEplot/EPT/2.pdf}
   \end{subfigure}
   \begin{subfigure}{.19\linewidth}
      \includegraphics[width=\linewidth,height= .8\linewidth]{figures/8gaussians-moons/KDEplot/EPT/3.pdf}
   \end{subfigure}
   \begin{subfigure}{.19\linewidth}
      \includegraphics[width=\linewidth,height= .8\linewidth]{figures/8gaussians-moons/KDEplot/EPT/4.pdf}
   \end{subfigure}
   \begin{subfigure}{.19\linewidth}
      \includegraphics[width=\linewidth,height= .8\linewidth]{figures/8gaussians-moons/KDEplot/EPT/5.pdf}
   \end{subfigure}

   \makebox[0pt][r]{\makebox[15pt]{\raisebox{15pt}{\rotatebox[origin=c]{90}{FM}}}}%
   \begin{subfigure}{.19\linewidth}
      \includegraphics[width=\linewidth,height= .8\linewidth]{figures/8gaussians-moons/KDEplot/CFM/1.pdf}
   \end{subfigure}
   \begin{subfigure}{.19\linewidth}
         \includegraphics[width=\linewidth,height= .8\linewidth]{figures/8gaussians-moons/KDEplot/CFM/2.pdf}
   \end{subfigure}
   \begin{subfigure}{.19\linewidth}
         \includegraphics[width=\linewidth,height= .8\linewidth]{figures/8gaussians-moons/KDEplot/CFM/3.pdf}
   \end{subfigure}
   \begin{subfigure}{.19\linewidth}
         \includegraphics[width=\linewidth,height= .8\linewidth]{figures/8gaussians-moons/KDEplot/CFM/4.pdf}
   \end{subfigure}
   \begin{subfigure}{.19\linewidth}
         \includegraphics[width=\linewidth,height= .8\linewidth]{figures/8gaussians-moons/KDEplot/CFM/5.pdf}
   \end{subfigure}

   \makebox[0pt][r]{\makebox[15pt]{\raisebox{15pt}{\rotatebox[origin=c]{90}{SI}}}}
   \begin{subfigure}{.19\linewidth}
      \includegraphics[width=\linewidth,height= .9\linewidth]{figures/8gaussians-moons/KDEplot/SI/1.pdf}
   \end{subfigure}
   \begin{subfigure}{.19\linewidth}
      \includegraphics[width=\linewidth,height= .9\linewidth]{figures/8gaussians-moons/KDEplot/SI/2.pdf}
   \end{subfigure}
   \begin{subfigure}{.19\linewidth}
      \includegraphics[width=\linewidth,height= .9\linewidth]{figures/8gaussians-moons/KDEplot/SI/3.pdf}
   \end{subfigure}
   \begin{subfigure}{.19\linewidth}
      \includegraphics[width=\linewidth,height= .9\linewidth]{figures/8gaussians-moons/KDEplot/SI/4.pdf}
   \end{subfigure}
   \begin{subfigure}{.19\linewidth}
      \includegraphics[width=\linewidth,height= .9\linewidth]{figures/8gaussians-moons/KDEplot/SI/5.pdf}
   \end{subfigure}

   \makebox[0pt][r]{\makebox[15pt]{\raisebox{15pt}{\rotatebox[origin=c]{90}{OT-CFM}}}}
   \begin{subfigure}{.19\linewidth}
      \includegraphics[width=\linewidth,height= .9\linewidth]{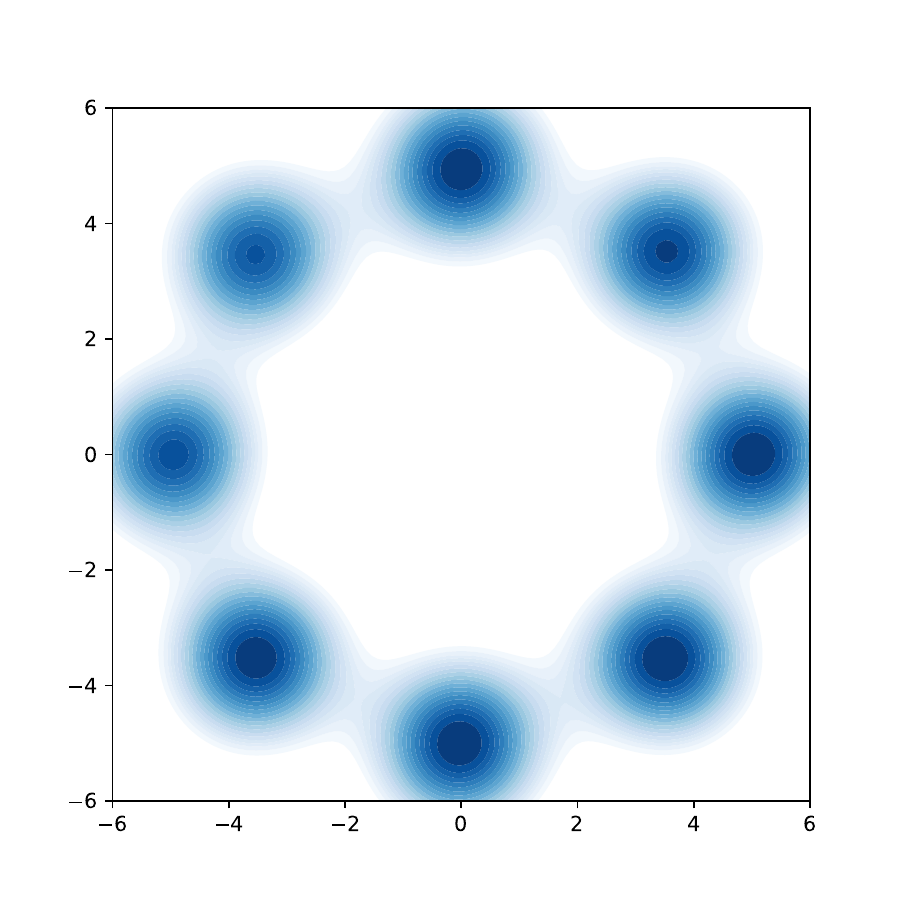}
   \end{subfigure}
   \begin{subfigure}{.19\linewidth}
      \includegraphics[width=\linewidth,height= .9\linewidth]{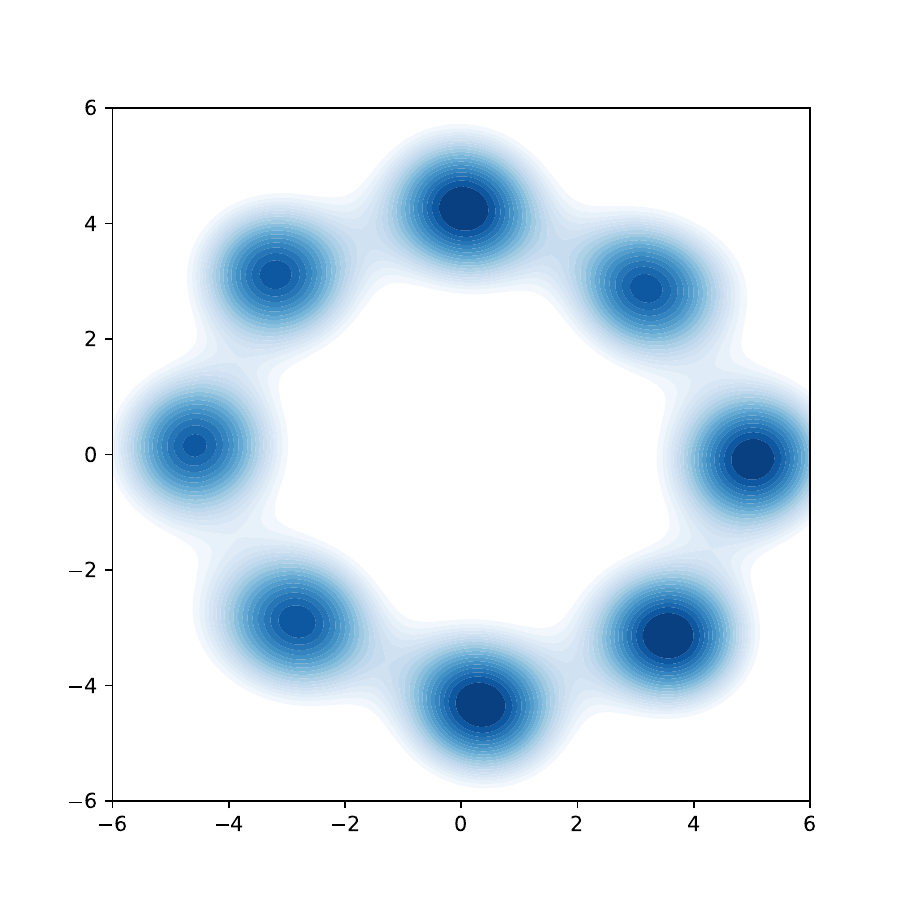}
   \end{subfigure}
   \begin{subfigure}{.19\linewidth}
      \includegraphics[width=\linewidth,height= .9\linewidth]{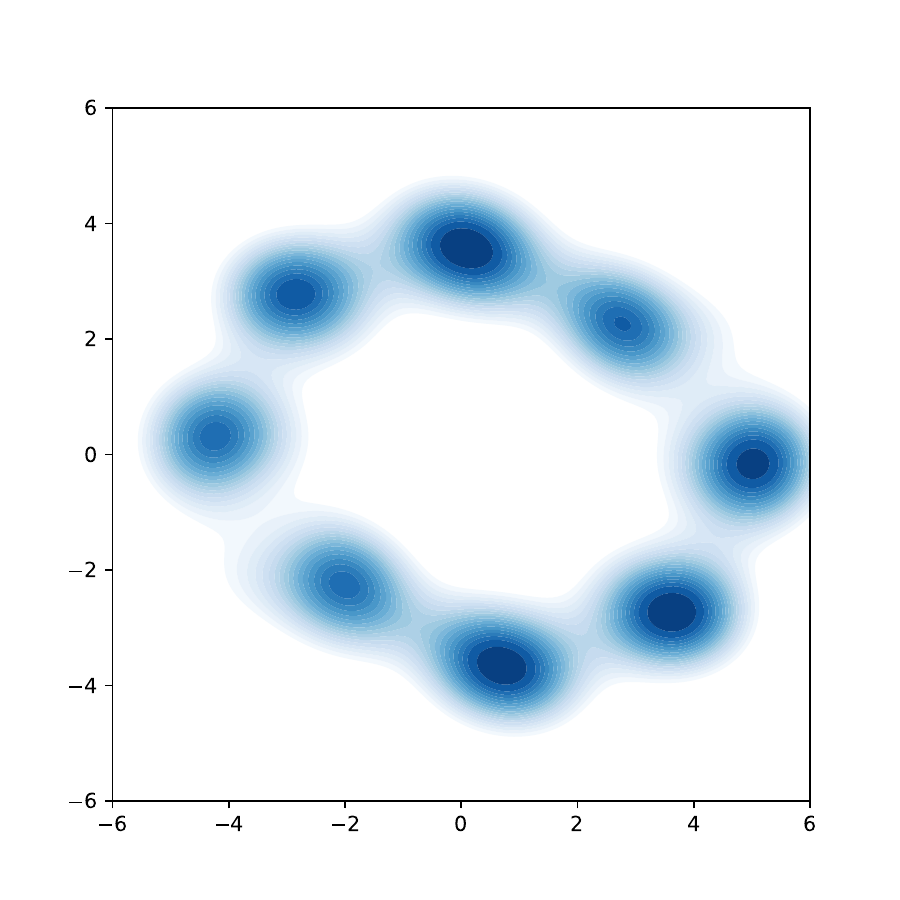}
   \end{subfigure}
   \begin{subfigure}{.19\linewidth}
      \includegraphics[width=\linewidth,height= .9\linewidth]{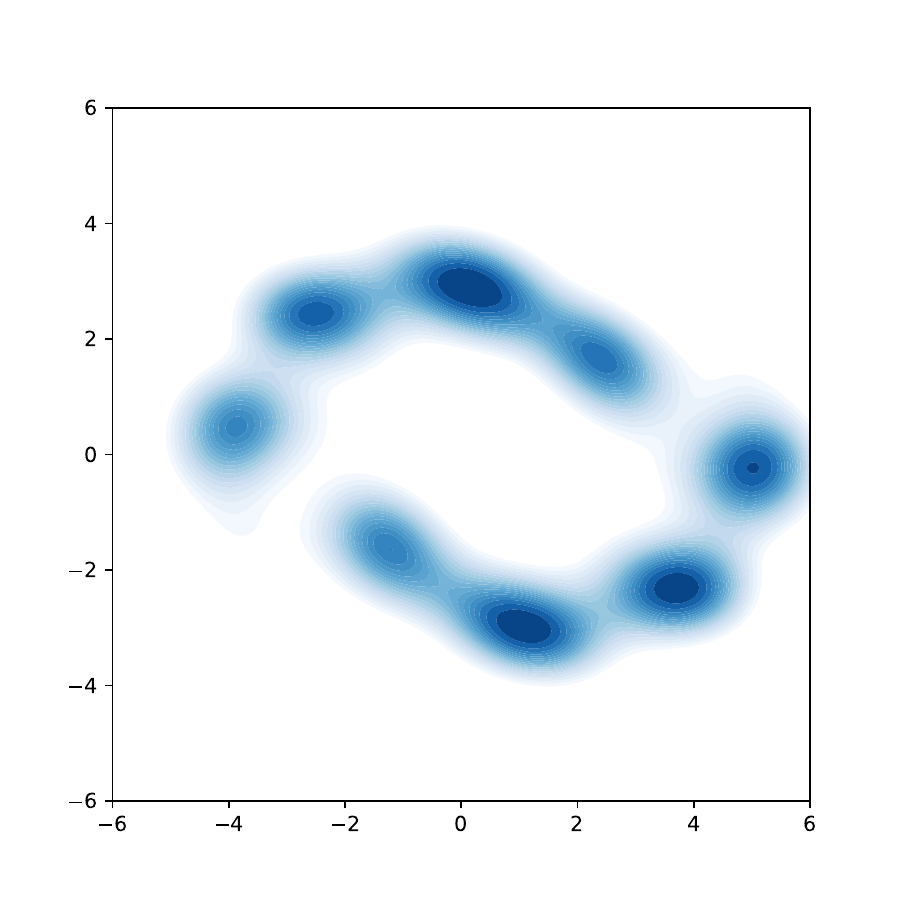}
   \end{subfigure}
   \begin{subfigure}{.19\linewidth}
      \includegraphics[width=\linewidth,height= .9\linewidth]{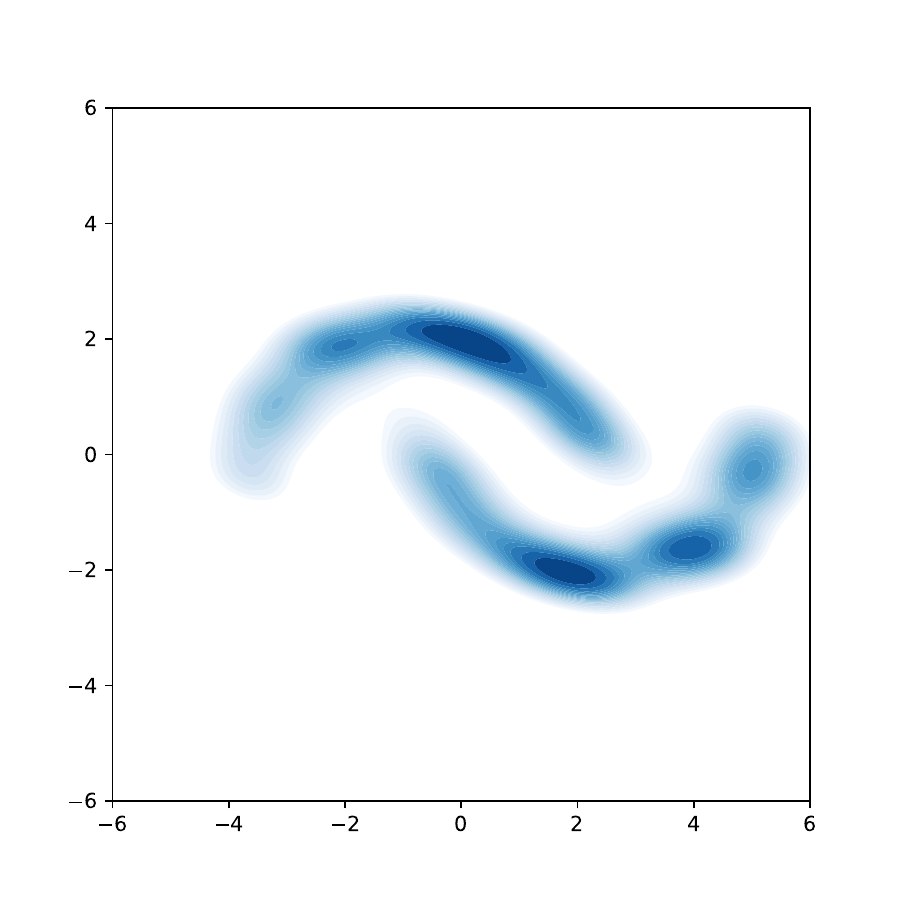}
   \end{subfigure}

   \makebox[0pt][r]{\makebox[15pt]{\raisebox{15pt}{\rotatebox[origin=c]{90}{RF-1}}}}
   \begin{subfigure}{.19\linewidth}
      \includegraphics[width=\linewidth,height= .9\linewidth]{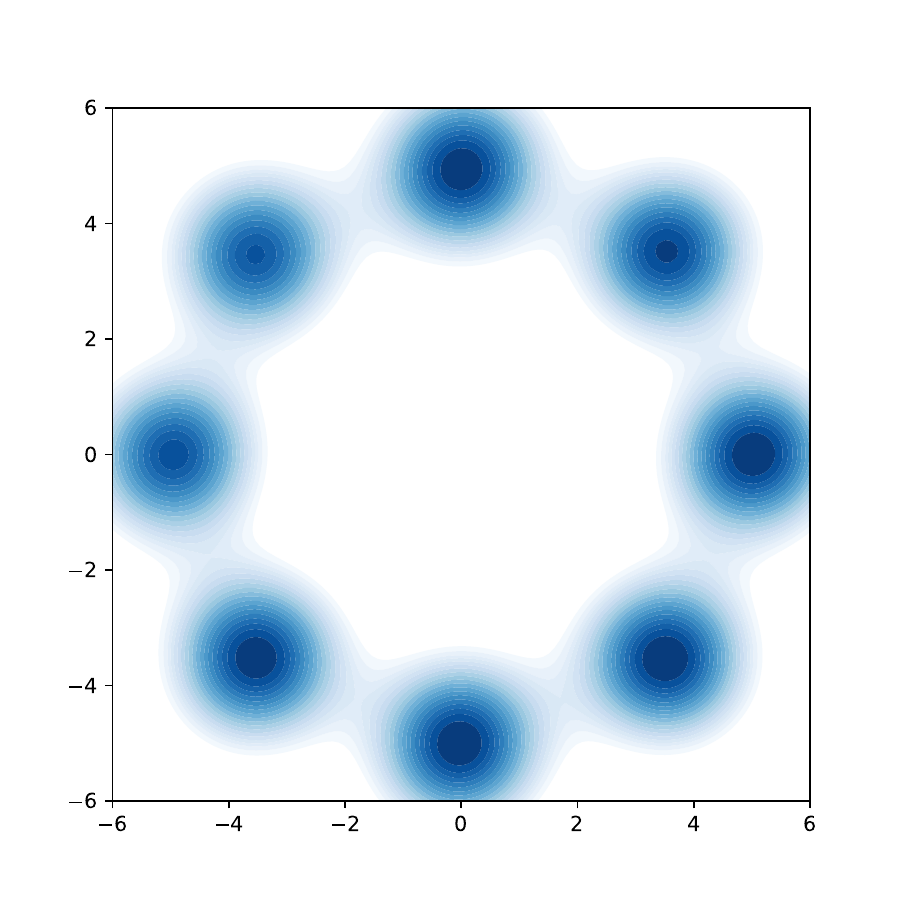}
   \end{subfigure}
   \begin{subfigure}{.19\linewidth}
      \includegraphics[width=\linewidth,height= .9\linewidth]{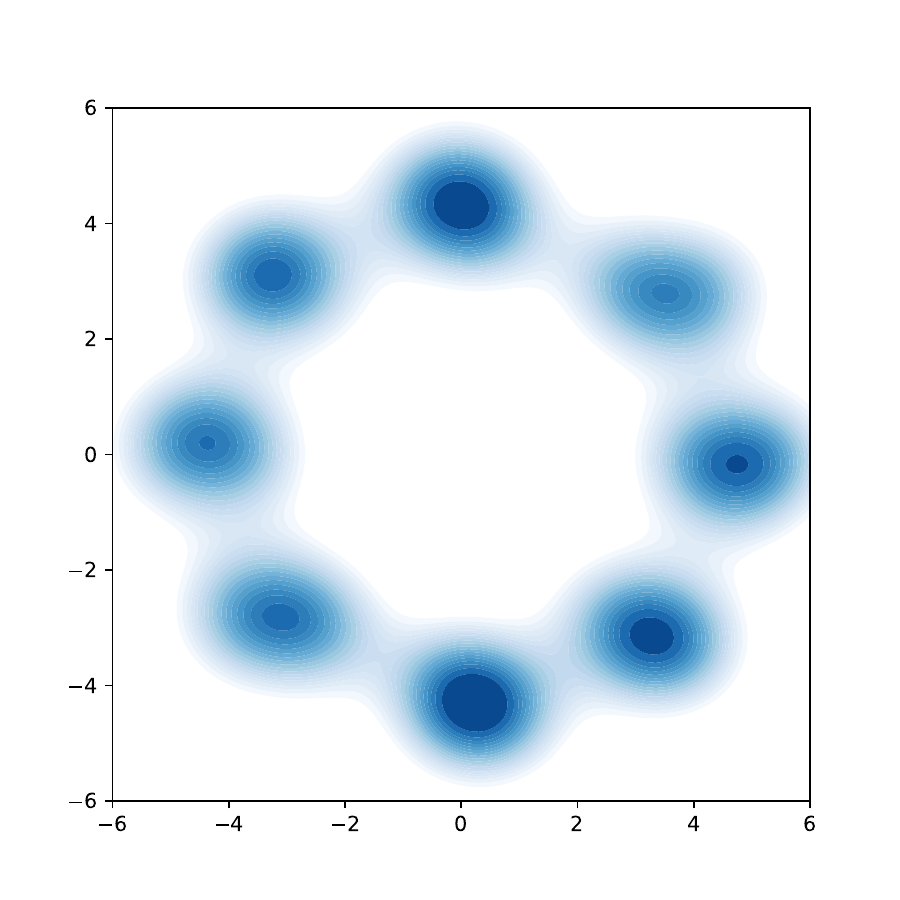}
   \end{subfigure}
   \begin{subfigure}{.19\linewidth}
      \includegraphics[width=\linewidth,height= .9\linewidth]{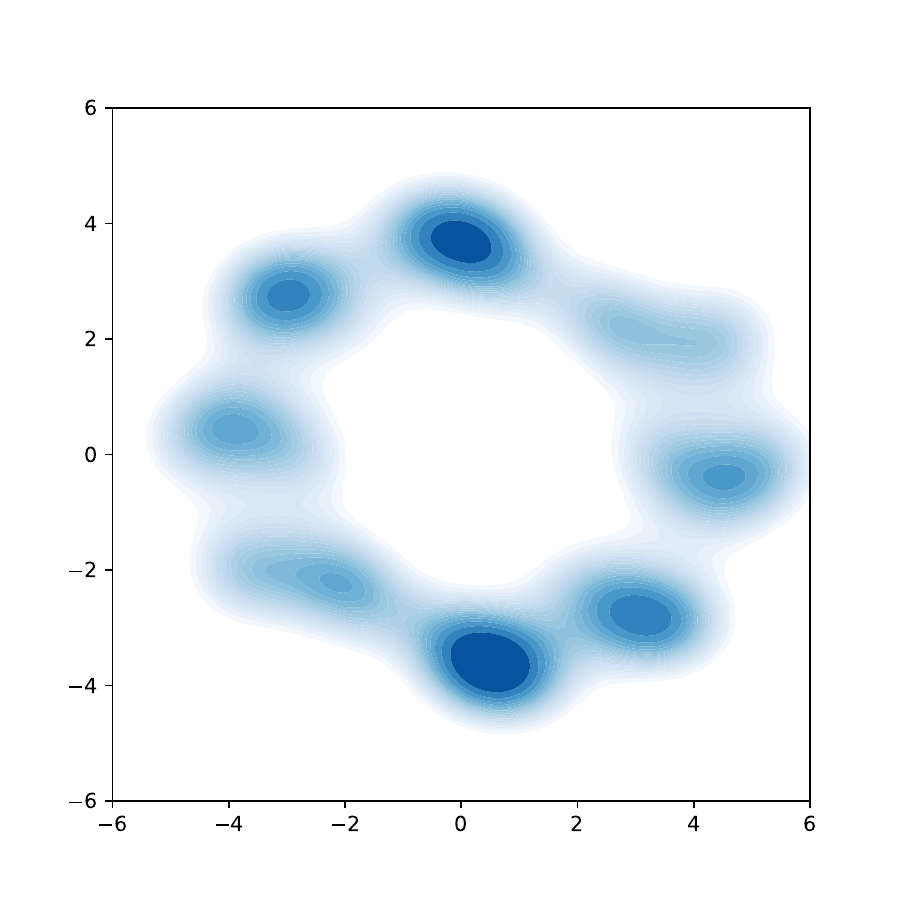}
   \end{subfigure}
   \begin{subfigure}{.19\linewidth}
      \includegraphics[width=\linewidth,height= .9\linewidth]{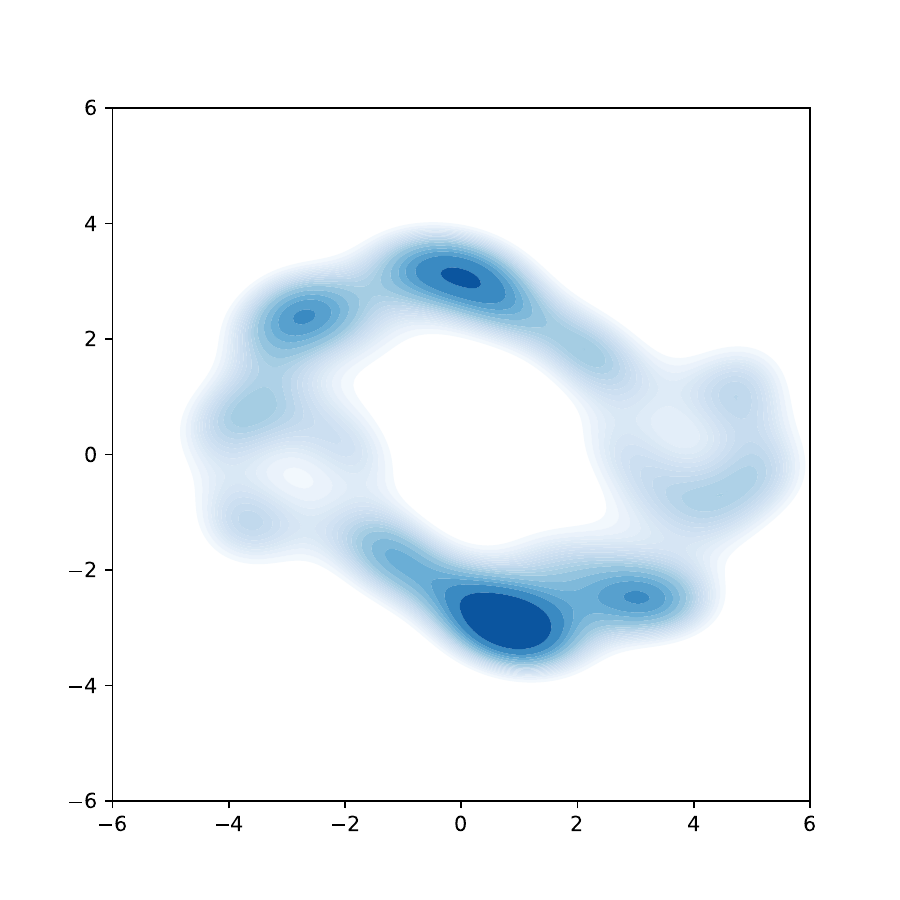}
   \end{subfigure}
   \begin{subfigure}{.19\linewidth}
      \includegraphics[width=\linewidth,height= .9\linewidth]{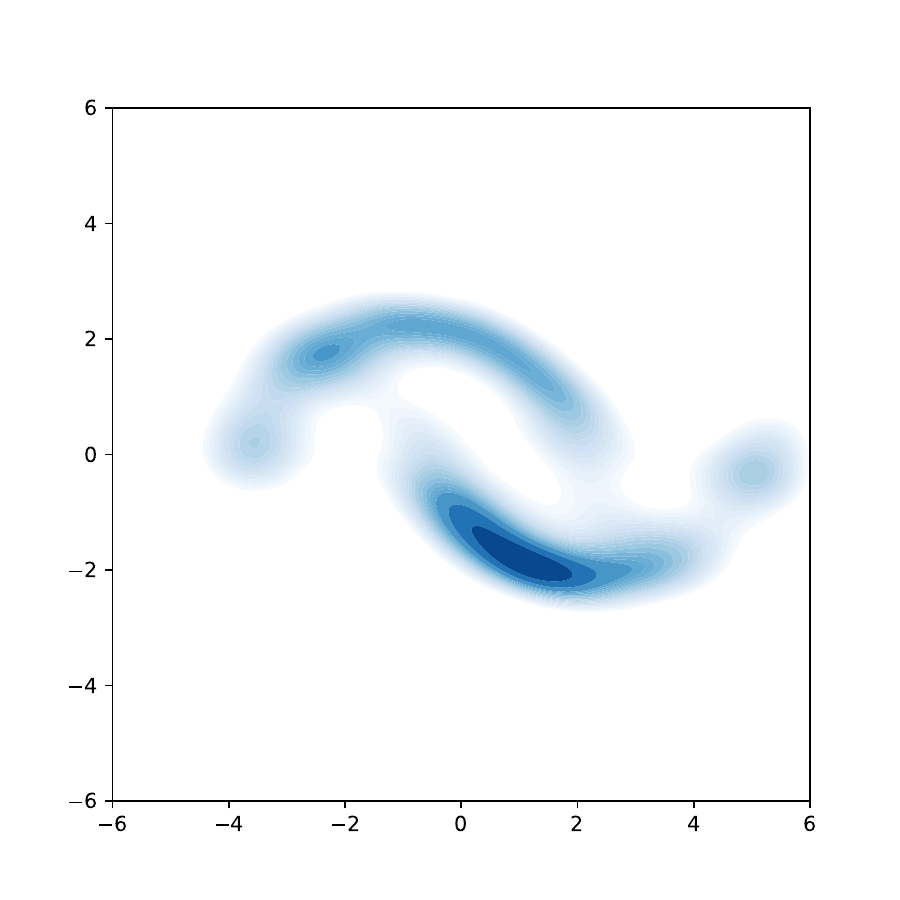}
   \end{subfigure}

   \makebox[0pt][r]{\makebox[15pt]{\raisebox{15pt}{\rotatebox[origin=c]{90}{RF-2}}}}
   \begin{subfigure}{.19\linewidth}
      \includegraphics[width=\linewidth,height= .9\linewidth]{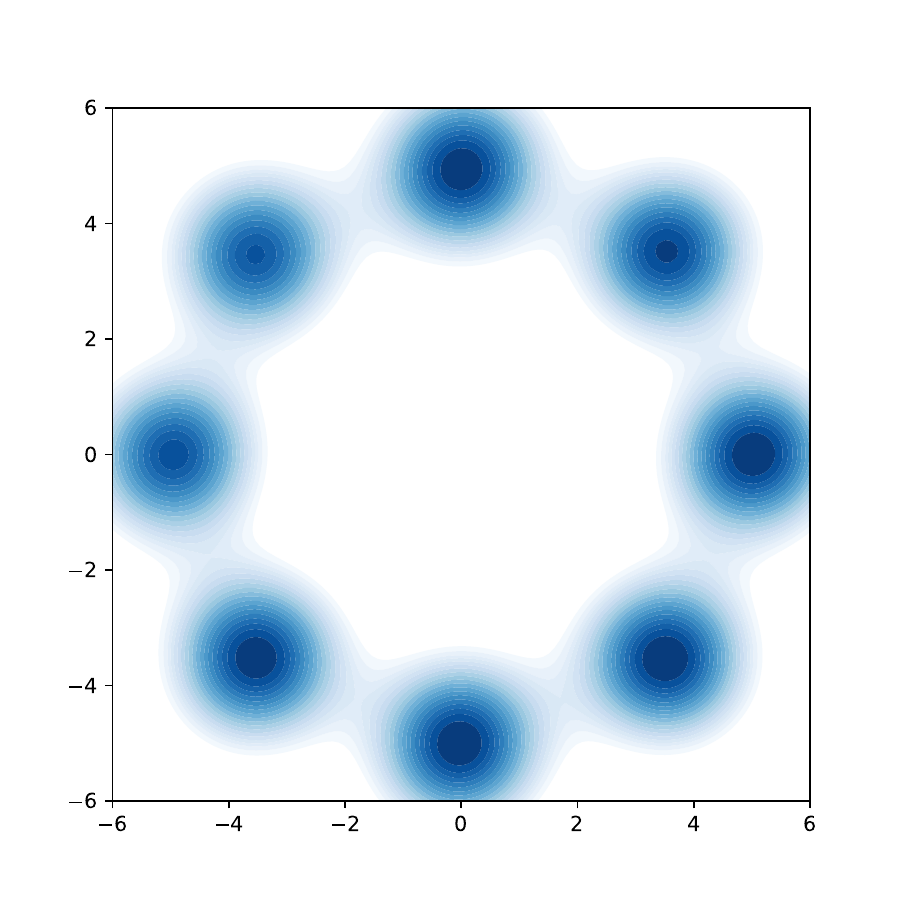}
   \end{subfigure}
   \begin{subfigure}{.19\linewidth}
      \includegraphics[width=\linewidth,height= .9\linewidth]{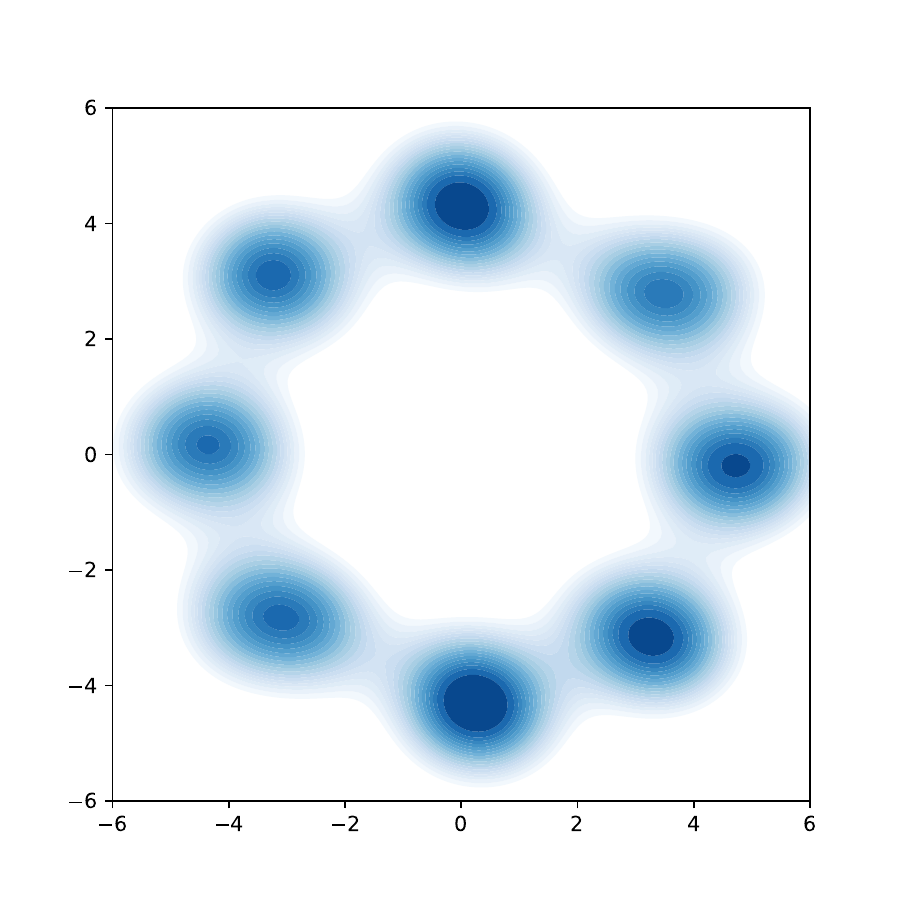}
   \end{subfigure}
   \begin{subfigure}{.19\linewidth}
      \includegraphics[width=\linewidth,height= .9\linewidth]{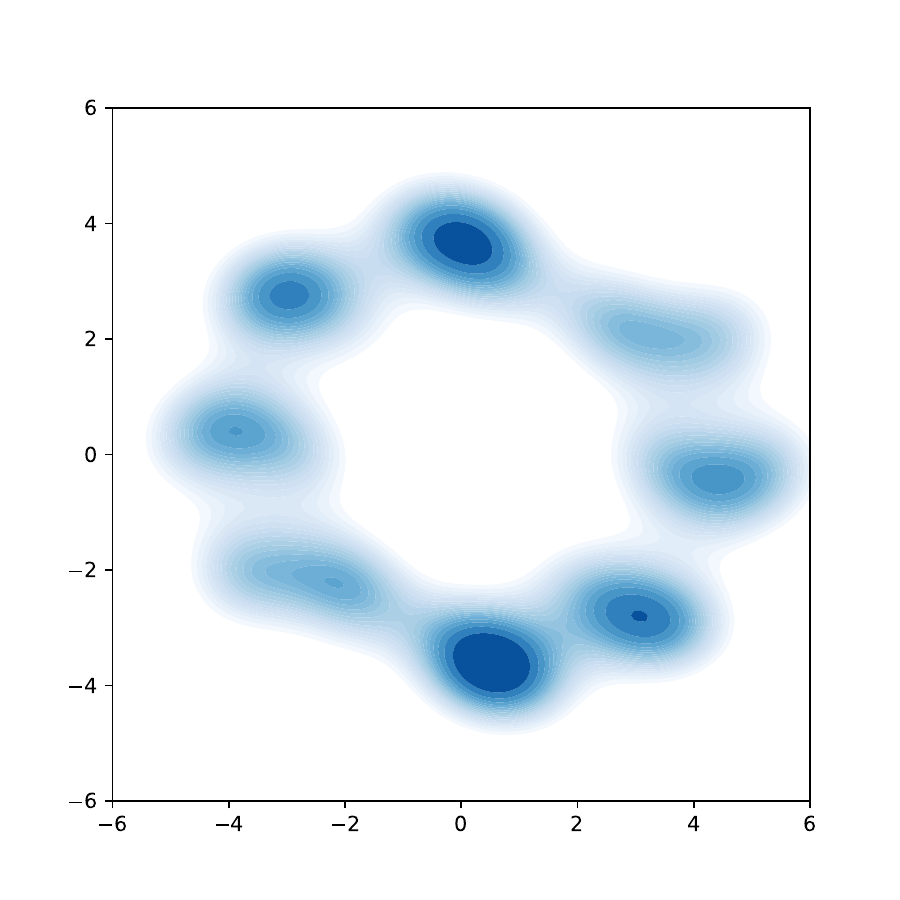}
   \end{subfigure}
   \begin{subfigure}{.19\linewidth}
      \includegraphics[width=\linewidth,height= .9\linewidth]{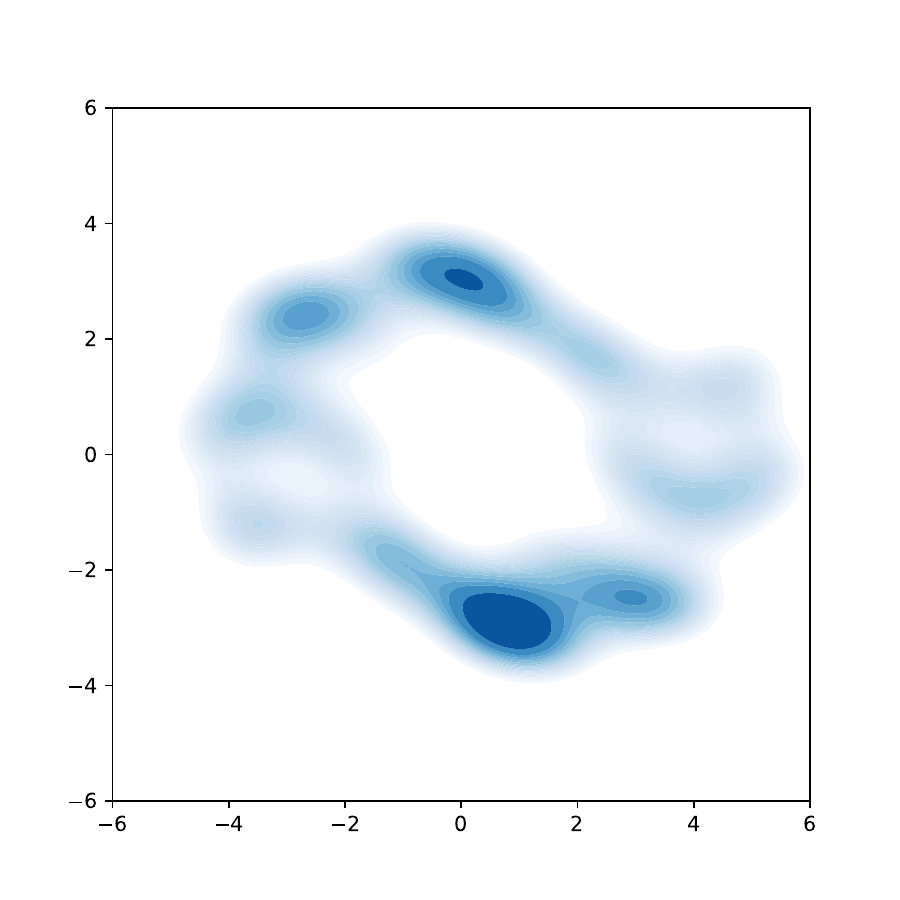}
   \end{subfigure}
   \begin{subfigure}{.19\linewidth}
      \includegraphics[width=\linewidth,height= .9\linewidth]{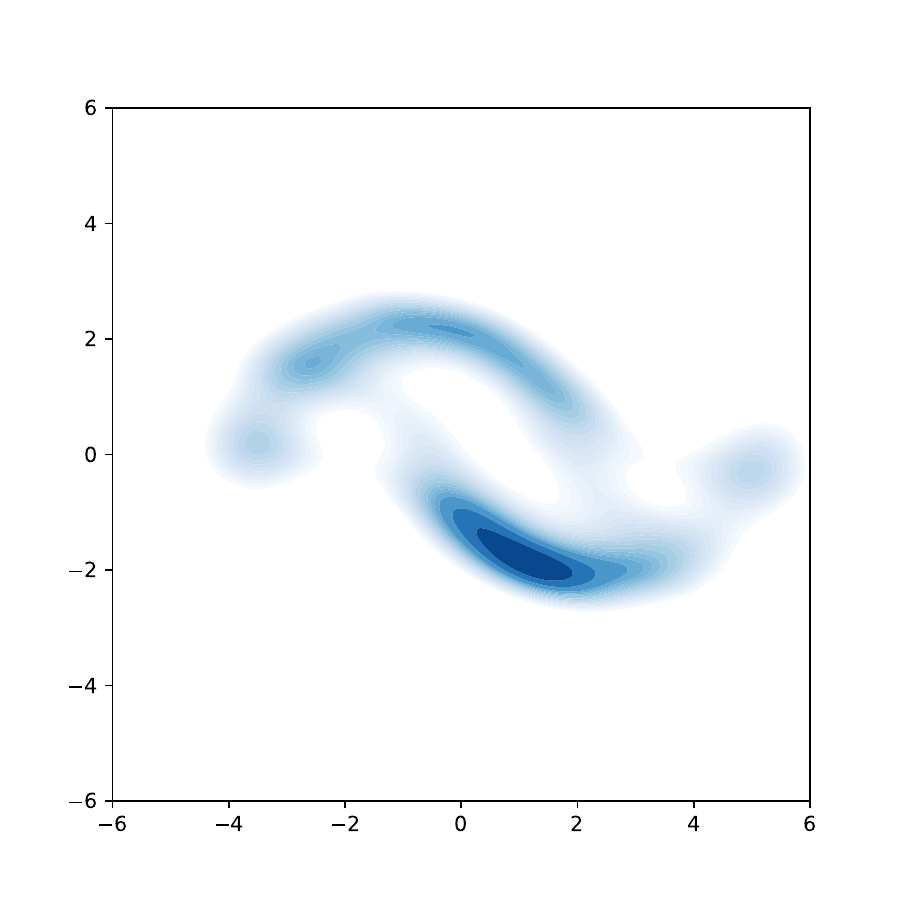}
   \end{subfigure}

   \caption{2-Wasserstein Distance of the generated process utilizing neural ODE-based diffusion models and NSGF. The FM/SI methods reduce noise roughly linearly, while NSGF quickly recovers the target structure and progressively optimizes the details in subsequent steps.}
   \label{KEDplot add}
   \end{figure}

In our supplementary materials, we present additional results from 2D simulated data to demonstrate the efficiency of the NSGF++ model in Figure \ref{2D stimulate data more results} and Figure \ref{KEDplot add}. These results indicate that NSGF++ achieves competitive performance with a more direct path and fewer steps compared to other neural Wasserstein gradient flow and flow-matching methods, highlighting its effectiveness and computational efficiency.

\subsubsection*{MNIST}
In our study, we include results from the MNIST dataset to showcase the efficiency of the NSGF++ model. As detailed in Table \ref{MNISTFID}, NSGF++ achieves competitive Fréchet Inception Distances (FIDs) while utilizing only 60\% of the Number of Function Evaluations (NFEs) typically required. This underscores the model's effectiveness in balancing performance with computational efficiency.  To evaluate our results, we use the Fréchet inception distance (FID) between 10K generated samples and the test dataset. Here, a smaller FID value indicates a higher similarity between generated and test samples.

\begin{table}[t]    
   \tiny
	\centering
    \resizebox{0.43\textwidth}{!}{%
	\begin{tabular}{c|cc}
		\toprule
		\multirow{2}*{Algorithm} & \multicolumn{2}{c}{MNIST}  \\
		~  & FID($\downarrow$) & NFE($\downarrow$) \\
		\hline
        \textbf{NSGF++(ours)}  & 3.8  &  60 \\
        SWGF\shortcite{liutkus2019sliced}  & 225.1  & 500 \\
        SIG\shortcite{dai2020sliced}  & 4.5  & / \\
      \hline
        FM \shortcite{lipman2023flow} & 3.4 & 100\\
        OT-CFM \shortcite{tong2023improving} & 3.3 & 100\\
        RF \shortcite{liu2023flow} & 3.1 & 100 \\
      \hline
         Training set & 2.27 & / \\
		\bottomrule
	\end{tabular}}
	\caption{Comparison of NSGF++ and other methods over MNIST, The last row states statistics of the FID scores between 10k training examples and 10k test examples}
    \label{MNISTFID}
\end{table}

\begin{figure}[t]
   \captionsetup[subfigure]{labelformat=empty}
   \centering
   \tiny
   \begin{subfigure}{0.49\linewidth}
       \centering
       \includegraphics[width=0.95\linewidth]{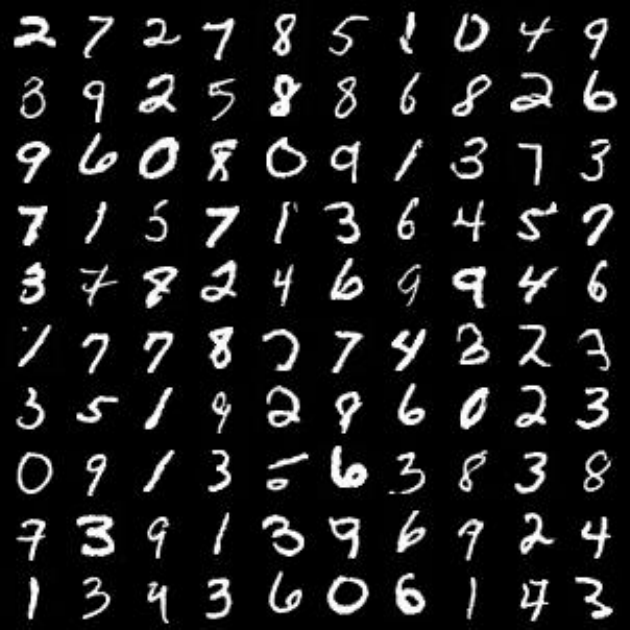}
   \end{subfigure}
   \begin{subfigure}{0.49\linewidth}
      \centering
      \includegraphics[width=0.95\linewidth]{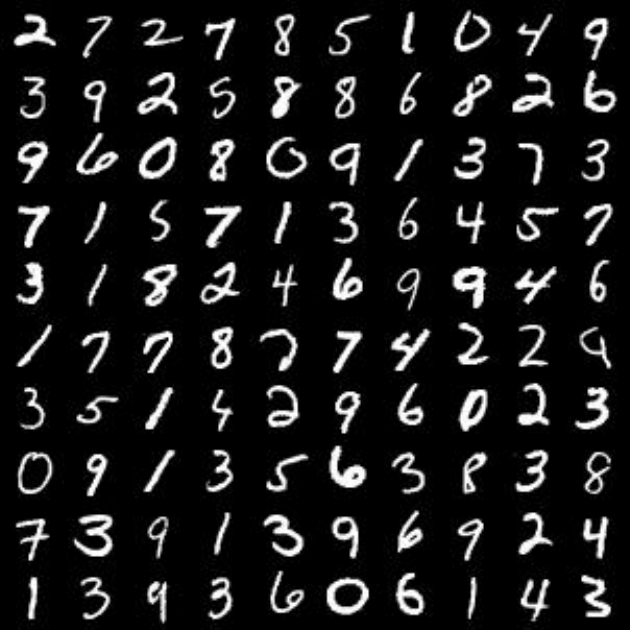}
   \end{subfigure}
   \caption{Uncurated samples on MNIST and $L2$-nearest neighbors from the training set (top: Samples, bottom: real)We observe that they are significantly different. Hence, our method generates really new samples and is not just reproducing the samples from the training set}
   \label{mnist nearest}
\end{figure}

\begin{figure}[b]
   \captionsetup[subfigure]{labelformat=empty}
   \centering
   \tiny
   \begin{subfigure}{0.49\linewidth}
       \centering
       \includegraphics[width=0.95\linewidth]{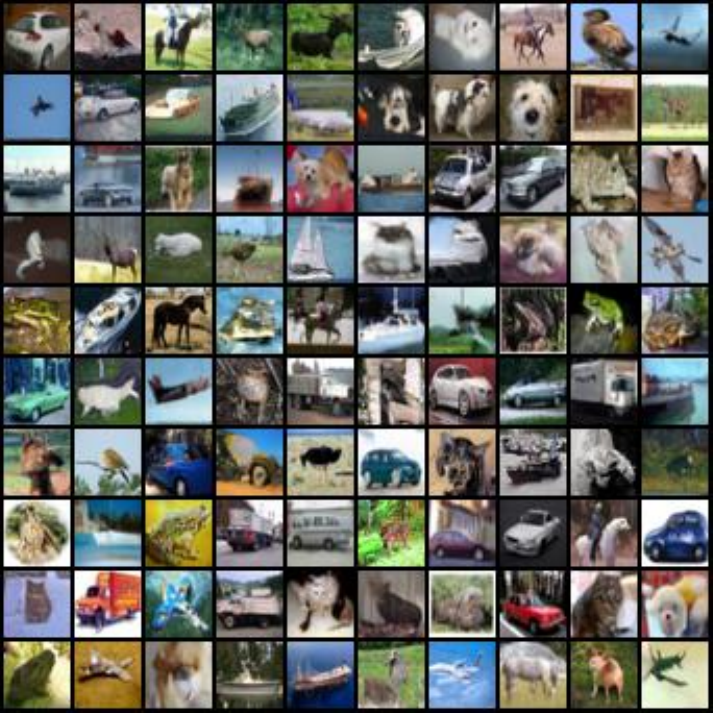}
   \end{subfigure}
   \begin{subfigure}{0.49\linewidth}
      \centering
      \includegraphics[width=0.95\linewidth]{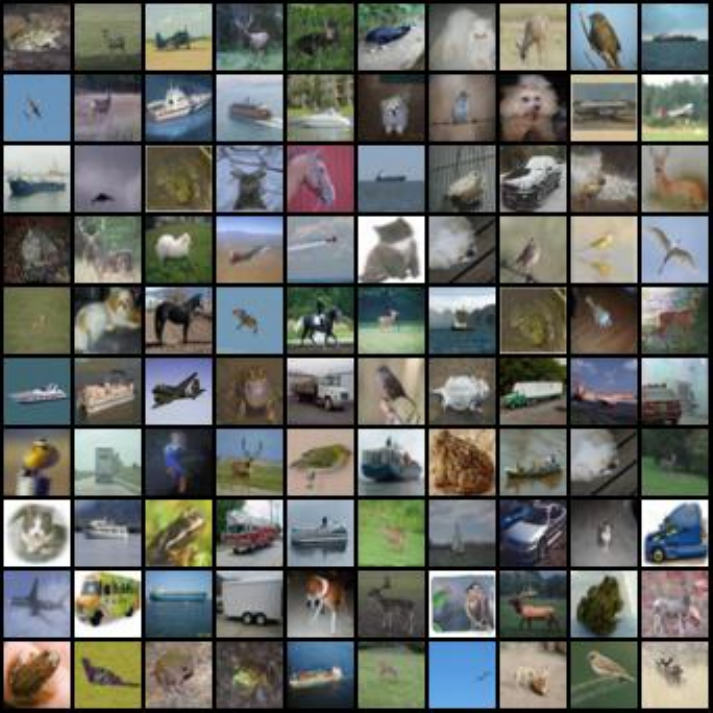}
   \end{subfigure}
   \caption{Uncurated samples on CIFAR-10 and $L2$-nearest neighbors from the training set (top: Samples, bottom: real)}
   \label{cifar10 nearest}
\end{figure}

\subsubsection*{CIFAR-10}
\begin{figure*}[t]
   \captionsetup[subfigure]{labelformat=empty}
   \centering
   \tiny
   \begin{subfigure}{0.8\linewidth}
       \centering
       \includegraphics[width=0.75\linewidth]{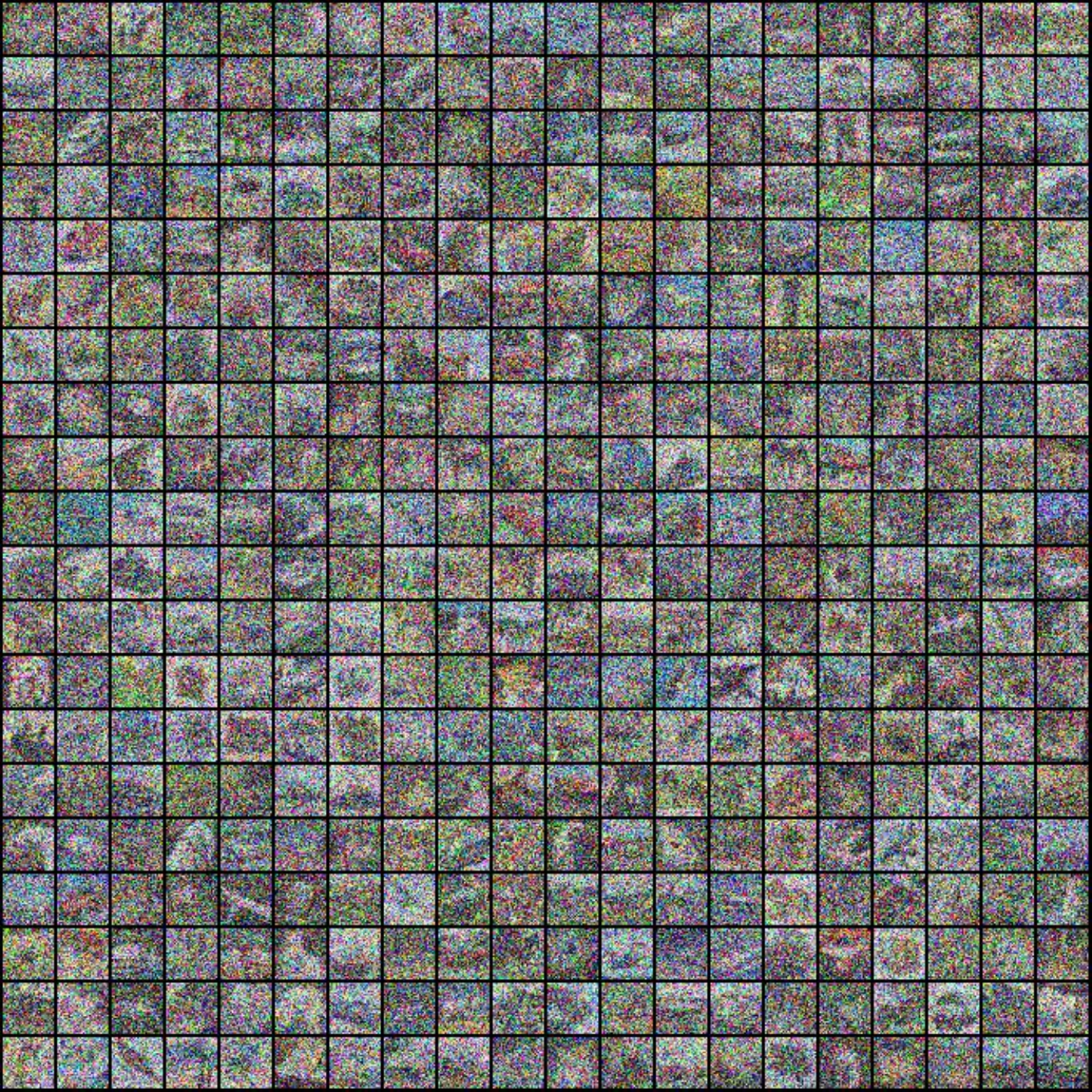}
   \end{subfigure}
   \begin{subfigure}{0.8\linewidth}
      \centering
      \includegraphics[width=0.75\linewidth]{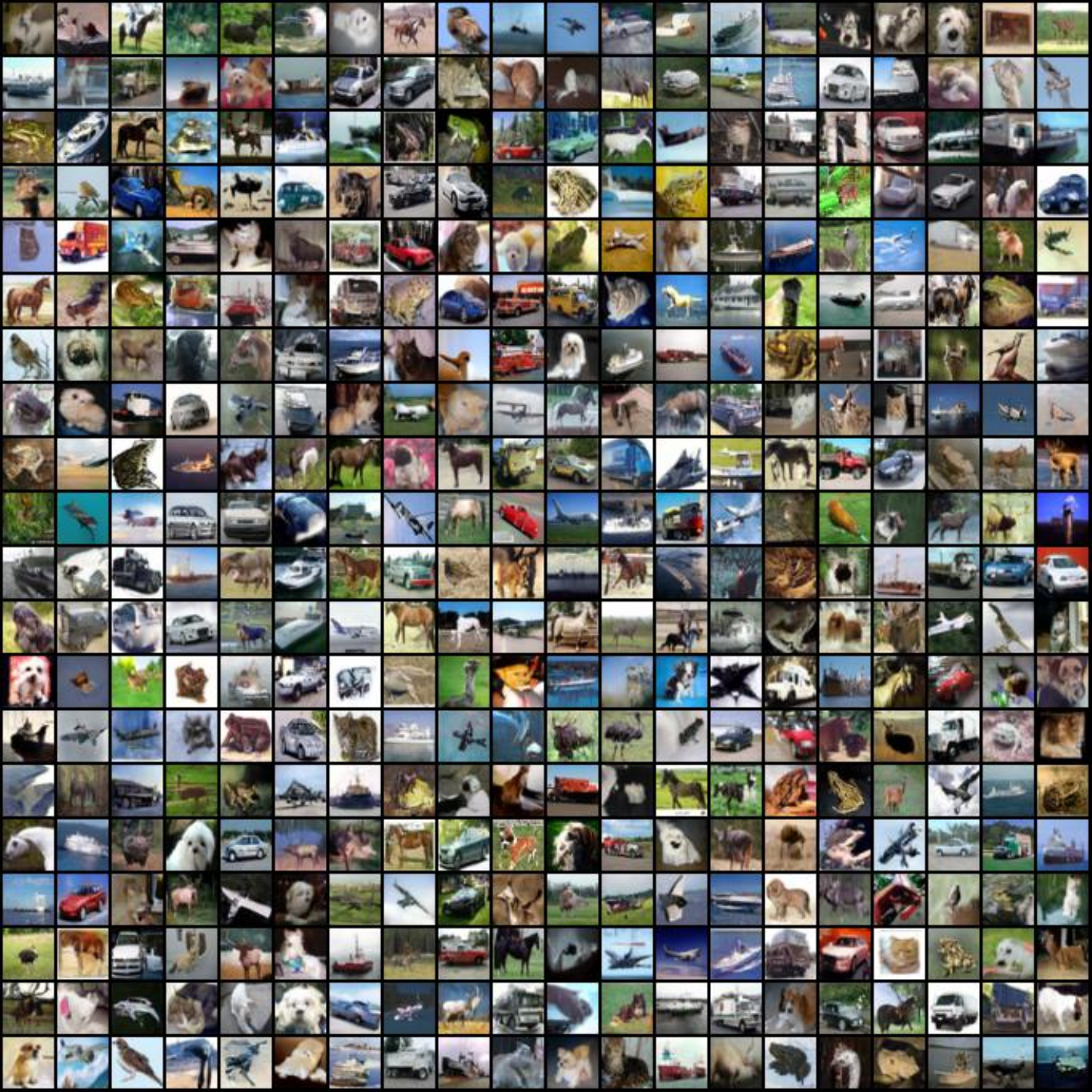}
  \end{subfigure}
   \caption{The extensive inference result of our NSGF++ model on CIFAR-10. The first row
   shows the result after 5 NSGF steps and the second row shows the
   final results}
   \label{multi cifar10}
\end{figure*}

In our work, we present further results of the NSGF++ model on the CIFAR-10 dataset, illustrated in Figures \ref{cifar10 nearest} and \ref{multi cifar10}. These experimental findings demonstrate that NSGF++ attains competitive performance in generation tasks, highlighting its efficacy.
\begin{figure*}[t]
   \captionsetup[subfigure]{labelformat=empty}
   \centering
   \tiny
   \begin{subfigure}{0.75\linewidth}
       \centering
       \includegraphics[width=0.8\linewidth]{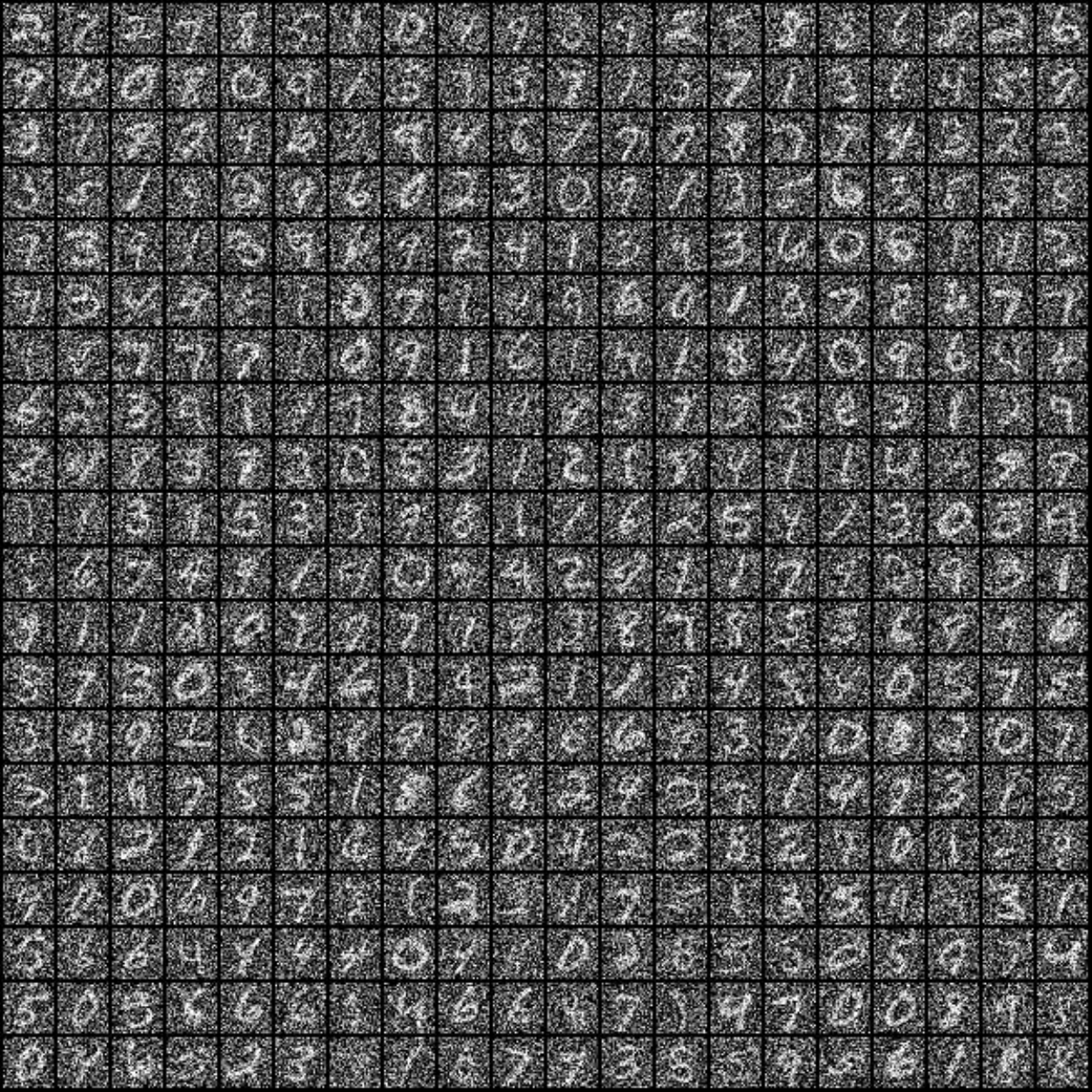}
   \end{subfigure}
   \begin{subfigure}{0.75\linewidth}
      \centering
      \includegraphics[width=0.8\linewidth]{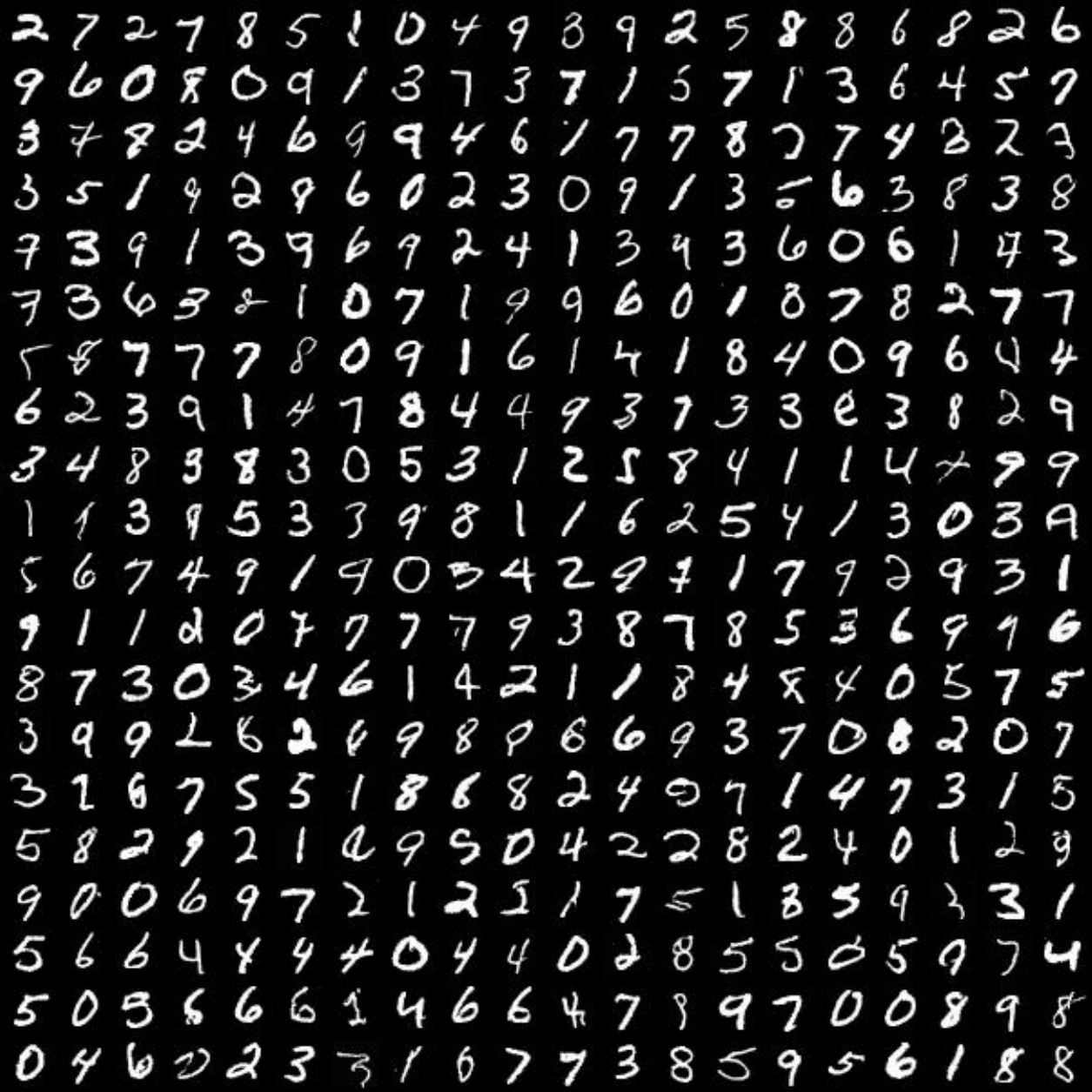}
  \end{subfigure}
   \caption{The extensive inference result of our NSGF++ model on MNIST. The first row
   shows the result after 5 NSGF steps and the second row shows the
   final results}
   \label{multi mnist}
\end{figure*}

\end{document}